\documentclass[12pt]{report} 
\usepackage
[
        left=1.5in,
        right=1.5in,
        top=2in,
        bottom=1in
]
{geometry}
\usepackage{mathtools,bm,amsfonts,amssymb,algorithm,algorithmicx,caption,algpseudocode,mathrsfs,multirow,multicol,subcaption,booktabs,threeparttable,balance,tikz,hyperref,breqn,dsfont,graphicx,fancyhdr,setspace,xspace,upgreek,newcent}

\usepackage[dvips]{changebar}
\nochangebars

\fancypagestyle{plain}{%
\fancyhead{}

}
\newcommand{\nas}{\textsc{NAS} }
\newcommand{\nasns}{\textsc{NAS}}

\newcommand{\mlas}{\textsc{MLAS} }
\newcommand{\mlasns}{\textsc{MLAS}}

\newcommand{\olas}{\textsc{OLAS} }
\newcommand{\olasns}{\textsc{OLAS}}

\newcommand{\amas}{\textsc{AMAS} }
\newcommand{\amasns}{\textsc{AMAS}}

\newcommand{\anet}{\texttt{AttNet} }
\newcommand{\anetns}{\texttt{AttNet}}

\newcommand{\snet}{\texttt{SeqNet} }
\newcommand{\snetns}{\texttt{SeqNet}}

\newcommand{\fnet}{\texttt{FusionNet} }
\newcommand{\fnetns}{\texttt{FusionNet}}

\newcommand{\mnet}{\texttt{MetricNet} }

\newcommand{\cnet}{\texttt{CoreNet} }
\newcommand{\cnetns}{\texttt{CoreNet}}

\newcommand{\vnet}{\textsf{PredictNet} }
\newcommand{\vnetns}{\textsf{PredictNet}}

\newcommand{\attn}{\texttt{Attention Block} }
\newcommand{\attns}{\texttt{Attention Block}}

\newcommand{\ie}{\emph{i.e.}}
\newcommand{\eg}{\emph{e.g.}}
\newcommand{\etc}{\emph{etc}}
\DeclareMathOperator*{\minimize}{minimize}

\DeclareMathOperator*{\argmin}{arg\,min}

\makeatletter
\newcommand*\bigcdot{\mathpalette\bigcdot@{.8}}
\newcommand*\bigcdot@[2]{\mathbin{\vcenter{\hbox{\scalebox{#2}{$\m@th#1\bullet$}}}}}
\makeatother

\usepackage{varwidth}
\DeclareCaptionFormat{centercaption}{%
  \begin{varwidth}{\linewidth}%
    \centering
    #1#2#3%
  \end{varwidth}%
}
  
\renewcommand{\vec}[1]{\bm{#1}}
\algnewcommand\algorithmicinput{\textbf{INPUT:}}
\algnewcommand\INPUT{\item[\algorithmicinput]}
\algnewcommand\algorithmicoutput{\textbf{OUTPUT:}}
\algnewcommand\OUTPUT{\item[\algorithmicoutput]}
\algnewcommand\algorithmicforeach{\textbf{for each}}
\algdef{S}[FOR]{ForEach}[1]{\algorithmicforeach\ #1\ \algorithmicdo}

\newtheorem{definition}{Definition}

\title{\vspace{-4.5cm}\Large\bf Deep Learning on Attributed Sequences\vspace{-3mm}}
\author{\large by \\ [3mm] Zhongfang Zhuang\vspace{3mm}}
\date{
 A Dissertation \\ \vfill
Submitted to the Faculty \\ \vfill
of the \\ \vfill
 WORCESTER POLYTECHNIC INSTITUTE\\ \vfill
In partial fulfillment of the requirements for the \\ \vfill
Degree of Doctor of Philosophy \\ \vfill
in\\ \vfill
Computer Science \\ \vfill
February 28, 2019 \\
\vfill
}

\begin{document}
\pagestyle{fancy}
\pagenumbering{arabic}
\maketitle
 \renewcommand{\baselinestretch}{2}
 \captionsetup[table]{font={stretch=1.5}}     
\captionsetup[figure]{font={stretch=1.5}} 
\selectfont

\newpage

\setcounter{secnumdepth}{4}
\tableofcontents

\newpage
\listoffigures


\newpage
\noindent
\noindent
\begin{center}
{\large \bf Abstract}
\end{center}
Recent research in feature learning has been extended to sequence data, where each instance consists of a sequence of heterogeneous items with a variable length. However, in many real-world applications, the data exists in the form of attributed sequences, which is composed of a set of fixed-size attributes and variable-length sequences with dependencies between them. In the attributed sequence context, feature learning remains challenging due to the dependencies between sequences and their associated attributes. In this dissertation, we focus on analyzing and building deep learning models for four new problems on attributed sequences. 

First, we propose a framework, called \nasns, to produce feature representations of attributed sequences in an \textit{unsupervised} fashion. The \nas is capable of producing task independent embeddings that can be used in various mining tasks of attributed sequences.  

Second, we study the problem of deep metric learning on attributed sequences. The goal is to learn a distance metric based on \textit{pairwise user feedback}. In this task, we propose a framework, called \mlasns, to learn a distance metric that measures the similarity and dissimilarity between attributed sequence feedback pairs. 

Third, we study the problem of one-shot learning on attributed sequences. This problem is important for a variety of real-world applications ranging from fraud prevention to network intrusion detection. We design a deep learning framework \olas to tackle this problem. Once the \olas is trained, we can then use it to make predictions for not only the new data but also for entire previously unseen new classes. 

Lastly, we investigate the problem of attributed sequence classification with attention model. This is challenging that now we need to assess the importance of each item in each sequence considering both the sequence itself and the associated attributes. In this work, we propose a framework, called \amasns, to classify attributed sequences using the information from the sequences, metadata, and the computed attention. 

Our extensive experiments on real-world datasets demonstrate that the proposed solutions significantly improve the performance of each task over the state-of-the-art methods on attributed sequences.

\newpage
\newpage
\noindent
\noindent
\begin{center}
{\large \bf Acknowledgments}
\end{center}

I would like to express my gratitude to my advisor Dr. Elke Rundensteiner and co-advisor Dr. Xiangnan Kong for their excellent guidance, patience, and support for my research. I would like to thank Dr. Mohamed Eltabakh for collaborating with me on the recurring query processing projects during my Ph.D. qualifier. My special thank you goes to Dr. Philip Yu for devoting his time and efforts to serve on my Ph.D. committee. 

My sincere thank you also goes to the Computer Science Department at Worcester Polytechnic Institute for the generous Teaching Assistantship offers in the early years of my Ph.D. study. This Teaching Assistantship enables and encourages me to follow my passion for computer science. I am also grateful for the generosity of Amadeus IT Group for supporting my research through the Research Assistantship as well as access to real-world data and problems.

I sincerely thank my friend and collaborator, Dr. Chuan Lei, for sharing his experiences in research during our collaborations of the big data system research. My sincere thank you also goes to Dr. Jihane Zouaoui and R{\'e}mi Domingues from Amadeus IT Group for the collaborations in deep learning research. 

My thank you also goes to current and alumni DSRG students Dr. Dongqing Xiao, Xiao Qin, Xinyue Liu, Mingrui Wei, and Dr. Chiying Wang for the support and advice during my Ph.D. years.

Last but not least, I want to thank my family for unconditionally supporting me to pursue my Ph.D.

\newpage
\pagestyle{fancy}
 \chapter{Introduction}
\label{chapter-intro}
\newpage
\section{The Prevalence of Attributed Sequences}
\label{ch1-section-attseq}
Sequential data arises naturally in a wide range of applications~\cite{bechet2015sequence, miliaraki2013mind, wang2016unsupervised, wei2013effective}. Examples of sequential data include clickstreams of web users, purchase histories of online customers, and DNA sequences of genes. Different from conventional multidimensional data~\cite{pedersen1999multidimensional} and time series data~\cite{khaleghi2016consistent}, sequential data~\cite{yang2003cluseq} are not represented as feature representations of continuous values, but as sequences of categorical items with variable-lengths. 

The sequential data in many real-world applications also often comes with a set of data attributes depicting the context. In this work, we will call this type of data, where each instance has both sequential data and the attributes, \textit{attributed sequences}. 
\begin{figure}[!ht]
    \centering
    \includegraphics[width=0.8\linewidth]{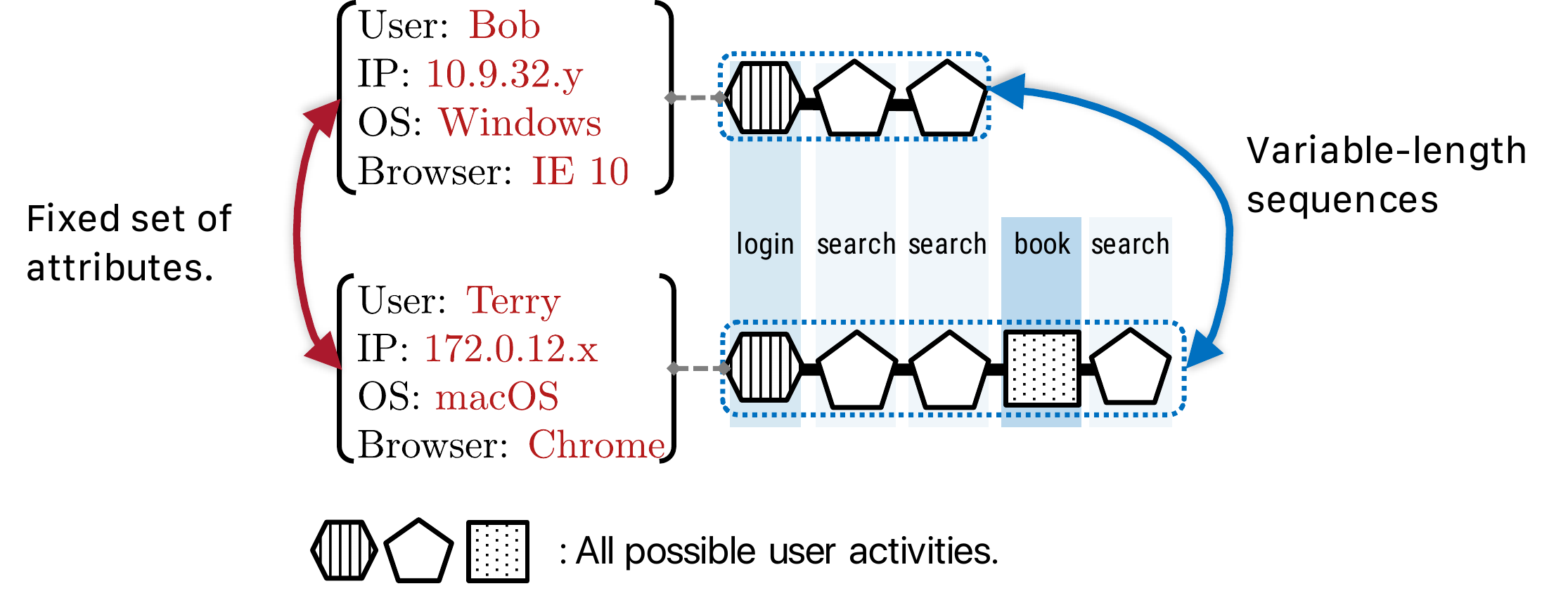}
    \caption[An Example of Attributed Sequences.]{An example of attributed sequences. Each attributed sequence represents a transaction in an online ticketing system, which includes a \textit{sequence} of user actions ({\eg}, ``\texttt{login}'', ``\texttt{search}'', ``\texttt{pick seats}'' and ``\texttt{confirm}'') and a set of  \textit{attributes} for the transactional context.}
    \label{fig-attrseq}
\end{figure}

For example, in online ticketing systems as shown in Figure~\ref{fig-attrseq}, each user transaction includes both a sequence of user actions (\eg, ``\texttt{login}'', ``\texttt{search}'' and ``\texttt{pick seats}'') and a set of attributes (\eg, ``\texttt{user name}'', ``\texttt{browser}'' and ``\texttt{IP address}'') indicating the context of the transaction. In gene function analysis, each gene can be represented by both a DNA sequence and a set of attributes indicating the expression levels of the gene in different types of cells. In the area of web search, an attributed sequence is composed of a static user profile (\eg, ``\texttt{geolocation}'') and the dynamic sequence of keywords searched (\eg, ``\texttt{snow storm}'' and ``\texttt{temperature}'').

Many real-world applications involve mining tasks over the sequential data. For example, in online ticketing systems, administrators are interested in finding fraudulent sequences from the clickstreams of users~\cite{tajer2014outlying, wei2013effective}. In user profiling systems, researchers are interested in grouping purchase histories of customers into clusters~\cite{wang2016unsupervised}. 
Motivated by these real-world applications, sequential data mining has received considerable attention in recent years~\cite{miliaraki2013mind, bechet2015sequence}. However, the attributes associated with these sequential data in these real-world applications have been overlooked despite their importance. Below, we highlight the importance of using attributed sequences in two real-world applications. 

\section{Applications using Attributed Sequences}
\label{ch1-section-motivation}
\begin{figure}[!ht]
    \centering
    \includegraphics[width=0.8\linewidth]{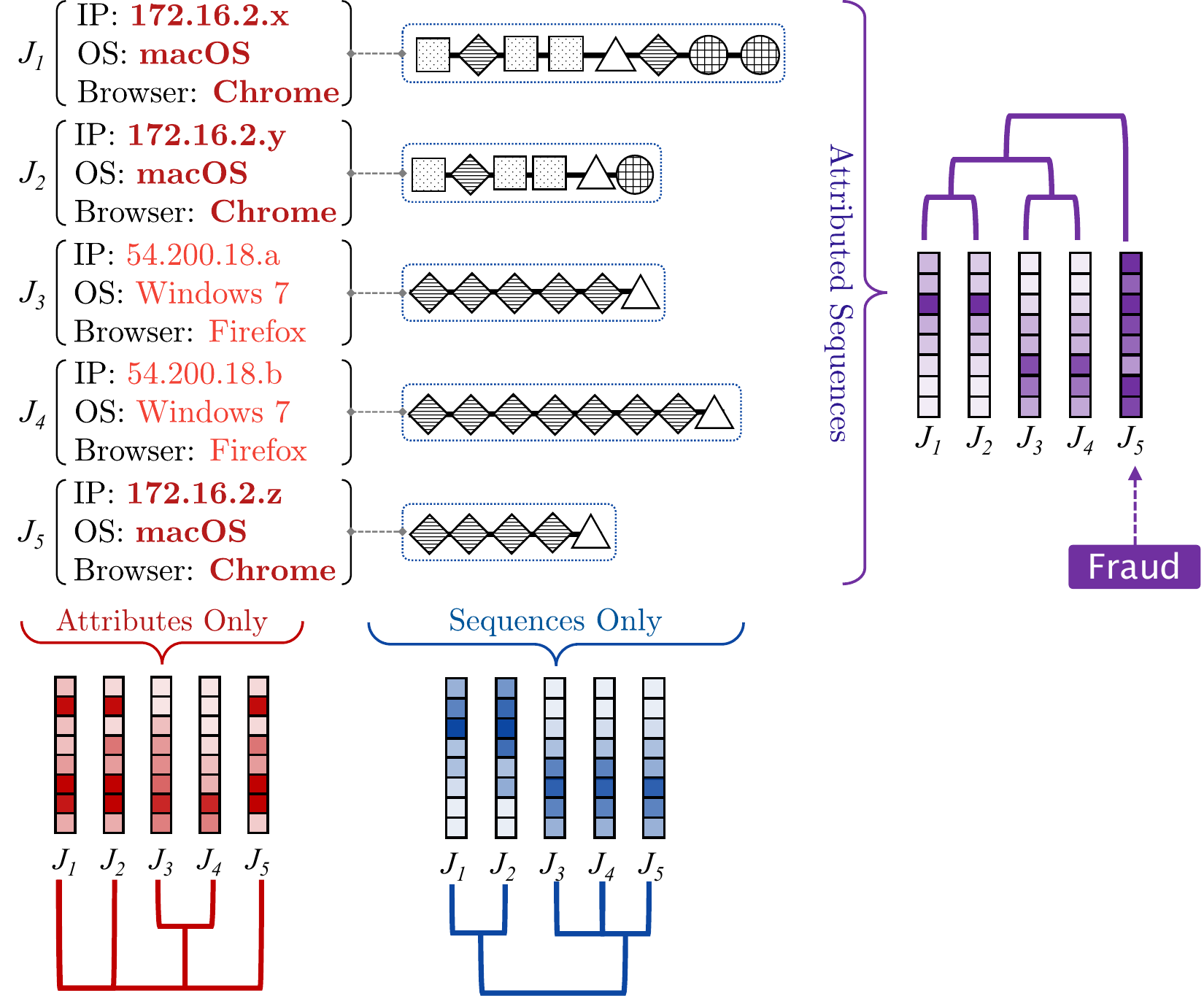}
    \caption[Fraud Detection using Attributed Sequence Embedding]{Dendrograms of feature representations learned from attributed sequences for fraud detection tasks. $J_5$ is a user committing fraud. However, it is considered a normal user session by the feature representation generated using either only attributes or only sequences. $J_5$ can only be caught as a fraud instance using the feature representation generated with regards to both attributes and sequences.}
    \label{fig-as2vec}
\end{figure}
\begin{enumerate}
    \item \textbf{Airline Ticket Booking Fraud.} In an airline ticketing system, whether a sequence of user actions is fraudulent depends closely upon the context of the transaction. A sequence of user actions, like ``\textit{apply business travel discount}, then \textit{book a ticket}'', will be normal for a \textit{business travel} context, but could be fraudulent for a \textit{leisure travel} context. 
    In Figure~\ref{fig-as2vec}, we present an example of fraud detection from a user privilege management system in Amadeus~\cite{amadeus}. This system logs each user session as an attributed sequence (denoted as $J_1 \sim J_5$). Each attributed sequence consists of a sequence of user's activities and a set of attributes derived from metadata values. The attributes ({\eg}, ``\texttt{IP}'', ``\texttt{OS}'' and ``\texttt{Browser}'') are recorded when a user logs into the system and remain unchanged during each user session. We use different shapes and colors to denote different user activities, {\eg}, ``\texttt{Reset password}'', ``\texttt{Delete a user}''. An important step in this fraud detection system is to ``\textit{red flag}'' suspicious user sessions for potential security breaches. In Figure~\ref{fig-as2vec}, we observe three groups of feature representations learned from the Amadeus application logs. For each group, we use a dendrogram to demonstrate the similarities between feature representations within that group. Neither of the feature representations using only sequences or only attributes detects any outliers due to not considering attribute-sequence dependencies. However, user session $J_5$ is discovered to be fraudulent using a learning algorithm that incorporates all three types of dependencies. 
\begin{figure}[!ht]
    \centering
       \includegraphics[width=\linewidth]{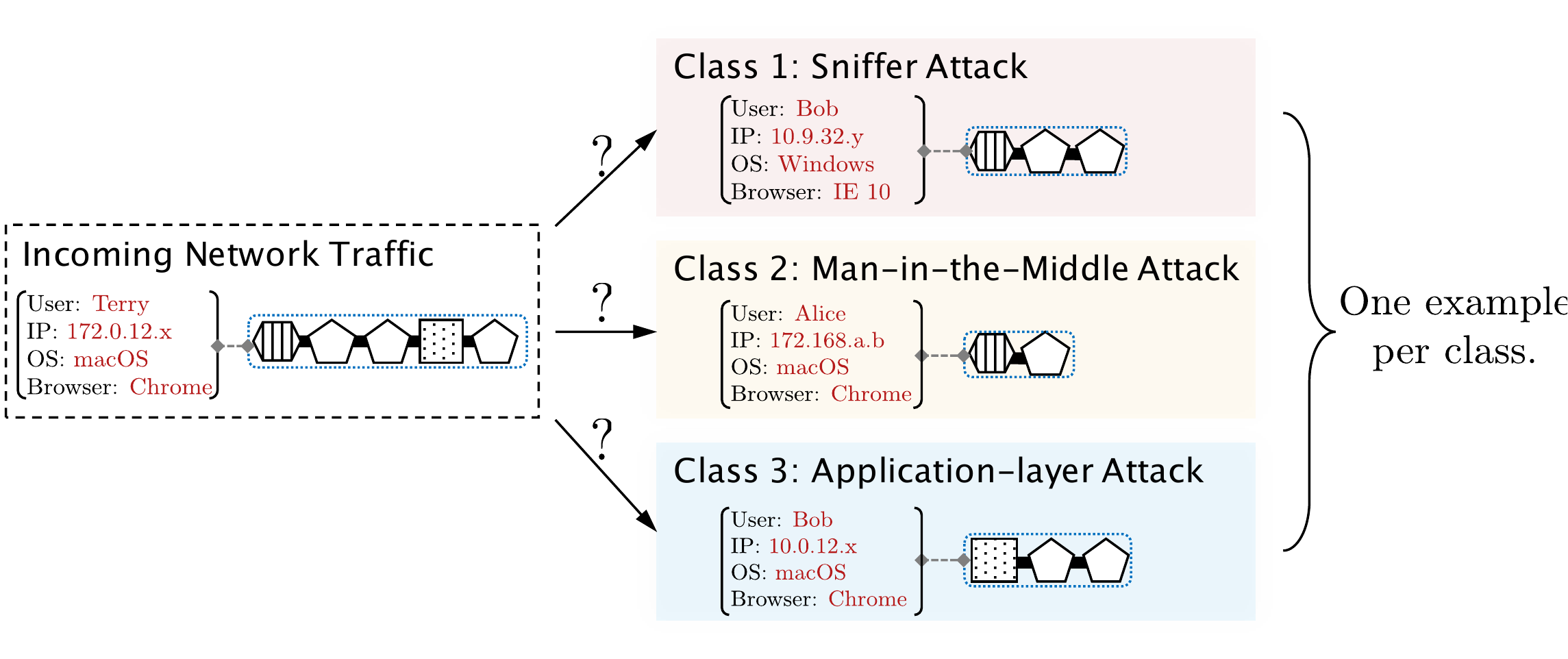}
       \caption[Motivating example: attack detection in one-shot]{Network attack detection using one-shot learning on attributed sequences. Each instance is composed of a user profile as the \textit{attributes} and a \textit{sequence} of user actions (depicted using different shapes). A system administrator is interested in finding out if the incoming network traffic is malicious with \textbf{only one} sample per class.}
    \label{fig-attributed-sequence}
\end{figure}
    \item \textbf{Bot Traffic Detection.}  With the rapid advance in e-commerce, more businesses than ever in history are using websites to advertise their products and services. Attributed sequences exist naturally on these websites: the activities of each visitor are recorded in log files as \textit{sequences} alongside the visitor's profile as \textit{attributes}. Aside from the \textit{real traffic} from potential customers on the website, there is another type of traffic, namely, the \textit{bot traffic}. Bot traffic~\cite{wikibottraffic} is from a series of applications that run scripts over the websites for various task. Many of those tasks are malicious to the websites, such as account hijacking with brute force, stealing web contents, undercutting prices and probing for potential attack opportunities~\cite{badbot2017}. Since the differences in attributes and sequences between bot and real traffic may not be significant, it is difficult to distill bot traffic from real traffic using only either attribute data or only sequence data. For instance, a real customer may have a similar or even identical profile (\eg, ``\texttt{OS}'', ``\texttt{IP}'') as the bot scripts; the tasks (\eg, \texttt{search information}, \texttt{click menus}) conducted by  real customers and bot scripts may also be similar. \\ Thus, a bot traffic system should be able to (1) Utilize attributed sequences in the log files to infer the different patterns of real traffic and bot traffic and (2) Generalize the patterns of both types of traffic to prevent future bot traffic. 

    \item \textbf{Network Intrusion Detection.} Network traffic can be modeled as attributed sequences. Namely, it consists of a sequence of packages being sent or received by the routers and a set of attributes indicating the context of the network traffic (\eg, user privileges, security settings, \etc). To respond in a timely fashion to potential network intrusion threats, one first has to determine what the intrusion type of incoming potentially malicious traffic even if only one or a few examples per known intrusion type have been seen previously (as depicted in Figure~\ref{fig-attributed-sequence}). 
\end{enumerate}
\section{State-of-the-Art}
\label{ch1-section-sota}
\subsection{Attributed Sequence Embedding} 
\begin{figure}[t]
    \centering
    \begin{subfigure}{0.45\linewidth}
        \begin{subfigure}{1.1\linewidth}
          \centering
          \includegraphics[width=1\textwidth]{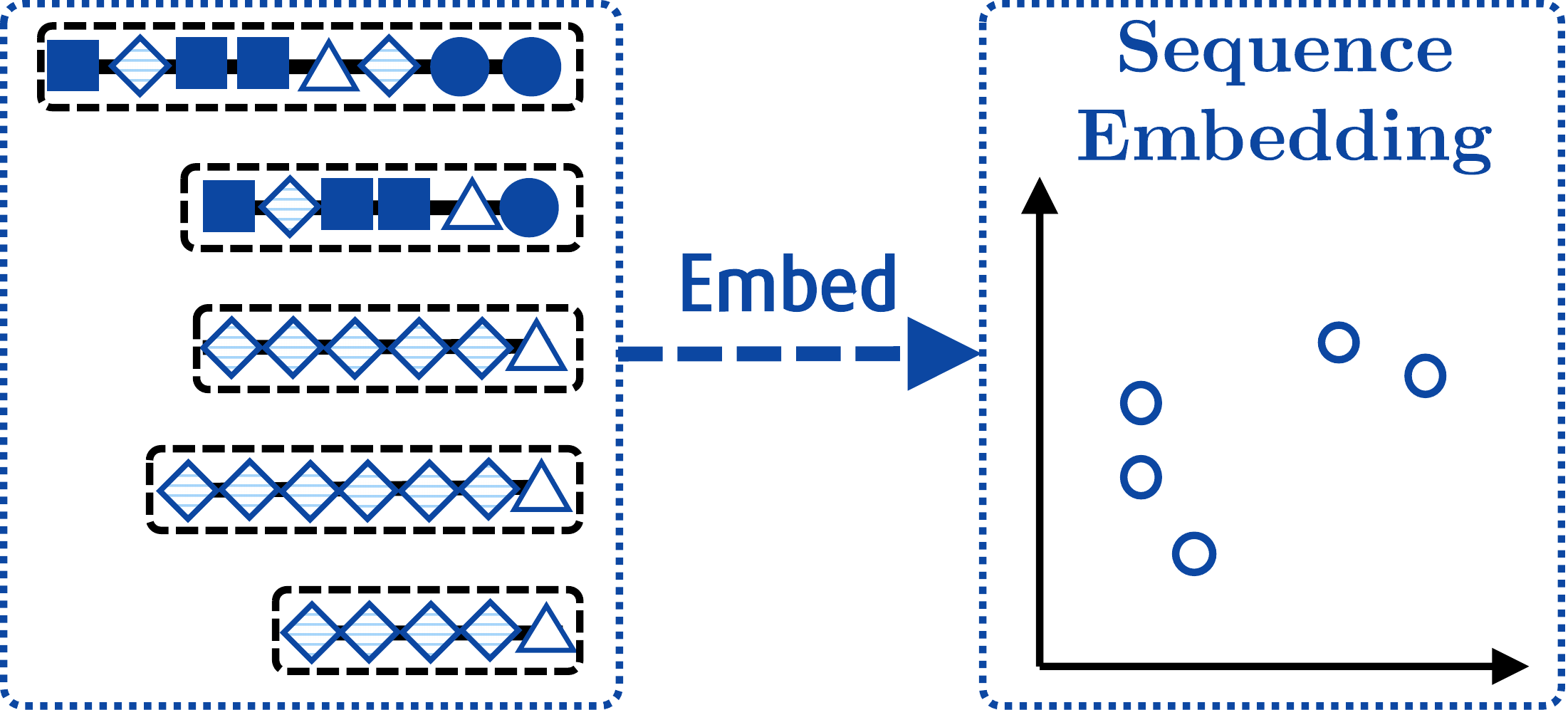}
          \caption{$\!\!$Sequence embedding~\cite{sutskever2014sequence}.}
          \label{fig-sota-seqonly}
         \vspace{5pt}
        \end{subfigure}
        \begin{subfigure}{1\linewidth}
          \centering
          \includegraphics[width=0.9\textwidth]{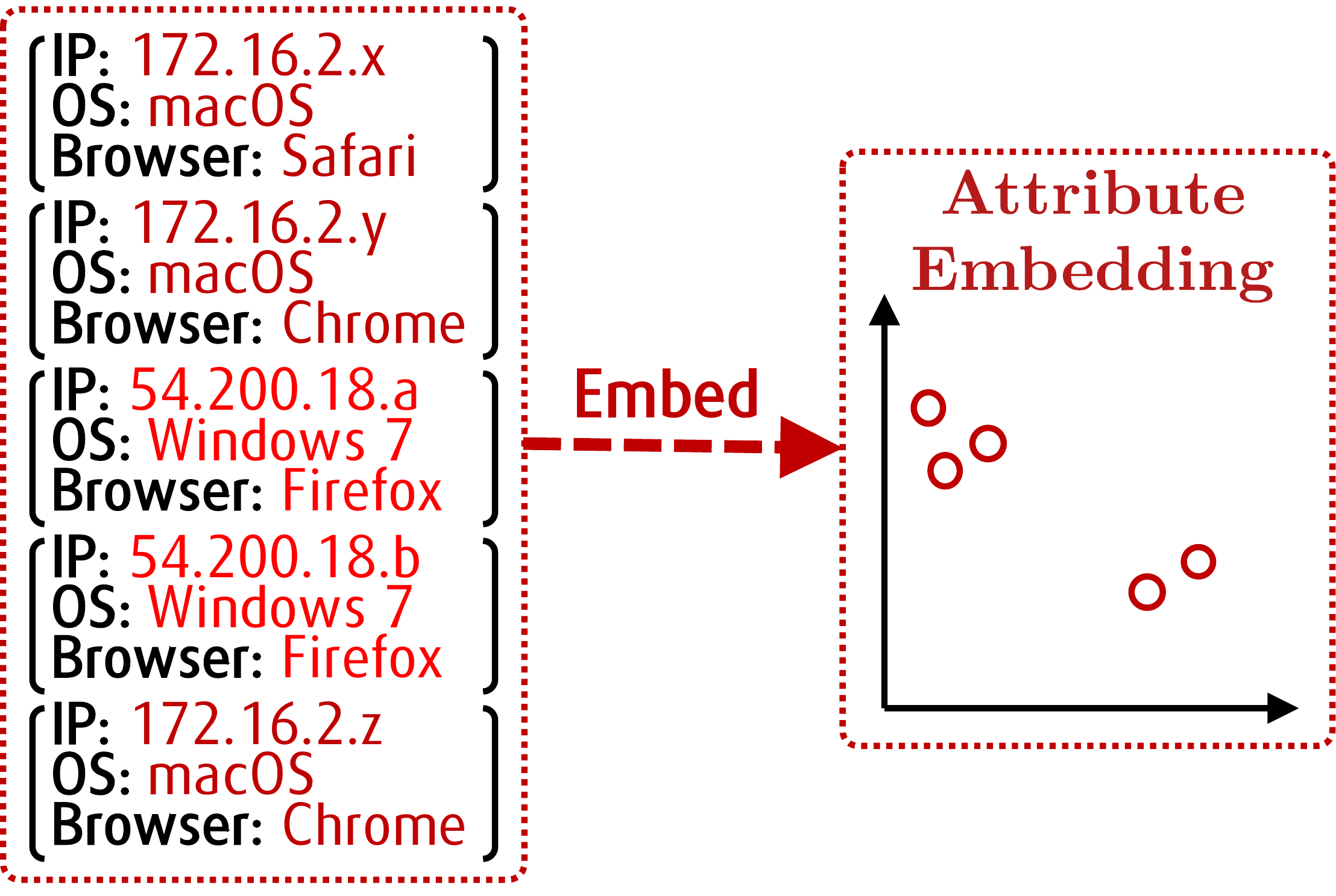}
          \caption{$\!\!\!$Attribute embedding\cite{wang2014generalized}.}
          \label{fig-sota-attr}
          \vspace{5pt}
        \end{subfigure}
        \begin{subfigure}{1\linewidth}
          \centering
          \includegraphics[width=0.95\textwidth]{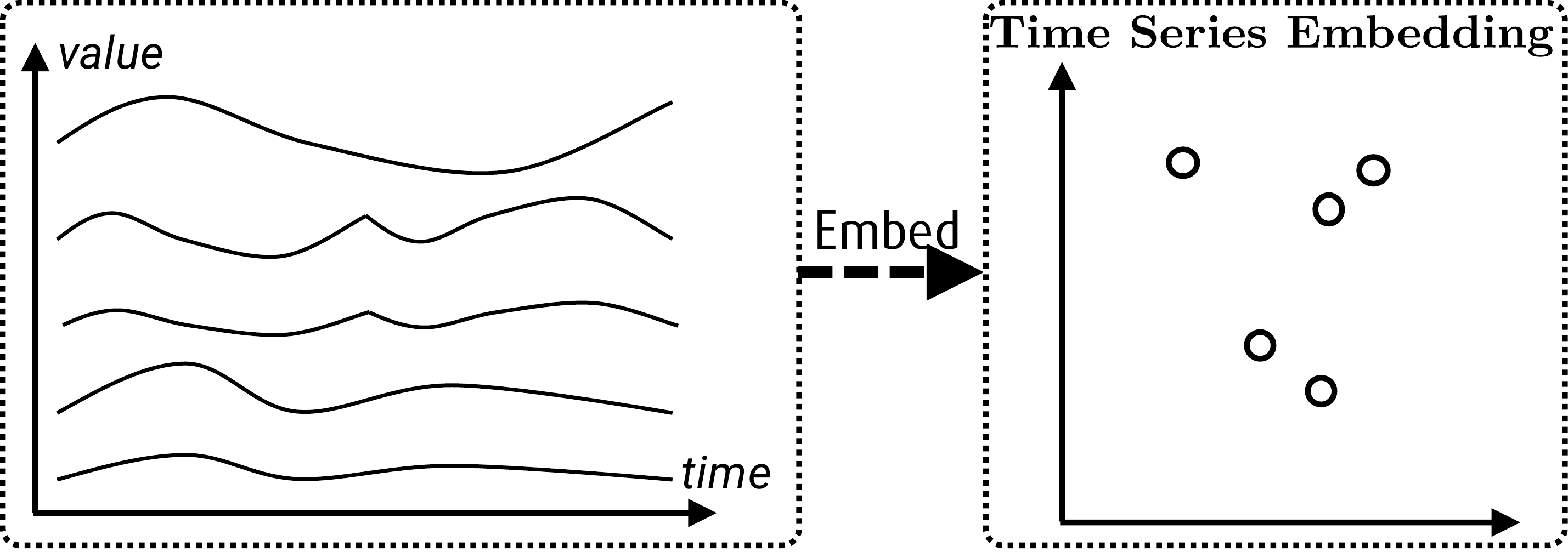}
          \caption{$\!\!$Time series embedding~\cite{khaleghi2016consistent}. }
          \label{fig-sota-ts}
        \end{subfigure}
    \end{subfigure}
    \hspace{5mm}
    \begin{subfigure}{0.40\linewidth}
    \centering
        \includegraphics[width=1\textwidth]{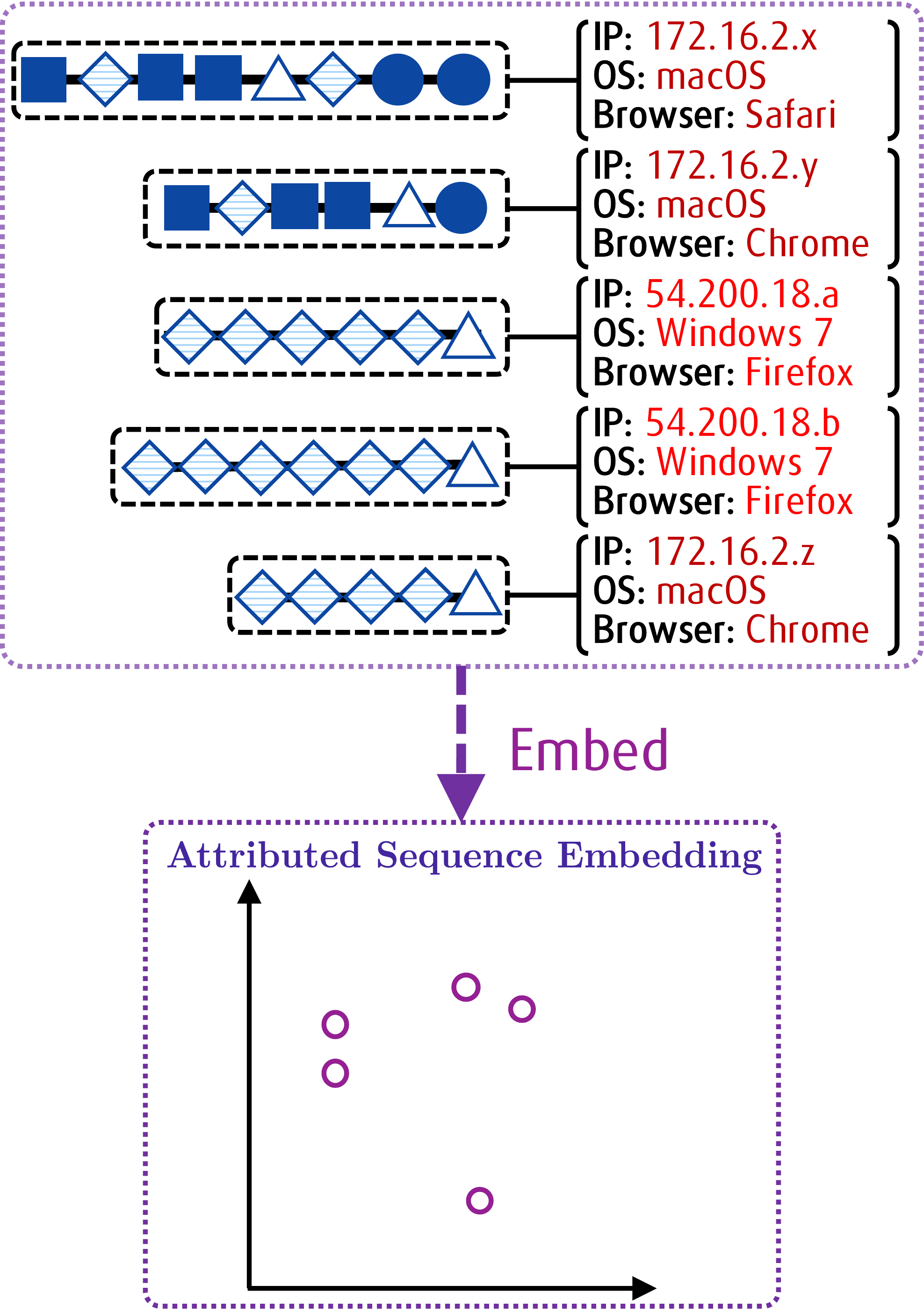}
        \caption{Attributed sequence embedding. }
        \label{fig-sota-nas}
    \end{subfigure}
    \caption{Comparison of different embedding problems. }
    \vspace{-15pt}
    \label{fig-as-setting}
\end{figure}
Sequential data usually requires a careful design of its feature representation before being fed to a learning algorithm. One of the feature learning problems on sequential data is called sequence embedding~\cite{cho2014learning, sutskever2014sequence}, where the goal is to transform a sequence into a fixed-length feature representation. Similarly, the attributed sequence embedding problem corresponds to transforming an attributed sequence into a fixed-length feature representation with continuous values.

In the sequence context, conventional methods only focus on sequential data~\cite{sutskever2014sequence, luong2015multi, cho2014learning, kalchbrenner2013recurrent, mueller2016siamese} to learn the dependencies between items within variable-length sequences -- neither support the attribute data nor learn the dependencies between attributes and sequences. 

On the other hand, the recent applications on image datasets using multilayer deep neural networks~\cite{wang2014generalized, akata2013label, cvpr-face-verify, sun2014deep} focus on the problems of pattern recognization in fixed-size images -- they neither support the variable-length sequences nor learn the attribute-sequence dependencies. Here in Figure~\ref{fig-as-setting}, we demonstrate the difference between our attributed sequence embedding problem and other embedding problems in the state-of-the-art. 

\subsection{Deep Metric Learning on Attributed Sequences} 
\begin{figure}[t]
    \centering
       \includegraphics[width=0.8\linewidth]{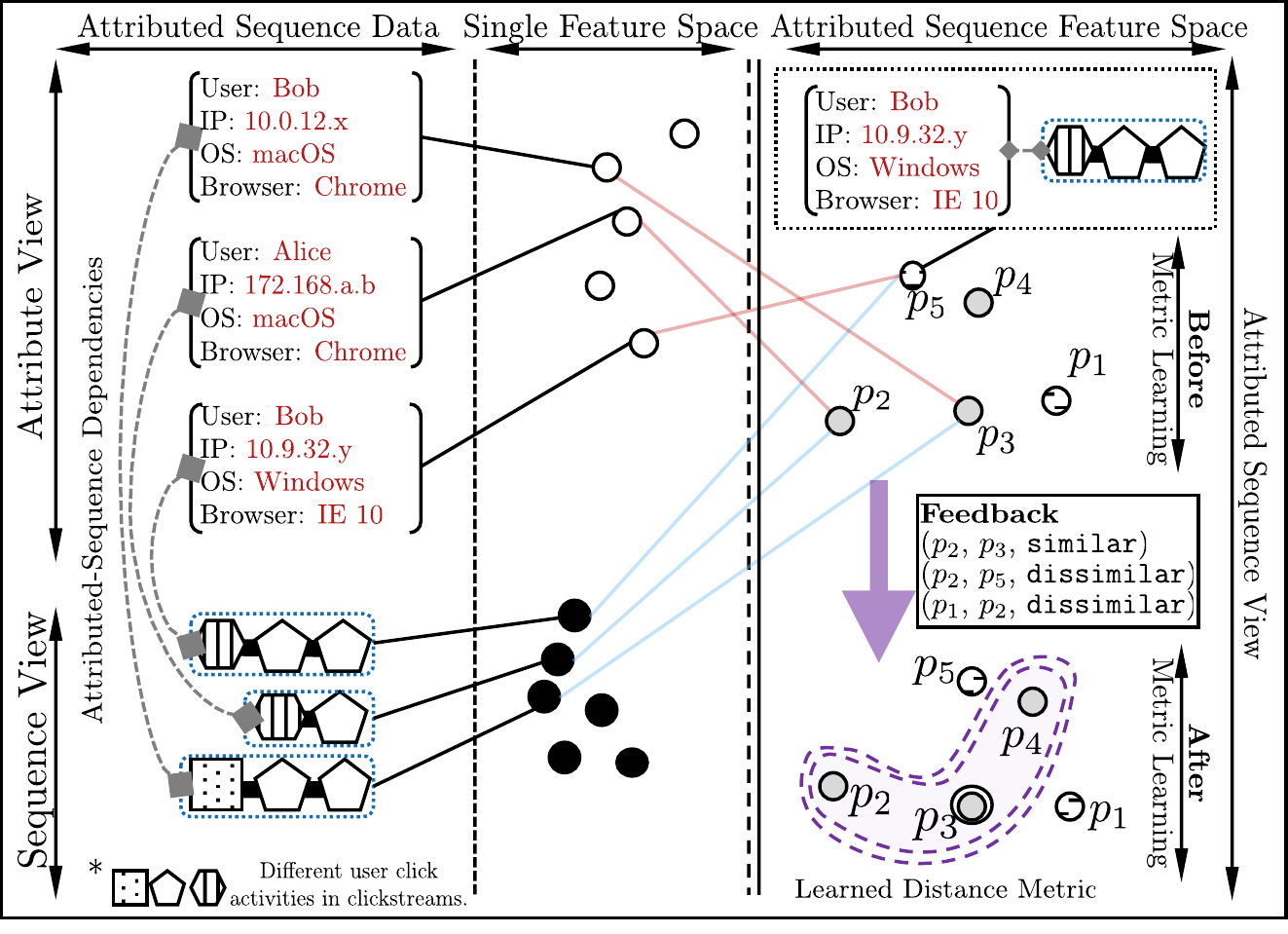}
    \caption{Distance metric learning on attributed sequences. }
    \label{fig-feedback}
    \vspace{-6mm}
\end{figure}
Conventional approaches on distance metric learning~\cite{xing2003distance, davis2007information, koestinger2012large, mignon2012pcca} mainly focus on learning a Mahalanobis distance metric, which is equivalent to learning a \textit{linear transformation} on data attributes. 
Recent research has extended distance metric learning to nonlinear settings~\cite{NIPS2012_4840, cvpr-face-verify}, where a nonlinear mapping function is first learned to project the instances into a new space, and then the final metric becomes the Euclidean distance metric in that space. 

Deep metric learning has been the method of choice in practice for learning nonlinear mappings~\cite{wang2012parametric, chatpatanasiri2010new, NIPS2012_4840, cvpr-face-verify}. 
Recent research on metric learning has explored sequential data~\cite{mueller2016siamese}, where we have structural information in the sequences, but no attributes are available. We use Figure~\ref{fig-feedback} to depict the differences in distance metrics learned from the data. 
\begin{figure}[!ht]
    \centering
    \begin{subfigure}[t]{.32\linewidth}
        \centering
        \includegraphics[width=\textwidth]{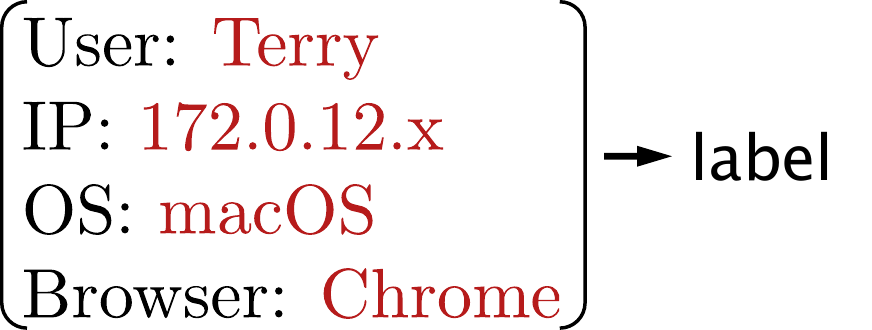}
        \caption{Classification on \\ data attributes~\cite{akata2013label}. }
        \label{fig-classify-att}
    \end{subfigure} 
    \hspace{1mm}
    \begin{subfigure}[t]{0.40\linewidth}
        \centering
        \includegraphics[width=\textwidth]{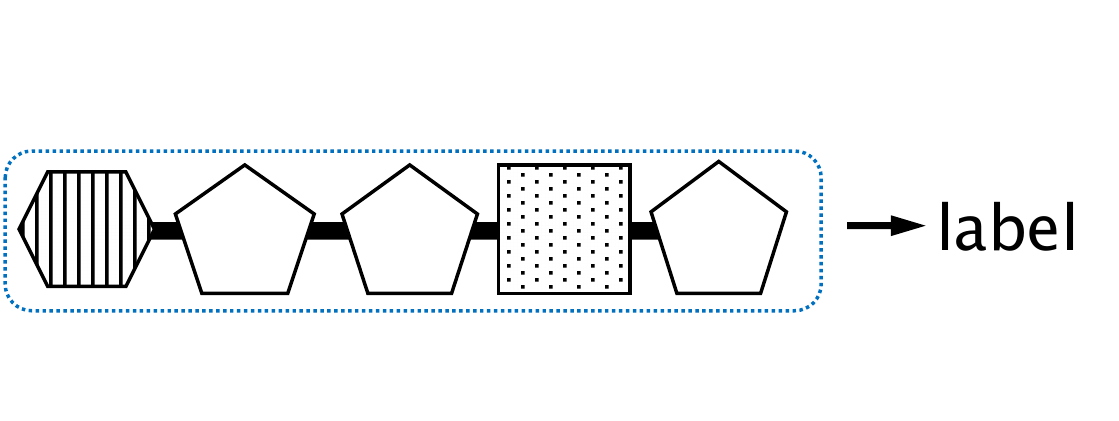}
        \caption{Sequence classification~\cite{xing2010brief}. }
        \label{fig-classify-att}
    \end{subfigure}
    \vspace{5mm}

    \begin{subfigure}[t]{0.28\linewidth}
        \centering
        \includegraphics[width=\textwidth]{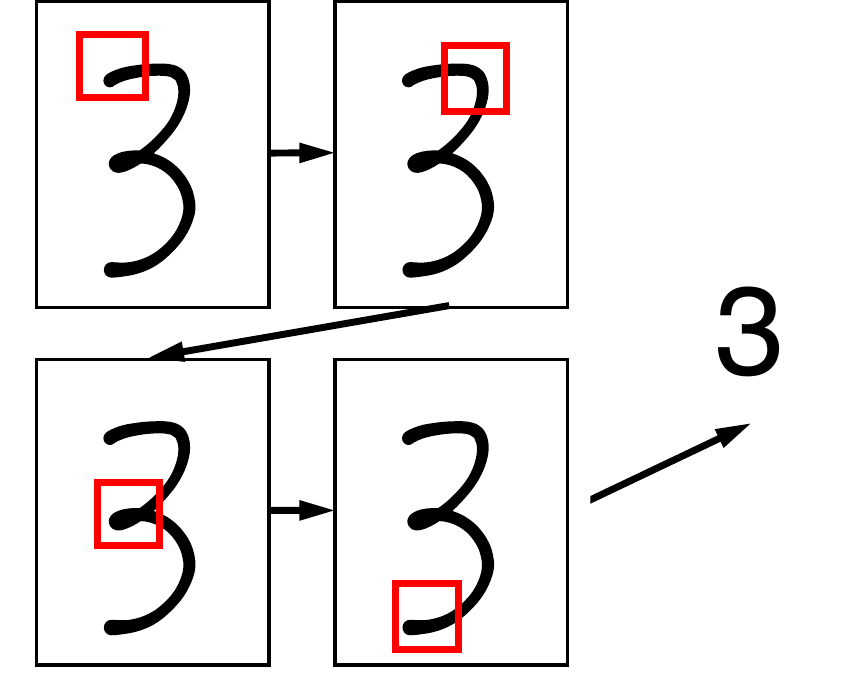}
        \caption{Image classification with attention~\cite{mnih2014recurrent}. }
        \label{fig-classify-att}
    \end{subfigure}
    \begin{subfigure}[t]{0.28\linewidth}
        \centering
        \includegraphics[width=\textwidth]{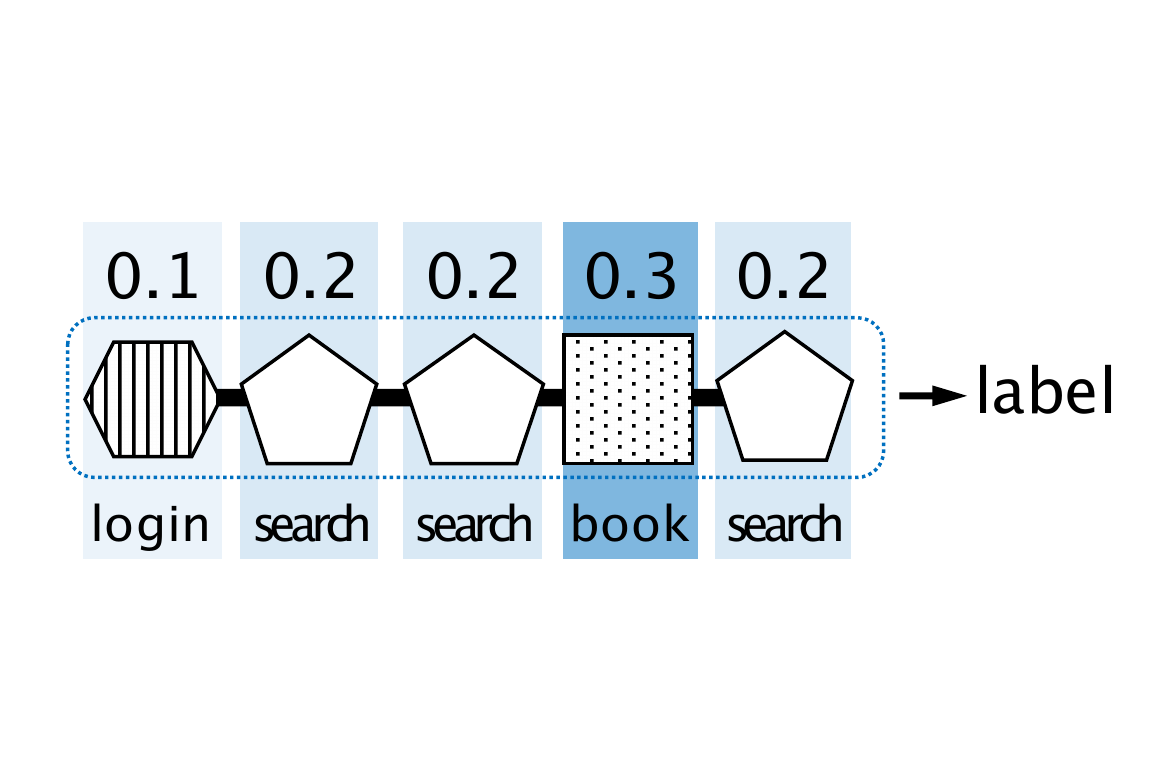}
        \caption{Sequence classification with attention~\cite{wang2016recurrent}. }
        \label{fig-classify-att}
    \end{subfigure}

    \vspace{5mm}
    \begin{subfigure}[t]{.65\linewidth}
        \centering
        \includegraphics[width=\textwidth]{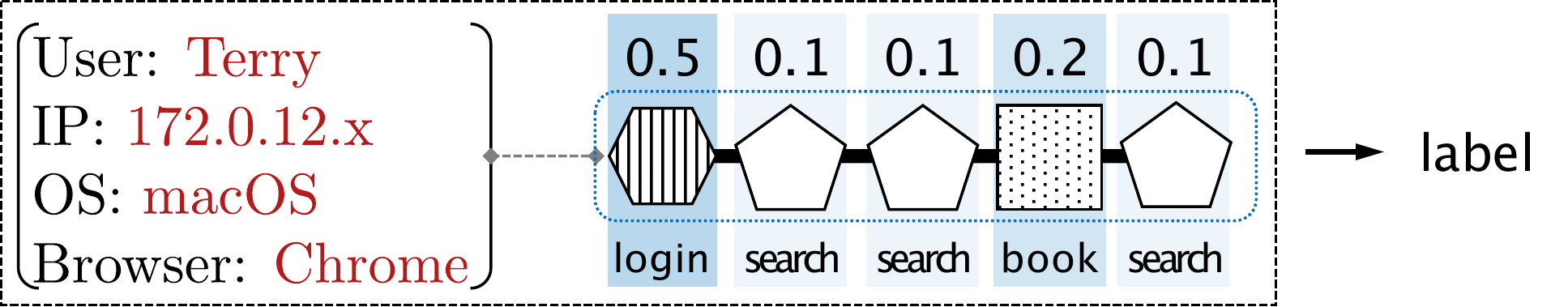}
        \caption{Sequence classification with attribute-guided attention.}
        \label{fig-classify-att}
    \end{subfigure}
    \vspace{-3mm}
    \caption{A comparison of related classification problems. }
    \label{fig-classify-setting}
    \vspace{-6mm}
\end{figure}
\subsection{One-shot Learning on Attributed Sequences} With the capacity of deep learning for generalizing the training examples, one-shot learning has attracted more interests recently when only one training sample is available~\cite{santoro2016meta, bertinetto2016learning, vinyals2016matching, koch2015siamese}. Conventional approaches to one-shot learning focus on using feature vectors as input in the learning process~\cite{koch2015siamese,wang2012parametric,xing2010brief}, in which each instance is represented as a fixed-size vector (\eg, images). However, neither sequential data nor attributed sequences has been studied in one-shot learning research. 

\subsection{Attention for Attributed Sequence Classification} In the areas of image processing, researchers designed a mechanism of only letting the model learn from certain ``useful'' and ``informative'' regions of image inputs~\cite{mnih2014recurrent}. This mechanism is called the attention network. Attention network has gained more research interest in recent years and has been generalized in various domains, \eg, image captioning~\cite{xu2015show, nam2016dual} and generation~\cite{gregor2015draw}, speech recognization~\cite{chorowski2015attention}, document classification~\cite{yang2016hierarchical}. However, attention model has yet been studied for attributed sequence classification. We use Figure~\ref{fig-classify-setting} to depict different classification problems on attributed sequences. 

\section{Research Challenges}
\label{ch1-section-research-challenge}
Here we summarize the research challenges of using attributed sequences in data mining applications:
\begin{figure}[!ht]
    \centering
    \includegraphics[width=0.8\linewidth]{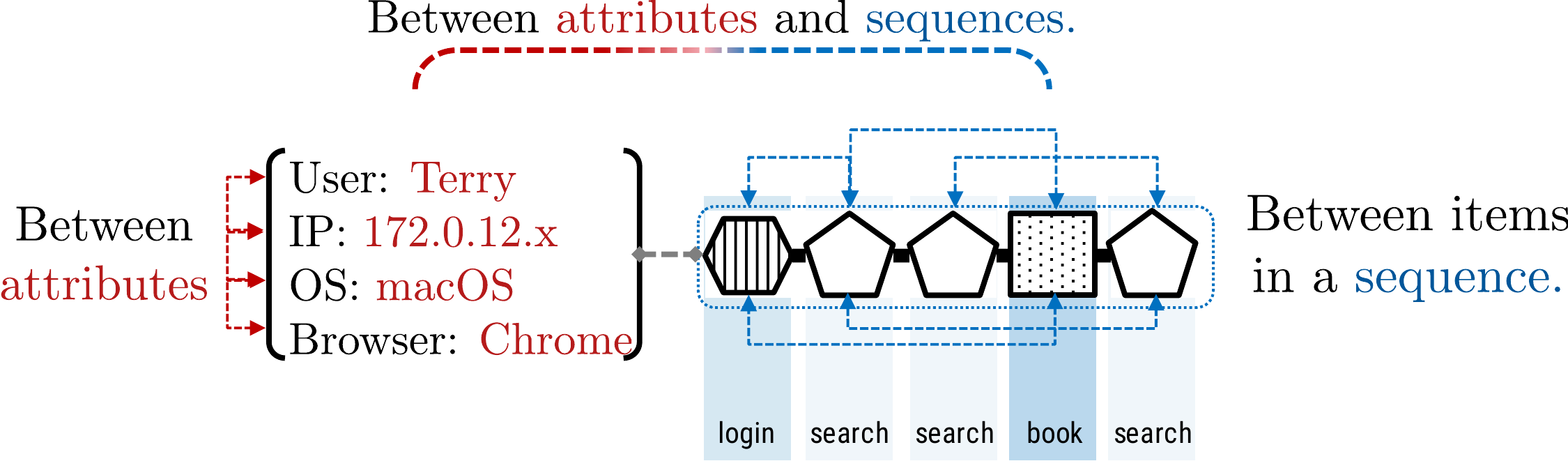}
    \caption[Three Types of Dependencies in An Attributed Sequence.]{The three types of dependencies in an attributed sequence: {\color{blue} item} dependencies, {\color{red} attribute} dependencies and {\color{red}attribute}-{\color{blue}sequence} dependencies.}
    \label{fig-complexrel}
\end{figure}


\subsection{Heterogeneous dependencies in attributed sequences} The bipartite structure of attributed sequences poses unique challenges in feature learning. Contrary to a straightforward thinking that attributes and sequences could first be learned separately then later rejoined, various dependencies arise in attributed sequences. There exist three types of dependencies in an attributed sequence: item dependencies, attribute dependencies and attribute-sequence dependencies. Thus, learning and capturing these \emph{attribute-sequence dependencies} are critical for attributed sequence embedding. For example, in a web search, each user session is composed of a session profile as attributes (\eg, \texttt{Device Type}, \texttt{OS}, \etc) as well as a sequence of search \texttt{keywords}. One \texttt{keyword} may depend on previous search terms (\eg, \texttt{temperature} following \texttt{snow storm}) and the keywords searched may depend on the device type (\eg, \texttt{Nearest restaurant} on \texttt{Cellphone}). In the user behavior studies using web search histories, it would be impractical without considering such dependencies. We illustrate these three types of dependencies in Figure~\ref{fig-complexrel}. 

\subsection{Lack of labeled data} With the continuously incoming volume of data and the high labor cost of manually labeling data, it is rare to find attributed sequences from many real-world applications with labels ({\eg}, \textit{fraud}, \textit{normal}) attached. Without proper labels, it is challenging to learn an embedding function that is capable of transforming attributed sequences into compact feature representations while capturing the three types of dependencies. 

\subsection{Metric Learning on Attributed Sequences} Metric learning focuses on the problems of differentiating the instances based on the similarities/dissimilarities in user feedback. One fundamental problem in metric learning on attributed sequences lies in the sequential structure in each instance. Conventional approaches to distance metrics learning assume, explicitly or implicitly, that the data are represented as features vectors (\ie, attributes)~\cite{xing2003distance, davis2007information, koestinger2012large, mignon2012pcca}. However, in attributed sequence data, the sequence in each data record is not represented as a feature vector, but rather as a variable-length sequence of discrete items. The information is encoded in the ordering of the items. A distance metric on attributed sequences thus must be able to capture structural similarities and dissimilarities between sequences.  

\subsection{Generalize from Only One Sample} This problem is different from previous one-shot learning work, as we now need to extract feature vectors from not only the \textit{attributes} but also the structural information from the \textit{sequences} and the \textit{dependencies} between attributes and sequences. The key difficulty in one-shot learning is to generalize beyond the single training example. It is more difficult to generalize from a more complex data type~\cite{chen2014big}, such as attributed sequence data, than from a simpler data type due to the larger search space and slower convergence. 

\subsection{Attention in Attributes and Sequences} Recent sequence learning research~\cite{yang2016hierarchical} has used neural attention models to improve the performance of sequence learning, such as document classification~\cite{yang2016hierarchical}. However, the attention mechanism focuses on learning the weight of certain time steps or sub-sequences in each sequential instance, without regards to its associated attributes. With information from the attributes, the weight of item or sub-sequence may be drastically different from the weight calculated by the attention mechanism using only sequences, which would consequently have different classification results. 

\section{Proposed Solutions}
\label{ch1-section-proposed-tasks}

In this dissertation, we address the challenges listed in Section~\ref{ch1-section-research-challenge} by extending the state-of-the-art deep learning models under various problem settings. 

\subsection{Attributed Sequence Embedding} The goal of the first task is to generate fix-sized feature representations (\ie, embeddings) of attributed sequences that can be used in downstream mining tasks (\eg, clustering and outlier detection). Many of these machine learning tasks use vectors of real numbers as the inputs. The embeddings, as mappings from attributed sequences to fixed-size vectors of real numbers, are both important and valuable in these tasks. Because the embeddings are the vectors mapped from inputs, the similarities measured by distance functions in the embedding space can be viewed as the similarities between the original inputs. 
This task offers the following contributions:
\begin{itemize}
    \item 
 We study the problem of attributed sequence embedding, which corresponds to a natural generalization of many real-world applications. We formally define the data model of attributed sequences, which consists of a sequence of categorical items and a set of static attributes. 
    \item 
We propose an innovative framework to exploit the dependencies among the attributed sequences. The \nas establishes the first attributed sequence embedding framework, which employs a three-phase process to effectively produce feature representations for attributed sequences. 
    \item We evaluate the \nas framework on various real-world datasets competing with state-of-the-art methods. We study the performance of \nas using clustering and outlier detection tasks. We also show that \nas is capable of producing feature representations in real-time. We also evaluate \nas using real-world case studies of user behaviors.
\end{itemize}

    \subsection{Incorporating Feedback of Attributed Sequences} Different from Task 1, now there is a limited amount of labeled feedback from domain experts in some real-world applications, such as \textit{fraud detection} and \textit{user behavior analysis}. The feedback is usually given as pairwise examples, where each feedback instance is composed of two attributed sequences and a label depicting whether they are similar or dissimilar. The feedback from domain experts is important in these applications since the feedback provides human insights of the datasets. For example, a domain expert can give feedback indicating a new online transaction is similar to a known fraud transaction. To address the above challenges, we propose a deep learning framework, called \mlas (\underline{M}etric \underline{L}earning on \underline{A}ttributed \underline{S}equences), to learn a distance metric that can effectively measure the dissimilarity between attributed sequences. 
Our \mlas framework includes three sub-networks: \anet (an attribute network to encode the attribute information using nonlinear transformations), \snet (a sequence network to encode structural information using LSTM), and \mnet (a metric network to produce the distance metric). 
We further designed three \mlas models: balanced, \anetns-centric and \snetns-centric.
    We offer the following main contributions in this task: 
\begin{itemize}
    \item We are the first to formulate and then study the problem of deep metric learning of attributed sequences. 
    \item We design three sub-networks: the \anet to encode the attribute information, \snet to encode the structural information, and a metric network \mnet to produce the distance metric. Together with these sub-networks, \mlas effectively learns the nonlinear distance metric on attributed sequences. 
    \item Our empirical results on real-world datasets demonstrate that our proposed \mlas framework significantly improves the performance of metrics learning on attributed sequences. 
\end{itemize}
    \subsection{Classify Attributed Sequences in One-shot} 
    To address the challenges in Section~\ref{}, we propose an end-to-end one-shot learning model, called \olasns, to accomplish one-shot learning for attributed sequences. The \olas model includes two main components: a \cnet to encode the information from attributes, sequences and their dependencies and a \vnet to learn the similarities and differences between different attributed sequence classes. The proposed \olas model is beyond a simple concatenation of \cnet and \vnetns. Instead, they are interconnected within one network architecture and thus can be trained synchronously. Once the \olas is trained, we can then use it to make predictions for not only the new data but also for entire previously unseen new classes. In this task, we offer the following core contributions: 
\begin{itemize}
    \item We formulate and analyze the problem of one-shot learning for attributed sequence classification. 
    \item We develop a deep learning model that is capable of inferring class labels for attributed sequences based on one instance per class. 
    \item We demonstrate that the \olas network model trained on attributed sequences significantly improves the accuracy of label prediction compared to state-of-the-art.
\end{itemize}

    \subsection{Attention Model for Attributed Sequence Classification} When presented an image, humans are capable of only focusing on some regions of the image and grasping the information that is deemed useful. The attention models in deep learning are designed to fulfill similar goals. Recently, the attention models have generated a lot of research interests in various areas~\cite{mnih2014recurrent, gregor2015draw, chorowski2015attention, yang2016hierarchical}. For instance, \textit{attention} has been employed to improve the performance of natural language processing tasks (\eg, document classification). Compared to treating every word and sentence equally in the document classification tasks without attention, the capability of enhanced processing of certain words or sentences using attention has demonstrated improvements in such tasks~\cite{yang2016hierarchical}. In this task, I design two attention models for attributed sequence classification. Specifically, this task offers the following contributions: 
\begin{itemize}
    \item We formulate the problem of attributed sequence classification. 
    \item We design a deep learning framework, called \amasns, with two attention-based models to exploit the information from attributes and sequences. 
    \item We demonstrate that the proposed models significantly improve the performance of attributed sequence classification using performance experiments and case studies. 
\end{itemize}

\section{Road Map}
\label{ch1-section-roadmap}
This rest of the dissertation is organized as follows. We introduce the attributed sequence data model in Chapter~\ref{chapter-background}. We start from the first task of attributed sequence embedding in Chapter~\ref{chapter-task1}. Chapter~\ref{chapter-task2} focuses on the task of deep metric learning on attributed sequences. In Chapters~\ref{chapter-task3} and~\ref{chapter-task4}, we present the tasks of one-shot learning and the attention network for attributed sequence classification tasks, respectively. Related work is discussed in Chapter~\ref{chapter-related}. We conclude the findings and discuss future works in Chapter~\ref{chapter-conclusion-future}.

 \newpage
\chapter{Background: Attributed Sequence}
\label{chapter-background}
\newpage

\begin{definition}[\small Sequence.]{\rm
\label{plain-sequence}
Given a set of $r$ categorical items $\mathcal{I} = \{e_1,\!\cdots, e_r\}$, the $k$-th sequence in the dataset $S_k = \big( \alpha_k^{(1)}, \cdots, \alpha_k^{(l_k)} \big)$ is an ordered list of items, where $\alpha_k^{(t)} \in \mathcal{I}, \forall t = 1,\cdots, l_k$.}
\end{definition}
Different sequences can have a varying number of items. For example, the number of user click activities varies between different user sessions. 
The meanings of items are different in different datasets. 
For example, in user behavior analysis from clickstreams, each item represents one action in user's click stream ({\eg}, $\mathcal{I}=\{$\texttt{search}, \texttt{select}\}, where $r=2$). Similarly in DNA sequencing, each item represents one canonical base ({\eg}, $\mathcal{I}\!=\!\{\texttt{A},\! \texttt{T},\! \texttt{G},\! \texttt{C}\}$, where $r=4$). There are dependencies between items in a sequence. Without loss of generality, we use the one-hot encoding of $S_k$, denoted as $ \mathbf{S}_k = \big(\vec{\alpha}_k^{(1)}, \cdots, \vec{\alpha}_k^{(l_k)}\big) \in \mathbb{R}^{l_k \times r} $
where each item $\vec{\alpha}_k^{(t)}\in \mathbb{R}^r$ in $\mathbf{S}_k$ is a one-hot vector corresponding to the original item $\alpha_k^{(t)}$ in the sequence $S_k$. 

In prior work~\cite{graves2013generating}, one common preprocessing step is to zero-pad each sequence to the longest sequence in the dataset and then to one-hot encode it. Without loss of generality, we denote the length of the longest sequence as $T$. 
For each $k$-th sequence $S_k$ in the dataset, we denote its equivalent one-hot encoded sequence as 
$\mathbf{S}_k = \left( \vec{\alpha}_k^{(1)}, \cdots, \vec{\alpha}_k^{(T_k)} \right)$, where $\vec{\alpha}_k^{(t)}\in \mathbb{R}^r$ corresponds to a vector represents the item $\alpha_k^{(t)}$ with the $l$-th entry in $\vec{\alpha}_k^{(t)}$ is ``1'' and all other entries are zeros if $\alpha_k^{(t)}=e_l, e_l \in \mathcal{I}$.

\begin{definition}[Attributes.]{\rm
	\label{def-attributes}
	The attribute values are concatenated into a vector $\mathbf{x}_k \in \mathbb{R}^{u}$, where $u$ is the number of attributes in $\mathbf{x}_k$. }
\end{definition}
The value of each attribute can be either categorical or numerical. $u$ is considered as a constant within any given datasets. 
For example, in a dataset where each instance has two attributes ``\texttt{IP}'' and ``\texttt{OS}'', $u=2$. 

\begin{definition}[\small Attributed Sequence.]{\rm
\label{attributed-sequence}
    Given a vector of attribute values $\mathbf{x}_k$ and a sequence $\mathbf{S}_k$, an attributed sequence $J_k\! =\! (\mathbf{x}_k,\! \mathbf{S}_k)$ is an ordered pair of the attribute value vector $\mathbf{x}_k$ and the sequence $\mathbf{S}_k$.}
\end{definition}
Common practices of distance metric learning involve pairwise examples as the training dataset~\cite{wang2015survey, wang2011integrating, yeung2007kernel, koestinger2012large, cvpr-face-verify, mueller2016siamese}. We thus define the feedback as a collection of similar (or dissimilar) attributed sequence pairs. 
\begin{definition}[Similar and Dissimilar Feedback]{\rm 
\label{def-as-feedback}
Let $\mathcal{J} = \{ J_1, \cdots, J_n\}$ be a set of $n$ attributed sequences. A feedback is a triplet $(p_i, p_j, \ell_{ij})$ consisting of two attributed sequences $p_i, p_j \in \mathcal{J}$ and a label $\ell_{ij} \in \{0, 1\}$ indicating whether $p_i$ and $p_j$ are similar ($\ell_{ij}=0$) or dissimilar ($\ell_{ij}=1$). We define a similar feedback set $\mathcal{S} = \{(p_i, p_j, \ell_{ij}) | \ell_{ij}=0\}$ and a dissimilar feedback set $\mathcal{D} = \{(p_i, p_j, \ell_{ij}) | \ell_{ij}=1\}$.}
\end{definition}

 \newpage
\chapter{Unsupervised Attributed Sequence Embedding}
\label{chapter-task1}
\newpage

\section{Problem Definition}
\label{section-task1-problem-definition}
Using the previously defined notations in Chapter~\ref{chapter-background}, we formulate our attributed sequence embedding problem as follows. 
\begin{definition}[Attributed Sequence Embedding.]{\rm
\label{def-main}
Given a dataset of attributed sequences $\mathcal{J}=\{J_1, \cdots, J_n\}$, the problem of attributed sequence embedding is to find a function $\Phi$ parameterized by $\theta$ that produces feature representations for $J_k$ in the form of vectors. The problem is formulated as
\begin{equation}
\label{equation-main}
    {\minimize}\sum_{k=1}^{n} 
    \sum_{t=1}^{l_k}-\log\Pr_{\phi} \left(\vec{\alpha}_k^{(t)} | \vec{\beta}_k^{(t)}, \mathbf{x}_k\right)
\end{equation}
\noindent where $\vec{\beta}_k^{(t)}\! =\! \big ( \vec{\alpha}_k^{(t-1)},\! \cdots,\! \vec{\alpha}_k^{(1)}\big ), \forall t = 2, \cdots, l_k$ represents the items prior to $\vec{\alpha}_k^{(t)}$ in the sequence. }\end{definition}
Our problem can be interpreted as: we want to minimize the prediction error of the $\vec{\alpha}_k^{(t)}$ in each attributed sequence given the attribute values $\mathbf{x}_k$ and all the items prior to $\vec{\alpha}_k^{(t)}$. 

\section{The \nas Framework}
\begin{figure}[t]
  \centering
  \includegraphics[width=0.9\linewidth]{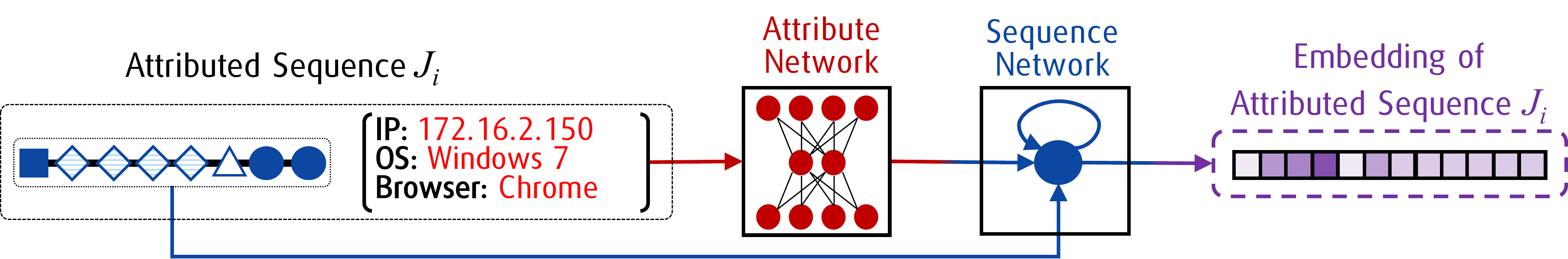}
      \vspace{-2mm}
  \caption[Process Flow of \nasns]{The process flow of the \nas framework from attributed sequences to the final feature representations is displayed. For an attributed sequence, the attribute feature representation from the attribute network is shared with the sequence network. In the end, the hidden layer states of the sequence network are saved as the feature representation for this attributed sequence. }
  \label{fig-train-vec}
\end{figure}

In this section, we introduce two main components in our proposed \nas framework, \ie, \textit{attribute network} and \textit{sequence network}. We illustrate how the two networks are integrated to serve the purpose of embedding attributed sequences in Figure~\ref{fig-train-vec}. 

In many real-world applications, attributes are often available before the sequences. For example, in the online ticketing system, the attributes that depict user profile are recorded when a user starts a ticket booking session. However, the sequence of one's actions on the webpage would not be available until such booking session ends. Our design follows the same paradigm, that is, the attributes are available before the respective sequence of items in each attributed sequence. In addition, the lengths of sequence in many real-world applications are short. For example, one may only need tens of actions to finish a ticket booking online. 

\subsection{Attribute Network}
\label{subsection-attribute-network}
Fully connected neural network~\cite{liou2014autoencoder} is capable of modeling the dependencies of the inputs and at the same time reduce the dimensionality. 
It has been widely used \cite{liou2014autoencoder, liou2008modeling, phan2016differential} for unsupervised data representations learning,  including tasks such as dimensionality reduction and generative data modeling. 
With the high-dimensional sparse input attribute values $\mathbf{x}_k \in \mathbb{R}^u$, it is ideal to use such a network to learn the attribute dependencies. We design our attribute network as 
\begin{equation}
    \label{autoencoder-formula}
    \begin{split}
        \mathbf{V}_k^{(1)} &= \rho\left(\mathbf{W}_A^{(1)}\mathbf{x}_k + \mathbf{b}_A^{(1)}\right) \\[-5mm]
        &\vdots \\[-5mm]
        \mathbf{V}_k^{(M)} &= \rho\left(\mathbf{W}_A^{(M)}\mathbf{V}_k^{(M-1)} + \mathbf{b}_A^{(M)}\right) \\[-2mm]
        \mathbf{V}_k^{(M+1)} &= \sigma\left(\mathbf{W}_A^{(M+1)}\mathbf{V}_k^{(M)} + \mathbf{b}_A^{(M+1)}\right) \\[-5mm]
        &\vdots \\[-5mm]
        \widehat{\mathbf{x}_{k}} &= \sigma\left(\mathbf{W}_A^{(2M)}\mathbf{V}_k^{(2M-1)} + \mathbf{b}_A^{(2M)}\right)
    \end{split}
\end{equation}
where $\rho$ is the encoder activation function and $\sigma$ is the decoder activation function. 
In this attribute network, we use \texttt{ReLU} activation function~\cite{nair2010rectified}, defined as 
\begin{equation}
    \rho(z) = \max(0,z)
\end{equation} 
and \texttt{sigmoid} activation function, defined as 
\begin{equation}
    \sigma(z) = \frac{1}{1 + e^{-z}}
\end{equation}
We denote the parameter set of attribute network as $\phi_A = \{\mathbf{W}_A, \mathbf{b}_A\}$, where $\mathbf{W}_A=\left(\mathbf{W}_A^{(1)},\cdots,\mathbf{W}_A^{(2M)}\right)$ and $\mathbf{b}_A=\left(\mathbf{b}_A^{(1)},\cdots,\mathbf{b}_A^{(2M)}\right)$.

The attribute network with $2M$ layers has two components, \ie, the encoder, and the decoder. The encoder is composed of the first $M$ layers, and the next $M$ layers work as the decoder. 
With $d_M$ hidden units in the $M$-th layer, the input attribute vector $\mathbf{x}_k \in \mathbb{R}^u$ is first transformed to $\mathbf{V}_k^{(M)}\in \mathbb{R}^{d_M}, d_M \ll u$ by the encoder. Then the decoder attempts to reconstruct the input and produce the reconstruction result $\widehat{\mathbf{x}_k}\in \mathbb{R}^{u}$. 
An ideal attribute network should be able to reconstruct the input from the $\mathbf{V}_k^{(M)}$. 
The smallest attribute network is built with $M=1$, where there are one encoder and one decoder.

\subsection{Sequence Network}
\label{subsection-sequence-network}

The proposed sequence network is a variation of the long short-term memory model (LSTM). The sequence network takes advantage of LSTM to learn the dependencies between items in sequences. Sequence network also accepts the feature representations from the attribute network to affect the learning process and thus learn the attribute-sequence dependencies. We define our sequence network as 
\begin{equation}
\label{seqnet-formula}
    \begin{split}
    \mathbf{i}_k^{(t)} &= \sigma\left(\mathbf{W}_i\vec{\alpha}_k^{(t)} + \mathbf{U}_i\mathbf{h}_k^{(t-1)} + \mathbf{b}_i\right) \\
    \mathbf{f}_k^{(t)} &= \sigma\left(\mathbf{W}_f\vec{\alpha}_k^{(t)} + \mathbf{U}_f\mathbf{h}_k^{(t-1)} + \mathbf{b}_f\right) \\
    \mathbf{o}_k^{(t)} &= \sigma\left(\mathbf{W}_o\vec{\alpha}_k^{(t)} + \mathbf{U}_o\mathbf{h}_k^{(t-1)} + \mathbf{b}_o\right) \\
    \mathbf{g}_k^{(t)} &= \sigma\left(\mathbf{W}_g\vec{\alpha}_k^{(t)} + \mathbf{U}_g\mathbf{h}_k^{(t-1)} + \mathbf{b}_g\right) \\
    \mathbf{c}_k^{(t)} &= \mathbf{f}_k^{(t)} \odot \mathbf{c}_k^{(t-1)} + \mathbf{i}_k^{(t)} \odot \mathbf{g}_k^{(t)} \\
    \mathbf{h}_k^{(t)} &= \mathbf{o}_k^{(t)} \odot \tanh\left(\mathbf{c}_k^{(t)}\right) + \mathds{1}(t=1)\odot\mathbf{V}_k^{(M)} \\
    \mathbf{y}_k^{(t)} &= \delta\left(\mathbf{W}_y\mathbf{h}_k^{(t)} + \mathbf{b}_y\right)
    \end{split}
\end{equation}
where $\sigma$ is a \texttt{sigmoid} activation function, $\mathbf{i}_k^{(t)}, \mathbf{f}_k^{(t)}, \mathbf{o}_k^{(t)}$ and $\mathbf{g}_k^{(t)}$ are the internal gates. With $d$ hidden units in the sequence network, $\mathbf{c}_k^{(t)}\in\mathbb{R}^{d}, \mathbf{h}_k^{(t)}\in\mathbb{R}^{d}$ are the cell states and hidden states of the sequence network. $\mathbf{y}_k^{(t)} \in \mathbb{R}^{r}$ is the predicted next item in sequence computed using \texttt{softmax} function. With the \texttt{softmax} activation function defined as
 \begin{equation}
 \delta(\mathbf{z}) = \frac{\exp{(z_j)}}{\sum_{i=1}^{r}\exp{(z_i)}}, \forall j = 1,\cdots,r
 \end{equation}
where the $i$-th element in $\mathbf{z}$ is denoted as $z_i$, the $\mathbf{y}_k^{(t)}$ can be interpreted as the probability distribution over $r$ items. 

We integrate the attribute network by conditioning the hidden states in the sequence network at the first time step (denoted as $\mathds{1}(t=1)\odot\mathbf{V}_k^{(M)}$). To do this, the attribute network and sequence network should have the same number of hidden units, \ie, $d_M = d$. 
After processing the last time step for an attributed sequence $\mathbf{S}_k$, the cell state of sequence network, namely $\mathbf{c}_k^{(l_k)}$, is used as the feature representation of $\mathbf{S}_k$. 
We denote the set of parameters in sequence network as $\phi_S = \{\mathbf{W}_S, \mathbf{U}_S, \mathbf{b}_S, \mathbf{W}_y, \mathbf{b}_y\}$, where $\mathbf{W}_S = (\mathbf{W}_i, \mathbf{W}_f, \mathbf{W}_o, \mathbf{W}_g)$, $\mathbf{U}_S = (\mathbf{U}_i, \mathbf{U}_f, \mathbf{U}_o, \mathbf{U}_g)$ and $\mathbf{b}_S = (\mathbf{b}_i, \mathbf{b}_f, \mathbf{b}_o, \mathbf{b}_g)$. 

\subsection{Attributed Sequence Learning}
\label{subsection-asl}

The attributed sequence embeddings would not be useful for downstream mining tasks if the embedding space is detached from the inputs. Thus, we choose two different learning objective functions for the attribute network and sequence network targeting at the different characteristics of attribute and sequence data. 

Attribute network aims at minimizing the differences between the input and reconstructed attribute values. The learning objective function of attribute network is defined as:
\begin{equation}
    L_{A} = \|\mathbf{x}_k - \widehat{\mathbf{x}_k}\|_2^2
\end{equation}

Sequence network aims at minimizing log likelihood of the incorrect prediction of the next item at each time step. Thus, the sequence network learning objective function can be formulated using categorical cross-entropy as:
\begin{equation}
    L_{S} = -\sum_{t=1}^{l_k} \vec{\alpha}_k^{(t)}\log\mathbf{y}_k^{(t)}
\end{equation}
\begin{algorithm}[!ht]
    \begin{algorithmic}[1]
      \caption{\nas Learning}
      \label{algorithm-learn}
      \INPUT $\mathcal{J}=\{J_1, \cdots, J_n\}$, attribute network layers $2M$, learning rate $\gamma$, iteration number of attribute network $\mathcal{T}_A$ and sequence network $\mathcal{T}_S$ and convergence error $\varepsilon_A$ and $\varepsilon_S$. 
      \OUTPUT Parameter sets $\phi_A, \phi_S$.
      \State{Initialize $\phi_A$ and $\phi_S$ using uniform distribution.}
      \ForEach{$J_k \in \mathcal{J}, k = 1,\cdots,n$}
          \ForEach{$\tau = 1, \cdots, \mathcal{T}_A$}
              \ForEach{$m = 1,\cdots, 2M$}
                  \State{Compute forward propagation.}
              \EndFor
              \ForEach{$m = 2M, \cdots, 1$}
                  \State{Compute the gradient of layer $m$.}
              \EndFor
              \ForEach{$m = 1, \cdots, 2M$}
                  \State{Update the parameter set $\phi_A$}
              \EndFor
              \State{Calculate $L_{A}^{\tau}$ }
              \If{$\tau > 1$ and $|L_{A}^{\tau} - L_{A}^{\tau - 1}|<\varepsilon_A$}
                  \State{Stop iterating. }
              \EndIf
          \EndFor
          \ForEach{$\tau = 1, \cdots, \mathcal{T}_S$}
              \ForEach{$t = 1, \cdots, l_k$}
                  \State{Compute forward propagation and get $\mathbf{y}_k^{t}$.}
                  \State{Compute the gradients of sequence network $\Phi_S$.}
                  \State{Update the parameter set $\phi_S$.}
              \EndFor
              \State{Calculate $L_{S}^{\tau}$ }
              \If{$\tau > 1$ and $|L_{S}^{\tau} - L_{S}^{\tau - 1}|<\varepsilon_S$}
                  \State{Stop iterating. }
              \EndIf
          \EndFor
      \EndFor
    \end{algorithmic}
  \end{algorithm}
  
Also, the learning processes are composed of a number of iterations, and the parameters are updated during each iteration based on the gradient computed. Without loss of generality, we denote $L_{A}^{\tau}$ and $L_{B}^{\tau}$ as the $\tau$-th iteration of attribute network and sequence network, respectively. We further denote the maximum numbers of iterations for attribute network and sequence network as $\mathcal{T}_A$ and $\mathcal{T}_B$. $\mathcal{T}_A$ and $\mathcal{T}_B$ may not be equal as the number of iterations needed for the attribute network, and sequence network may not be the same. 
Algorithm~\ref{algorithm-learn} summarizes the learning paradigm under our proposed \nas framework. 
After the attributed sequence learning process, we use the parameters in the attribute network and sequence network to embed each attributed sequence. 

In this section, we evaluate \nas framework using real-world application logs from Amadeus and public datasets from Wikispeedia~\cite{west2012human,west2009wikispeedia}. 
We evaluate the quality of embeddings generated by different methods by measuring the performance of outlier detection and clustering algorithms using different embeddings. Outlier detection and clustering tasks are frequently being used for many applications, such as fraud detection and user behavior analysis. We also include three methods not using neural networks in outlier detection tasks. 
We also study four case studies in the security management system in Amadeus to demonstrate the embeddings produced by \nas is useful for real-world applications. 
Lastly, we show that the \nas framework is capable of embedding attributed sequences in real-time. 
\section{Experimental Setup}
\subsection{Data Collection} 
We use four datasets in our experiments: two from Amadeus application log files and two from Wikispeedia\footnote{Personal identity information are not collected for privacy reasons. }. The numbers of attributed sequences in all four datasets vary between $\sim$58k and $\sim$106k. 


\begin{itemize}
    \item \texttt{AMS-A}: We extract $\sim$58k instances from log files of an Amadeus internal application. Each record is composed of a profile containing information ranging from system configurations to office name, and a sequence of functions invoked by click activities on the web interface. There are 288 distinct functions, 57,270 distinct profiles in this dataset. The average length of the sequences is 18. 
    \item \texttt{AMS-B}: We use $\sim$106k instances from Amadeus internal application log files with 573 distinct functions and 106,671 distinct profiles. The average length of the sequences is 22. 
    \item \texttt{Wiki-A}: This dataset is sampled from Wikispeedia dataset\footnote{Download link: \texttt{http://goo.gl/8Z9h9f}}. Wikispeedia dataset originated from a human-computation game, called Wikispeedia~\cite{west2009wikispeedia}. In this game, each user is given two pages ({\ie}, source, and destination) from a subset of Wikipedia pages and asked to navigate from the source to the destination page.  We use a subset of $\sim$2k paths from Wikispeedia with the average length of the path as 6. We also extract 11 sequence context (\eg, the category of the source page, average time spent on each page) as attributes. 
    \item \texttt{Wiki-B}: This dataset is also sampled from Wikispeedia dataset. In \texttt{Wiki-B}, we use $\sim$1.5k paths from Wikispeedia with the average length of the path as 8. We also extract 11 sequence context as attributes. 
\end{itemize}

\subsection{Compared Methods} 
\begin{table}[t]
  \centering
  \caption{Compared methods.}
    \label{table-methods}%
    \begin{tabular}{clp{5.5cm}l}
    \toprule 
    Methods & Data Used & {Short Descriptions} & {Related Publication}\\
    \midrule  
    \texttt{LEN}    &  Attributes & The encoded attributes is used.  & \cite{akata2013label} \\
    \midrule
    \texttt{MCC}   &  Sequences & A markov chain based method. & \cite{bernhard2016clickstream} \\
    \midrule
    \texttt{ATR} & Attributes & An autoencoder for attribute data.  & \cite{wang2014generalized} \\
    \midrule
    \texttt{SEQ} & Sequences & Sequence embedding using LSTM.  & \cite{sutskever2014sequence} \\
    \midrule
    \multirow{2}{*}{\texttt{EML}} & Attributes & Aggregation of the scores of & \multirow{2}{*}{\cite{yager2014probabilistically}} \\
    & Sequences &  \texttt{LEN} and \texttt{MCC} methods. & \\
    \midrule
    \multirow{2}{*}{\texttt{CSA}} &  Attributes & Concatenation of embeddings  & \multirow{2}{*}{\cite{ngiam2011multimodal}} \\
          &  Sequences & generated by \texttt{ATR} and \texttt{SEQ} methods.  &  \\
    \midrule
    \multirow{2}{*}{\nasns} &  Attributes & Attribute network is used to & \multirow{2}{*}{This Work}  \\
          &  Sequences & condition sequence network. &  \\
    \bottomrule
    \end{tabular}%
  \vspace{-5mm}
\end{table}%
To study \nas performance on attributed sequences in real-world applications, we use the compared methods in Table~\ref{table-methods} in outlier detection and clustering tasks. 
\begin{itemize}
\item 
\texttt{LEN}~\cite{akata2013label}: The attributes are encoded and directly used in the mining algorithm. 
\item 
\texttt{MCC}~\cite{bernhard2016clickstream}: 
\texttt{MCC} uses the sequence component of attributed sequence as input and produces log likelihood for each sequence.
\item 
\texttt{SEQ}~\cite{sutskever2014sequence}: Only the sequence inputs are used by an LSTM to generate fixed-length embeddings. 
\item 
\texttt{ATR}~\cite{wang2014generalized}: A fully connected neural network with two layers is applied on only attributes to generate embeddings.
\item 
\texttt{EML}\cite{yager2014probabilistically}: The scores from \texttt{MCC} and \texttt{LEN} are aggregated.  
\item 
\texttt{CSA}~\cite{ngiam2011multimodal}: The attribute embedding and the sequence embedding are first generated by \texttt{ATR} and \texttt{SEQ}, respectively. Then the two embeddings are concatenated together. 
\item 
\nasns: Our proposed \nas framework using both attributes and sequences to generate embeddings. 
\end{itemize} 

\subsection{Network Parameters}
Following the previous work in~\cite{glorot2010understanding}, we initialize weight matrices $\mathbf{W}_A$ and $\mathbf{W}_S$ using the uniform distribution. The recurrent matrix $\mathbf{U}_S$ is initialized using the orthogonal matrix as suggested by~\cite{saxe2013exact}. All the bias vectors are initialized with zero vector $\pmb0$. We use stochastic gradient descent as optimizer with the learning rate of 0.01. We use a two-layer encoder-decoder stack as our attribute network. 
\begin{figure}[!ht]
    \centering
    \includegraphics[width=0.55\linewidth]{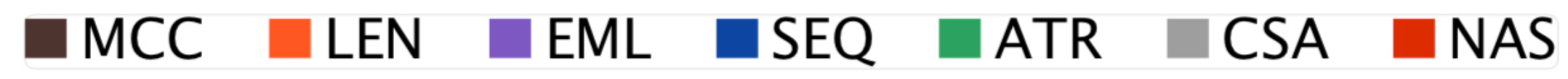}
    
    \begin{subfigure}[!ht]{0.42\linewidth}
        \centering
        \includegraphics[page=1, width=\linewidth]{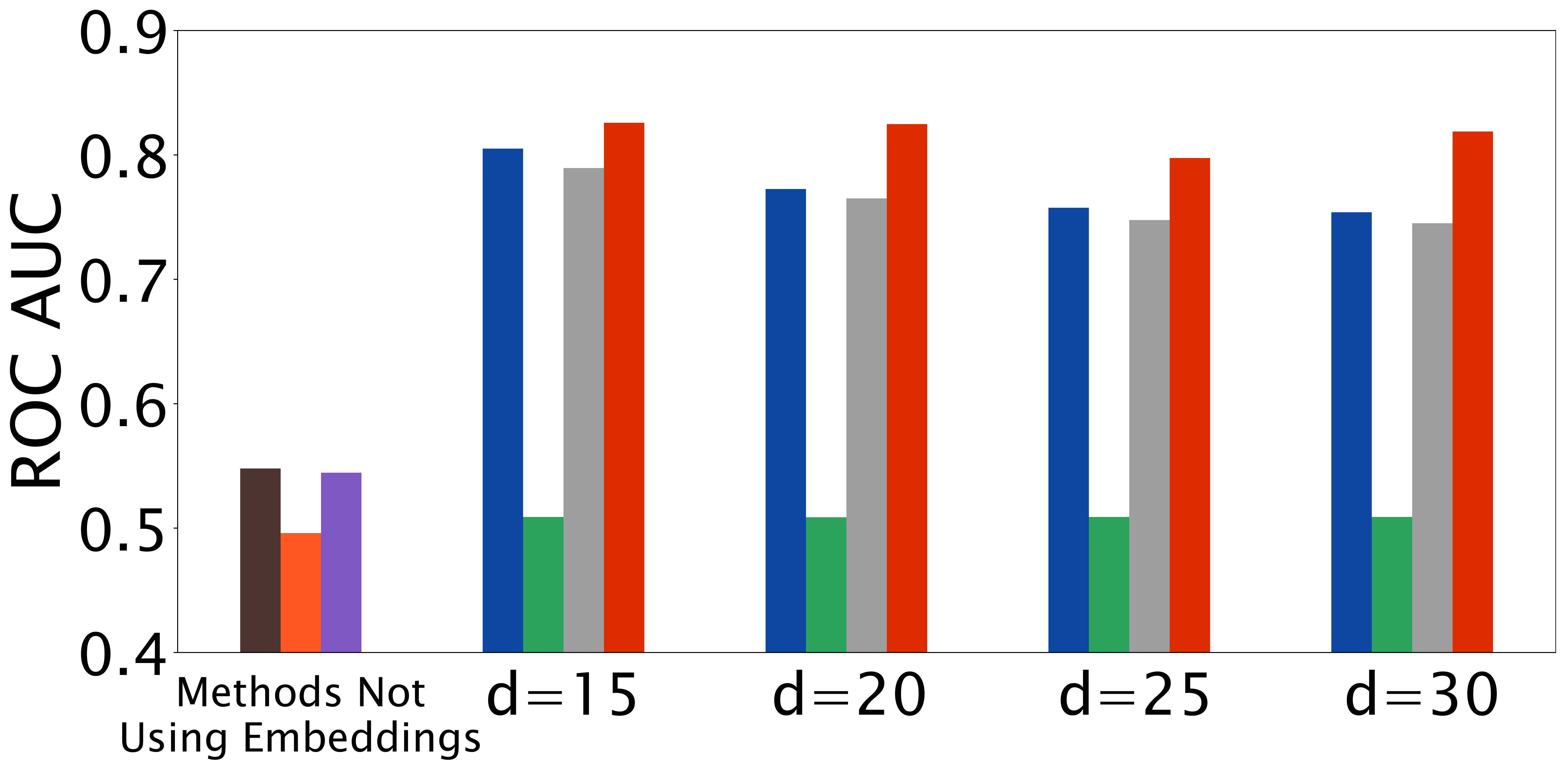}
        \caption{\texttt{AMS-A} Dataset}
        \label{fig-outlier-perf-auc-a}
    \end{subfigure}
    \hspace{5mm}
    \begin{subfigure}[!ht]{0.42\linewidth}
        \centering
        \includegraphics[page=2, width=\linewidth]{figures/NAS/exp/auc_perf_bar.pdf}
        \caption{\texttt{AMS-B} Dataset}
        \label{fig-outlier-perf-auc-b}
    \end{subfigure}
    \begin{subfigure}[!ht]{0.42\linewidth}
        \centering
        \includegraphics[page=3, width=\linewidth]{figures/NAS/exp/auc_perf_bar.pdf}
        \vspace{-5mm}
        \caption{\texttt{Wiki-A} Dataset}
        \label{fig-outlier-perf-wiki-A}
    \end{subfigure}
    \hspace{5mm}
    \begin{subfigure}[!ht]{0.42\linewidth}
        \centering
        \includegraphics[page=4, width=\linewidth]{figures/NAS/exp/auc_perf_bar.pdf}
        \vspace{-5mm}
        \caption{\texttt{Wiki-B} Dataset}
        \label{fig-outlier-perf-wiki-b}
    \end{subfigure}
        \vspace{-5mm}
    \caption[\nas performance of $k$-NN outlier detection]{The performance of $k$-NN outlier detection ($k=5$). The \textit{methods not using embeddings} are placed on the left. We vary the number of dimensions on the right. The higher score is better. We observe that the combinations of $k$-NN and \nas embeddings have the best performance on the four datasets. }
    \label{fig-outlier-perf}
    \vspace{-9mm}
\end{figure}
\section{Outlier Detection Tasks}
\label{outlier}
We first use outlier detection tasks to evaluate the quality of embeddings produced by \nasns. We select $k$-NN outlier detection algorithm as it has only one important parameter (\ie, the $k$ value). We use ROC AUC as the metric in this set of experiments. 

For each of the \texttt{AMS-A} and \texttt{AMS-B} datasets, we ask domain experts to select two users as inlier and outlier. These two users have completely different behaviors ({\ie}, sequences) and metadata ({\ie}, attributes). The percentages of outliers in \texttt{AMS-A} and \texttt{AMS-B} are 1.5\% and 2.5\% of all attributed sequences, respectively. 
For the \texttt{Wiki-A} and \texttt{Wiki-B} datasets, each path is labeled based on the category of the source page. Similarly to the previous two datasets, we select paths with two labels as inliers and outliers where the percentage of outlier paths is 2\%. The feature used to label paths is excluded from the learning and embedding process. 

\subsection{Performance}
\begin{figure}[ht]
    \centering
    \includegraphics[page=1, width=0.35\linewidth]{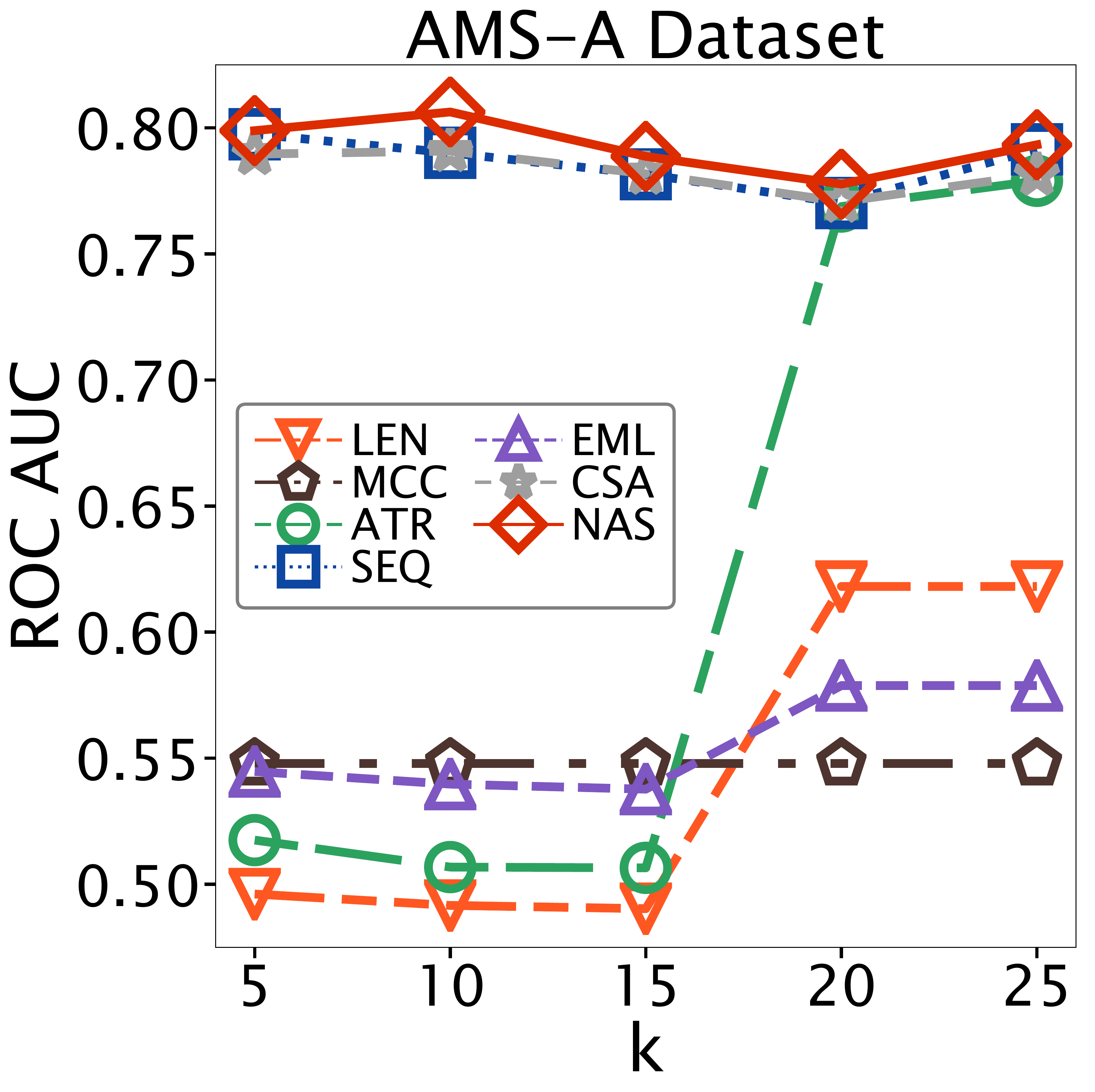}
    \includegraphics[page=2, width=0.35\linewidth]{figures/NAS/exp/auc_k.pdf}
    \includegraphics[page=3, width=0.35\linewidth]{figures/NAS/exp/auc_k.pdf}
    \includegraphics[page=4, width=0.35\linewidth]{figures/NAS/exp/auc_k.pdf}     
    \caption[\nas parameter sensitivity to $k$ values]{Parameter sensitivity to different $k$ values. It is shown that the embeddings generated by \nas always have the best performance under different $k$ values. }
    \label{exp-auc-k}
\end{figure}
\begin{figure}[ht]
\centering
    \includegraphics[page=1, width=0.35\linewidth]{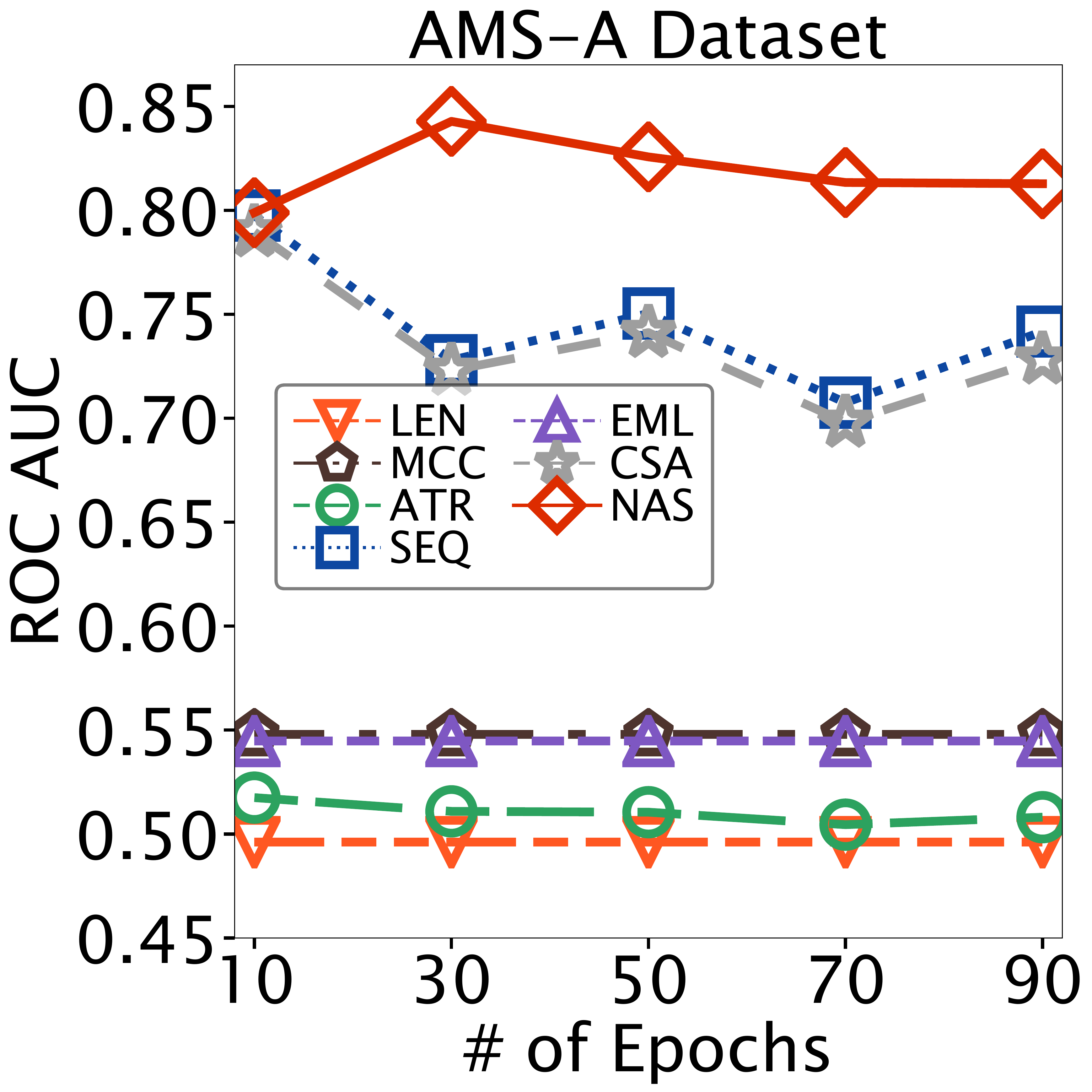}
    \includegraphics[page=2, width=0.35\linewidth]{figures/NAS/exp/auc_epoch.pdf}
    \includegraphics[page=3, width=0.35\linewidth]{figures/NAS/exp/auc_epoch.pdf}
    \includegraphics[page=4, width=0.35\linewidth]{figures/NAS/exp/auc_epoch.pdf}         
    \caption[\nas performance in outlier detection (varying epoches)]{Performance comparisons using outlier detection tasks. The embeddings generated by \nas can always achieve the best performance compared to baseline methods when the number of training epochs increases. }
    \label{exp-auc-epoch}
\end{figure}

The goal of this set of experiments is to demonstrate the performance of outlier detection using all our compared methods. Each method is trained with all the instances. 
For \texttt{SEQ}, \texttt{ATR} and \nasns, the number of learning epochs is set to 10 and we vary the number of embedding dimensions $d$ from 15 to 30. 
We set $k=5$ for outlier detection tasks for \texttt{LEN}, \texttt{SEQ}, \texttt{ATR}, \texttt{CSA} and \nasns. Choosing the \textit{optimal} $k$ value in the outlier detection tasks is beyond the scope of this work, thus we omit its discussions. We summarize the performance results in Figure~\ref{fig-outlier-perf}. 

\textbf{Analysis.} We find that the results based on the embeddings generated by \nas are superior to other methods. 
That is, \nas maximally outperforms other state-of-the-art algorithms by 32.9\%, 27.5\%, 44.8\% and 48\% on \texttt{AMS-A}, \texttt{AMS-B}, \texttt{Wiki-A} and \texttt{Wiki-B} datasets, respectively. It is worth mentioning that \nas outperforms a similar baseline method \texttt{CSA} by incorporating the information of attribute-sequence dependencies. 

\subsection{Parameter Study}
There are two \textit{key} parameters in our evaluations, \ie, $k$ value for the $k$-NN algorithm and the learning epochs. 

In Figure \ref{exp-auc-k}, we first show that the embeddings (dimension $d=15$) generated by our \nas assist $k$-NN outlier detection algorithm to achieve superior performance under a wide range of $k$ values ($k=5, 10, 15, 20, 25$). We omit the detailed discussions of selecting the optimal $k$ values due to the scope of this work. 

Next, we evaluate the performance changes as the number of training epochs increases. 
We do not use the early stopping in this set of experiments as we want to fix the number of training epochs. 
Figure \ref{exp-auc-epoch} presents the results when we fix $k=5, d=15$ and vary the number of epochs in the learning phase. We notice that compared to its competitor, the embeddings generated by \nas can achieve a higher AUC even with a relatively fewer number of learning epochs. Compared to other neural network-based methods (\ie, \texttt{SEQ}, \texttt{ATR} and \texttt{CSA}), \nas have a more stable performance. The \nas performance gain is not due to the advantage of using both attributes and sequences, but because of taking the various dependencies into account, as the other two competitors (\ie, \texttt{CSA} and \texttt{EML}) also utilize the information from both attributes and sequences. 

\section{Clustering Tasks}
\label{clustering}
We use clustering tasks to evaluate the quality of the embeddings generated by compared methods. We use the HDBSCAN clustering algorithm from~\cite{campello2015hierarchical} since the results remain stable over different runs, which is a fair and ideal choice for studying the performance of representation learning algorithms. We use normalized mutual information (NMI) to measure the quality of the clustering. The highest NMI score is 1. 

\subsection{Performance}
\begin{figure}[!ht]
    \centering
    \includegraphics[width=0.35\linewidth]{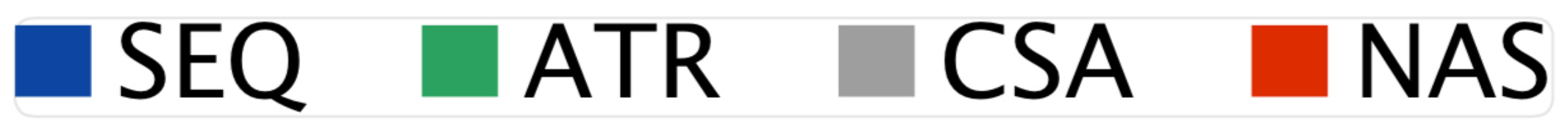}

    \begin{subfigure}[t]{0.42\linewidth}
        \centering
        \includegraphics[page=1, width=\linewidth]{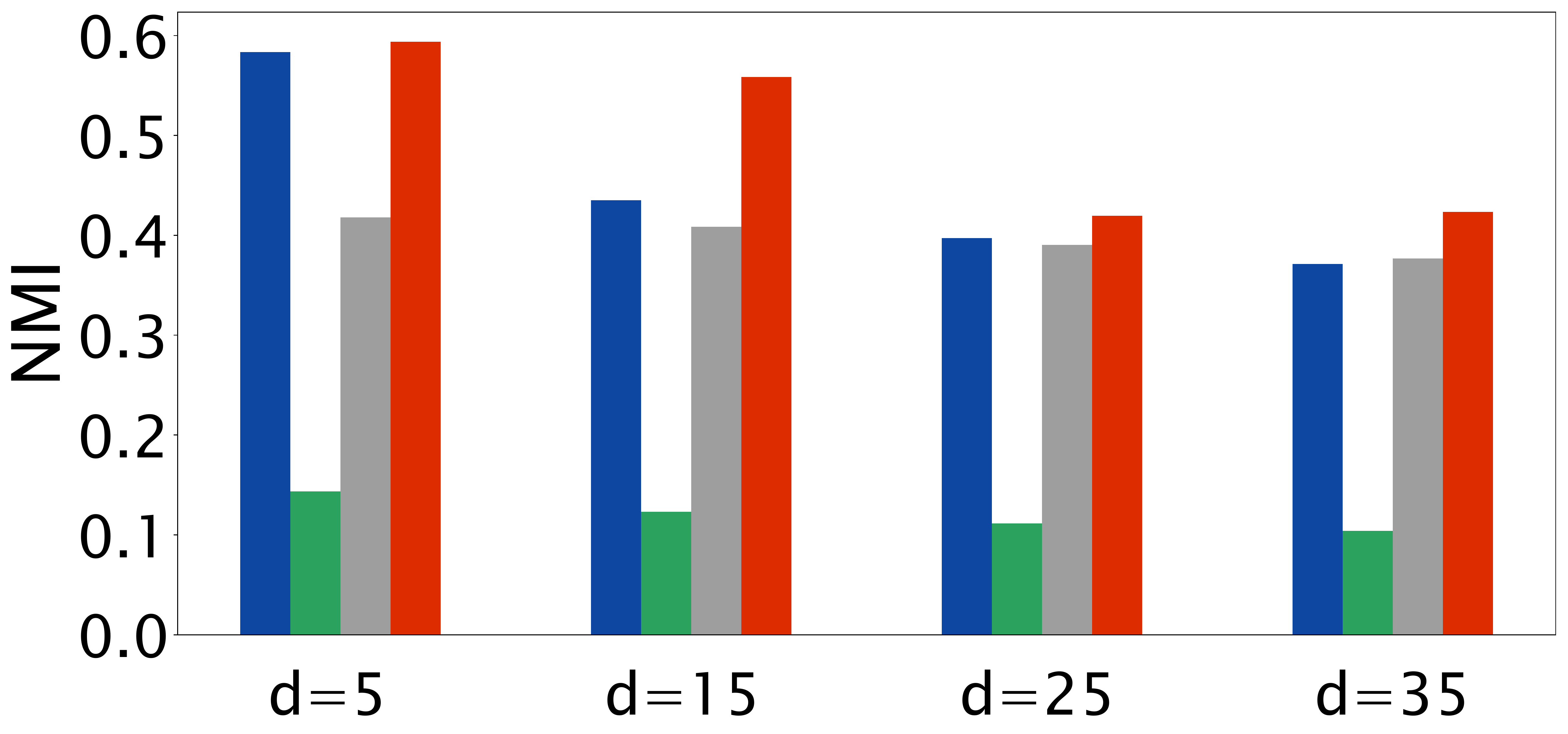}
        \caption{\texttt{AMS-A} Dataset}
        \label{fig-clustering-perf-nmi-a}
    \end{subfigure}
    \hspace{5mm}
    \begin{subfigure}[t]{0.42\linewidth}
        \centering
        \includegraphics[page=2, width=\linewidth]{figures/NAS/exp/nmi_perf_bar.pdf}
        \caption{\texttt{AMS-B} Dataset}
        \label{fig-clustering-perf-nmi-b}
    \end{subfigure}
    \begin{subfigure}[t]{0.42\linewidth}
        \centering
        \includegraphics[page=3, width=\linewidth]{figures/NAS/exp/nmi_perf_bar.pdf}
        \caption{\texttt{Wiki-A} Dataset}
        \label{fig-clustering-perf-wiki-a}
    \end{subfigure}
    \hspace{5mm}
    \begin{subfigure}[t]{0.42\linewidth}
        \centering
        \includegraphics[page=4, width=\linewidth]{figures/NAS/exp/nmi_perf_bar.pdf}
        \caption{\texttt{Wiki-B} Dataset}
        \label{fig-clustering-perf-wiki-b}
    \end{subfigure}
    \caption[\nas performance in clustering]{Performance of HDBSCAN clustering algorithm with the minimum cluster size of 640. The higher score is better. By using the embeddings produced by \nasns, the clustering algorithm performs better than using embeddings produced by other baseline methods. }
    \label{fig-clustering-perf}
\end{figure}
In Figure~\ref{fig-clustering-perf} we report the result when the minimum cluster size is fixed to 640, and we vary the number of embedding dimensionality $d$ from 5 to 35. 
\nas is capable of taking advantage of using the information not only from both attributes and sequences but also the attributed-sequence dependencies when generating the embeddings. Specially, when compared to the \texttt{CSA} method where the information from both attributes and sequences is used but without taking advantage of the attribute-sequence dependencies, \nas outperforms \texttt{CSA} by 24.7\%, 20.6\%, 3.8\%, 4.3\% on average on \texttt{AMS-A}, \texttt{AMS-B}, \texttt{Wiki-A} and \texttt{Wiki-B} datasets, respectively. 

\subsection{Parameter Study}

\begin{figure}[t]
    \centering
    \includegraphics[page=1, width=0.35\linewidth]{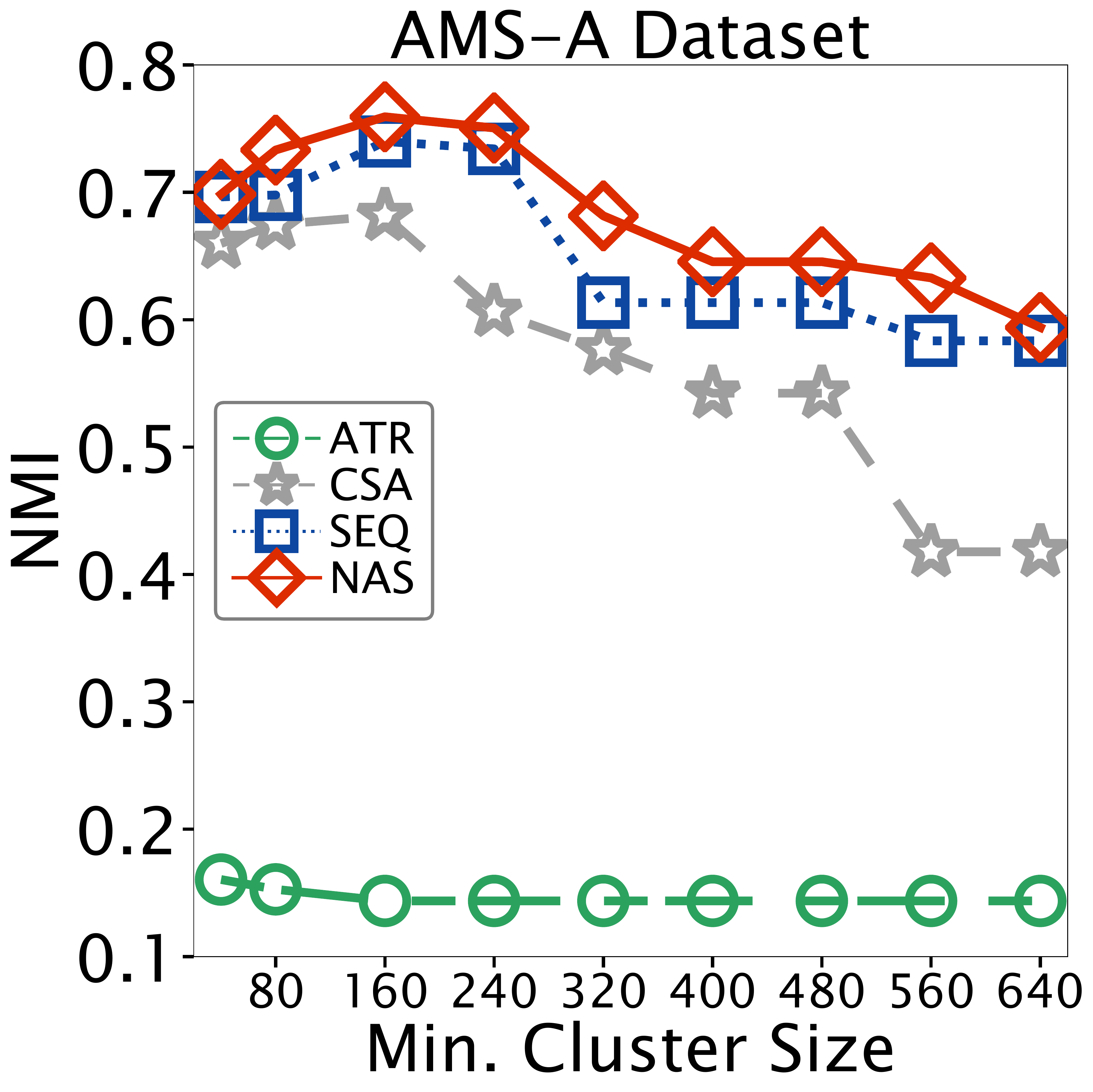}
    \includegraphics[page=2, width=0.35\linewidth]{figures/NAS/exp/nmi_cs.pdf}
    \includegraphics[page=3, width=0.35\linewidth]{figures/NAS/exp/nmi_cs.pdf}
    \includegraphics[page=4, width=0.35\linewidth]{figures/NAS/exp/nmi_cs.pdf}    
    \caption[\nas parameter study in clustering]{Clustering algorithm using the embeddings generated by \nas can achieve better performance than baselines under different minimum cluster size parameter. }
    \label{exp-nmi-cs}
\end{figure}

In this set of experiments, we investigate the performance of \nas under different parameter settings. We first vary the minimum size of clusters for all three datasets. For each dataset, we evaluate on the same range of the minimum cluster size. Figure \ref{exp-nmi-cs} depicts the NMI of the HDBSCAN clustering algorithm when the dimensionality of embedding is fixed to 5, and the number of learning epochs is set to 10. We observe that \nas is always achieving the highest NMI among the four methods. 
\begin{figure}[!ht]
\centering
    \includegraphics[page=1, width=0.35\linewidth]{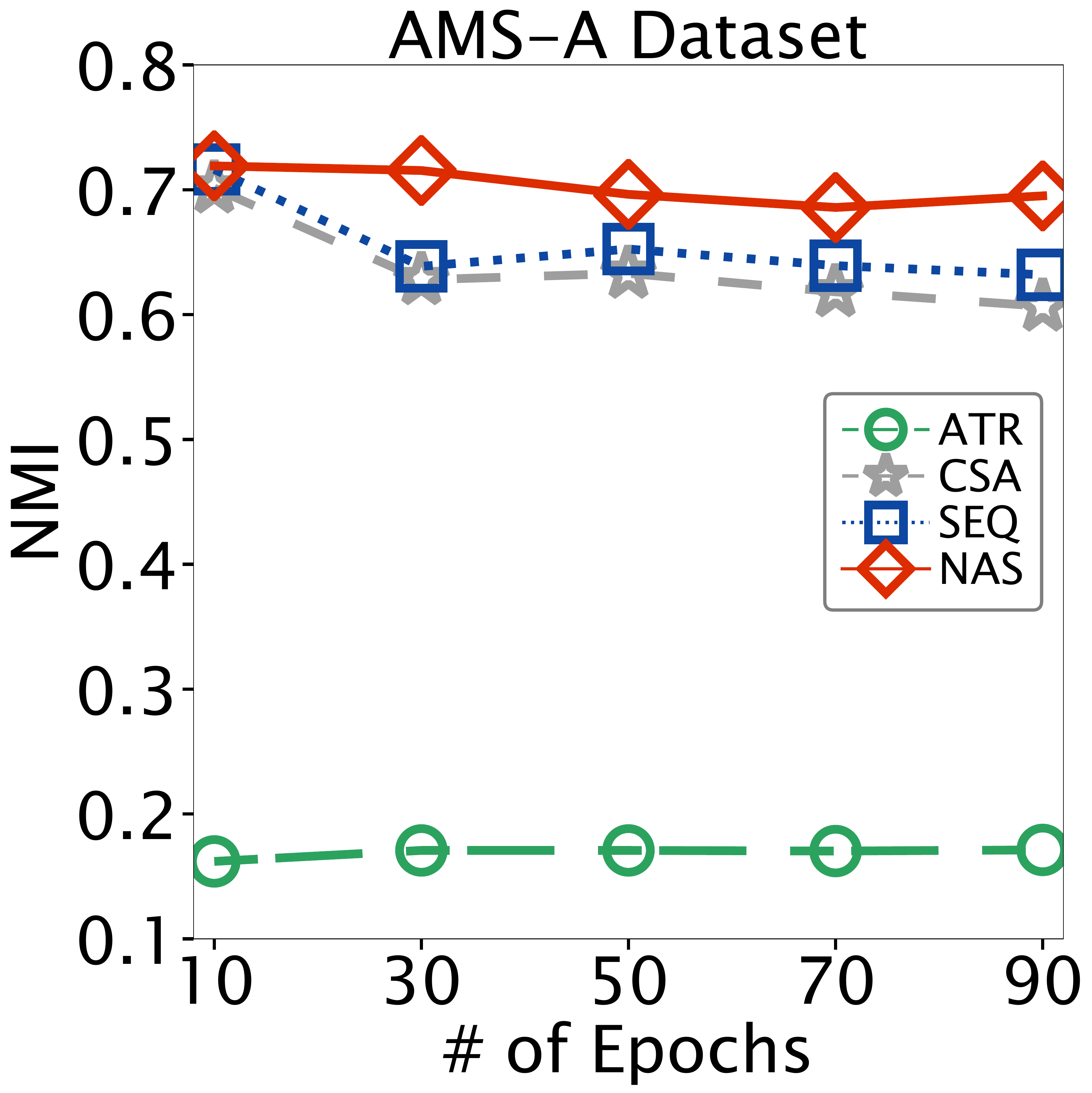}
    \includegraphics[page=2, width=0.35\linewidth]{figures/NAS/exp/nmi_epoch.pdf}
    \includegraphics[page=3, width=0.35\linewidth]{figures/NAS/exp/nmi_epoch.pdf}
    \includegraphics[page=4, width=0.35\linewidth]{figures/NAS/exp/nmi_epoch.pdf}        
    \caption[\nas parameter study with varying epochs]{The embeddings generated by \nas have better performance than the baseline methods across different numbers of training epochs. }
    \label{exp-nmi-epoch}
\end{figure}
\begin{figure}[!ht]
    \centering
    \includegraphics[width=0.8\textwidth]{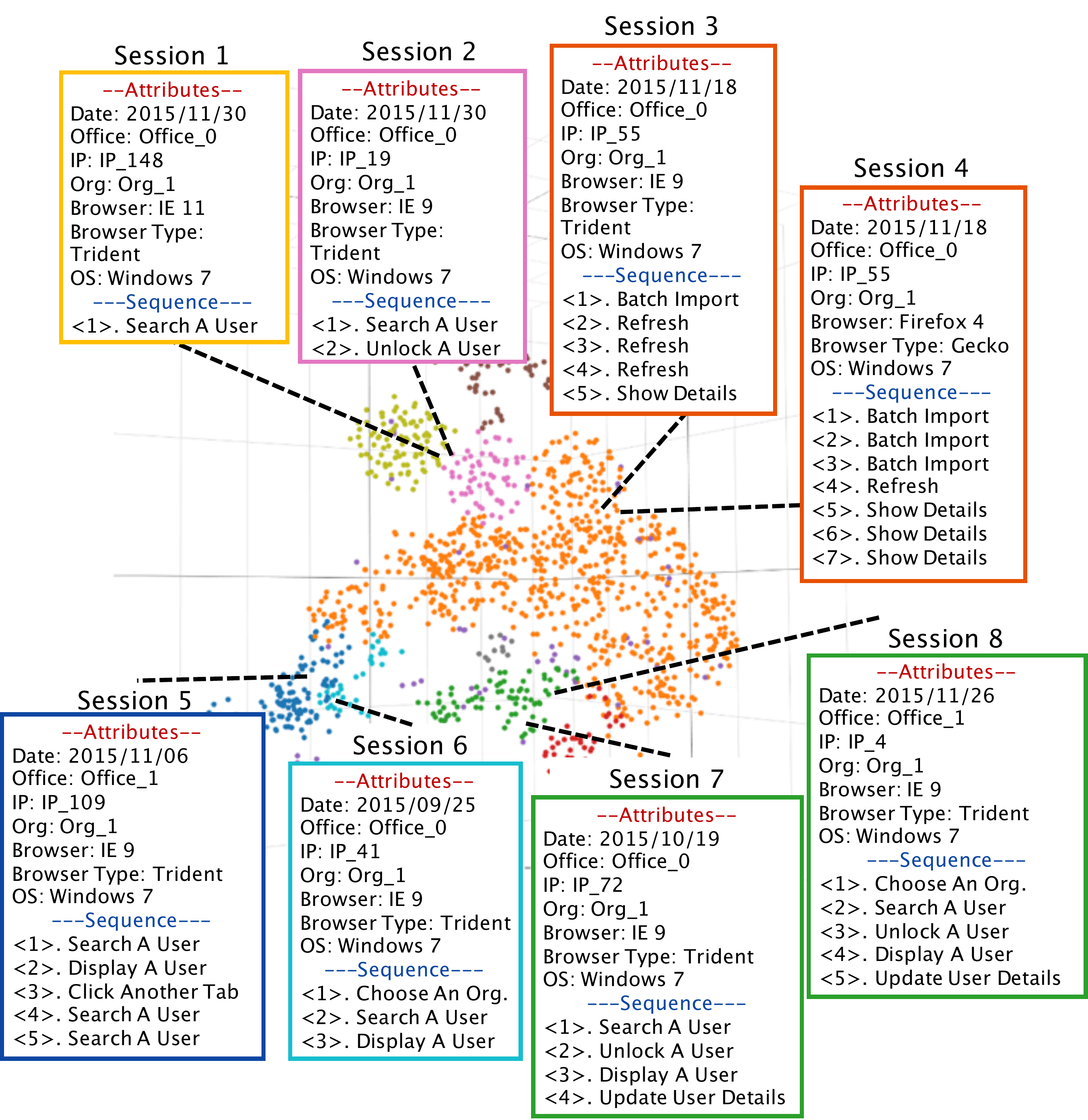}
     \vspace{-3mm}
    \caption[Case studies: \nas in clustering]{Clustering results for case studies. Each color represents one cluster. Only major attributes are shown here. Actual attribute values and item names in sequences are replaced for privacy reasons. (Better view in color.)}
    \vspace{-6mm}
    \label{fig-clustering-case-study}
\end{figure}

In Figure \ref{exp-nmi-epoch} we choose a fixed minimum cluster size of 40 and vary the number of learning epochs from 10 to 90. We observe that \nas is superior to its competitors across all parameter settings. 

\subsection{Case Study: Security Management System}

In this case study, we build HDBSCAN clusters on the \texttt{AMS-A} dataset. Each point in Figure \ref{fig-clustering-case-study} is a user session in \texttt{AMS-A} dataset. We apply t-SNE~\cite{maaten2008visualizing} on the embeddings to generate a 3D plot with each color for one cluster. 
We ask domain experts to closely examine the similarities and differences between the attributed sequences in each cluster. Based on eight user sessions in Figure~\ref{fig-clustering-case-study}, we summarize four case studies as follows: 

\begin{itemize}
\item \noindent \textbf{Case Study 1.} The \textit{similarities} and \textit{differences} between the two ``nearby'' points from the \textit{same} cluster. Sessions 3 and 4 share the same set of actions, and the order of each kind of action is the same. Further, they both share the same attributes. In a more complex case of Sessions 7 and 8, although they have different sets of actions ({\ie}, Session 8 has ``\texttt{Choose An Org}.'' which Session 7 does not) and not all attributes are the same, they are clustered together since \textit{both} of them are ``high-level administrative'' sessions. 
\item \noindent \textbf{Case Study 2.} \textit{Differences} between two ``\textit{nearby}'' points that belong to \textit{different} clusters. In Sessions 1 and 2, although they share similar attributes ({\ie}, ``\texttt{IE}'', ``\texttt{Windows 7}'', ``\texttt{Office\_0}'', etc.) and only one action is different, the Session 2 belongs to another cluster since that action (``\texttt{Unlock A User}'') changes the type of the session from ``routine'' to ``administrative''. 
\item \noindent \textbf{Case Study 3.}  \textit{Similarities} between two ``\textit{nearby}'' points that belong to \textit{different} clusters. There are a number of differences between Sessions 5 and 6, which cause them to be distant from each other, such as Sessions 5 and 6 belong to different offices, have different IP addresses and have different sets of user actions in the sequences. The similarities between them are obvious: first, they both share a number of similar attributes ({\ie}, ``\texttt{browser}'', ``\texttt{OS}'', and ``\texttt{organization}''); second, the majority of actions remains the same, and they share a subsequence of actions ({\ie}, ``\texttt{Search A User}'' and ``\texttt{Display A User}''). Thus, although they belong to different clusters, the distance between Sessions 5 and 6 is smaller compared to sessions from other clusters.
\item \noindent \textbf{Case Study 4.} A \textit{global} view of the eight clusters. The \nas is capable of differentiating user sessions despite that the 8 user sessions have some similar or identical attribute values (\eg, six of them are from the same office, all of them are using Windows 7), and there are common user actions shared by user sessions. 
\end{itemize}

The above four case studies conclude that the clustering results based on the embeddings generated by \nas can be easily explained in real-world cases. 

\section{Scalability}
\begin{figure}[H]
\centering
    \begin{subfigure}[t]{0.45\linewidth}
    \centering
        \includegraphics[width=0.8\linewidth]{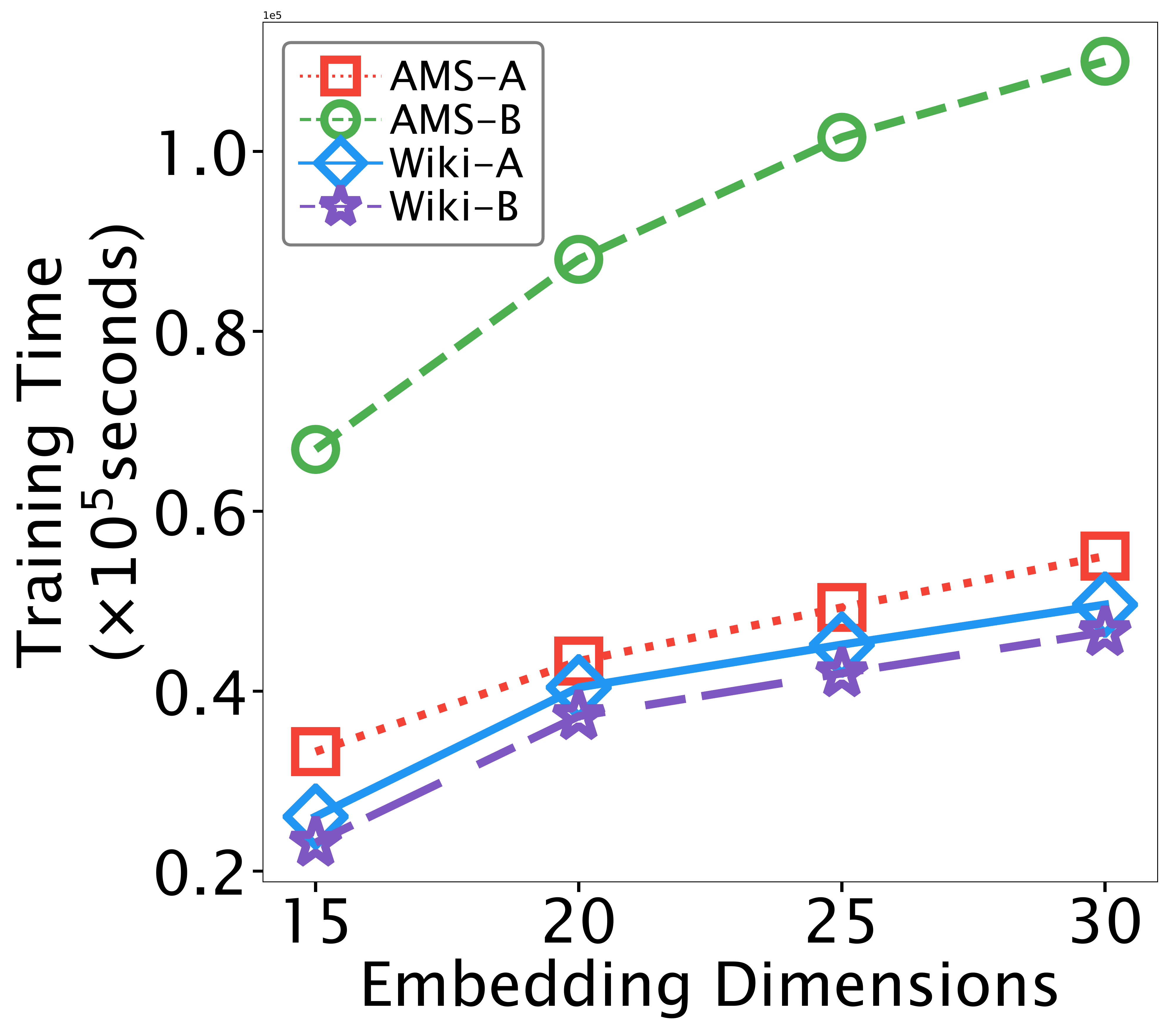}\par 
        \caption{Learning Time}
        \label{fig-scale-1}
    \end{subfigure}
     \hspace{5mm}
    \begin{subfigure}[t]{0.45\linewidth}
    \centering
        \includegraphics[width=0.8\linewidth]{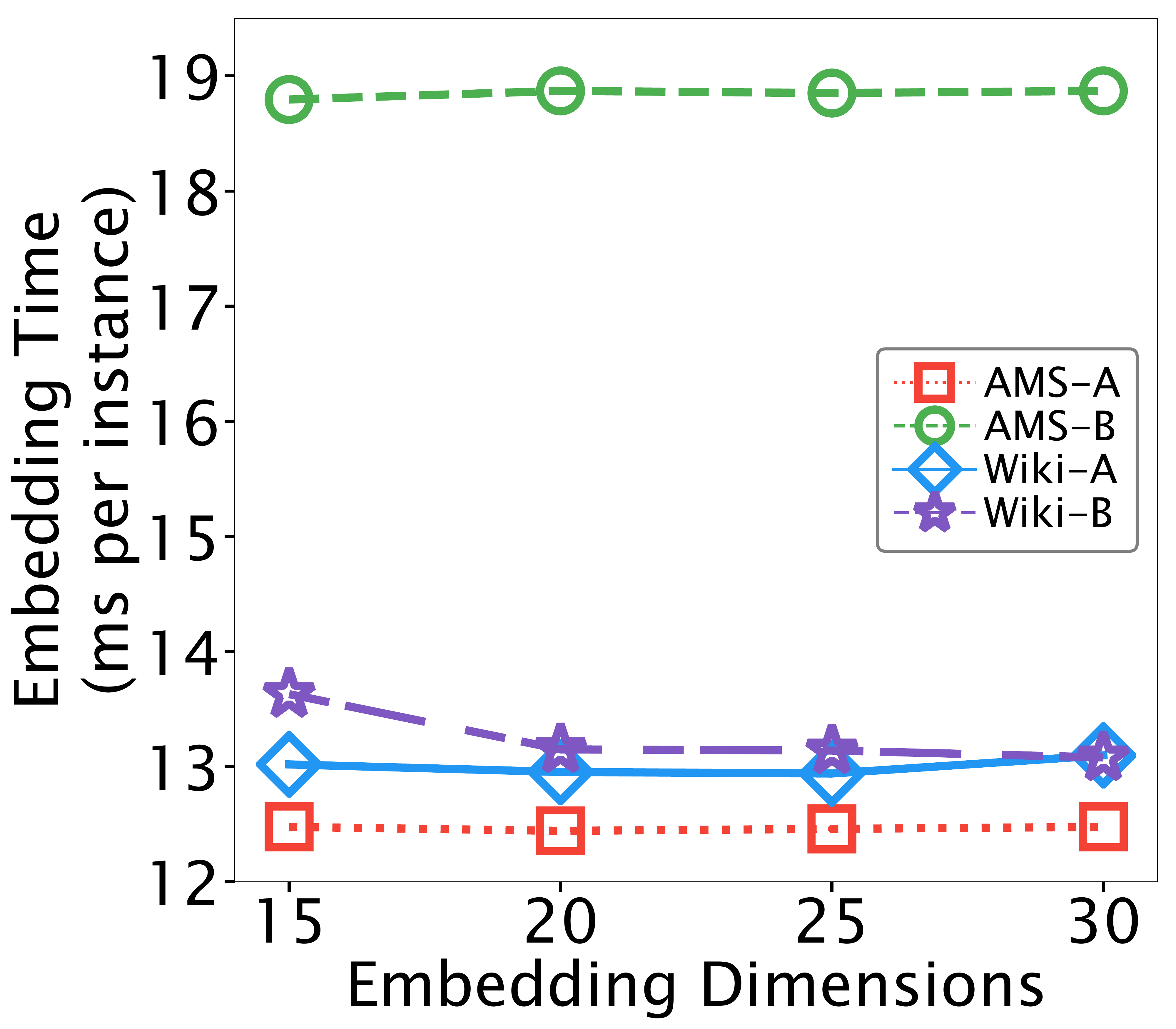}\par 
        \caption{Embedding Time}
    \label{fig-scale-2}
    \end{subfigure}
    \caption[\nas runtime scalability]{Runtime scalability}
    \label{fig-scalability}
\end{figure}
In this set of experiments, we evaluate the learning time and embedding time of the proposed \nas framework.
We implement the \nas framework using Theano 0.8~\cite{theano2012} on Ubuntu 14.04. We conduct our experiments on a machine with 24 E5-2690 v2 cores and 256GB memory. The I/O time for reading datasets into memory and writing embeddings to the disk is excluded. Each setting corresponds to 10 runs times and averaged. 

The embeddings of \nas come from the cell states in each artificial neuron. An increase in the embedding dimensions increases the number of artificial neurons in the model, which results in a larger model with more parameters that needs longer time to train. However, the model learning process is designed to be an offline process. Thus, it does not interfere with the real-time attributed sequence embedding process. Although the learning time increases as the number of embedding dimensions increases, the embedding time per attributed sequence remains at the millisecond level. Thus, \nas is capable of transforming one attributed sequence into one embedding in real-time that is sufficient for real-world data mining tasks.


 \newpage
\chapter{Deep Metric Learning on Attributed Sequences}
\label{chapter-task2}
\newpage

\section{Problem Definition}
\label{section-task2-problem-definition}
Given a nonlinear transformation function $\Theta$ and two attributed sequences $p_i$ and $p_j$ as inputs, deep metric learning approaches~\cite{cvpr-face-verify} often apply the Mahalanobis distance function to the $d$-dimensional outputs of function $\Theta$ as 
\begin{equation}
\label{eq-mahalanobis-distance}
    D_\Theta(p_i, p_j) = \sqrt{(\Theta(p_i) - \Theta(p_j))^\top \pmb\Lambda (\Theta(p_i) - \Theta(p_j))}
\end{equation}
where $\pmb\Lambda \in \mathbb{R}^{d \times d}$ is a symmetric, semi-definite, and positive matrix. When $\pmb\Lambda = \mathbf{I}$, Eq.~\ref{eq-mahalanobis-distance} is transformed to Euclidean distance~\cite{xing2003distance} as:
\begin{equation}
\label{eq-euclidean}
    D_\Theta(p_i, p_j) = \|\Theta(p_i)-\Theta(p_j)\|_2.
\end{equation}
 
Given feedback sets $\mathcal{S}$ and $\mathcal{D}$ of attributed sequences as per Def.~\ref{def-as-feedback} and a distance function $D_\Theta$ as per Eq.~\ref{eq-euclidean}, the goal of deep metric learning on attributed sequences is to find the transformation function $\Theta: (\mathbb{R}^u, \mathbb{R}^{T \times r}) \mapsto \mathbb{R}^d$ with a set of parameters $\theta$ that is capable of mapping the attributed sequence inputs to a space that the distances between each pair of attributed sequences in the similar feedback set $\mathcal{S}$ are minimized while increasing the distances between attributed sequence pairs in the dissimilar feedback set $\mathcal{D}$. Inspired by~\cite{xing2003distance}, we adopt the learning objective as 
\begin{equation}
\label{eq-problem-def}
\begin{split}
\minimize_{\theta} &\sum_{(p_i, p_j, \ell_{ij})\in \mathcal{S}} D_\Theta\left(p_i, p_j\right) \\
\textrm{s.t.} & \sum_{(p_i, p_j, \ell_{ij})\in \mathcal{D}} D_\Theta\left(p_i, p_j\right) \geq g\\
\end{split}
\end{equation}
where $g$ is a group-based margin parameter that stipulates the distance between two attributed sequences from dissimilar feedback set should be larger than $g$. This prevents the dataset from being reduced to a single point~\cite{xing2003distance}.  

\section{The \mlas Network}
\label{section-mlas}

In this section, we first design two distinct networks, called \anet and \snetns, to learn the attribute data and sequence data, respectively. Next, we present several types of \fnet designs to integrate the two networks. Lastly, the existing network composition is augmented by \mnet to employ user feedback into the learning process. 

\subsection{AttNet and SeqNet}
\label{subsection-attnet-seqnet}

\anetns, designed to learn the relationships within attribute data, utilizes a fully connected neural network with multiple layers of nonlinear transformations. 
In particular, for an \anet with $M$ layers, we denote the weight and bias parameters of the $m$-th layer as $\mathbf{W}_A^{(m)}$ and $\mathbf{b}_A^{(m)}, \forall m=1,\cdots, M$. 
Given an attribute vector $\mathbf{x}_k \in \mathbb{R}^u$ as the input, with $d_m$ hidden units in the $m$-th layer of \anetns, the corresponding output is $\mathbf{V}_k^{(m)}\in\mathbb{R}^{d_m}, \forall m=1,\cdots, M$. The structure of \anet is designed as 
\begin{equation}
\begin{split}
\label{eq-att-network}
\mathbf{V}_{k}^{(1)} &= \delta\left(\mathbf{W}_A^{(1)}\mathbf{x}_k + \mathbf{b}_A^{(1)}\right) \\
\mathbf{V}_{k}^{(2)} &= \delta\left(\mathbf{W}_A^{(2)}\mathbf{V}_{k}^{(1)} + \mathbf{b}_A^{(2)}\right) \\
\vdots \\
\mathbf{V}_{k}^{(M)} &= \delta\left(\mathbf{W}_A^{(M)}\mathbf{V}_{k}^{(M - 1)} + \mathbf{b}_A^{(M)}\right)
\end{split}
\end{equation}
where $\delta: \mathbb{R}^{d_m}\mapsto\mathbb{R}^{d_m}$ is a nonlinear activation function. Possible choices of $\delta$ include \texttt{sigmoid}, \texttt{ReLU}~\cite{nair2010rectified} and \texttt{tanh} functions. 

The mechanism of \anet is that, given the attribute vector $\mathbf{x}_k \in \mathbb{R}^u$ as the input, the first layer uses the weight matrix $\mathbf{W}_A^{(1)} \in \mathbb{R}^{d_1 \times u}$ and bias vector $\mathbf{b}_A^{(1)}\in\mathbb{R}^{d_1}$ to map $\mathbf{x}_k$ to the output $\mathbf{V}_{k}^{(1)} \in \mathbb{R}^{d_1}$ with $d_1<u$. The $\mathbf{V}_k^{(1)}$ is subsequently used as the input to the next layer. 
For simplicity, with the $d_M$ hidden units in the $M$-th layer, we denote the \anet as 
\begin{equation}
\Theta_A: \mathbb{R}^{u} \mapsto \mathbb{R}^{d_{M}}
\end{equation}
$\Theta_A$ is parameterized by $\mathbf{W}_A$ and $\mathbf{b}_A$, where $\mathbf{W}_A = \left(\mathbf{W}_A^{(1)}, \cdots, \mathbf{W}_A^{(M)}\right)$ and $\mathbf{b}_A = \left(\mathbf{b}_A^{(1)}, \cdots, \mathbf{b}_A^{(M)}\right)$.

The \snet is designed to learn the dependencies between items in the input sequences. \snet takes advantage of long short-term memory (LSTM)~\cite{hochreiter1997long} network to learn both long and short-term dependencies within the sequences.  

Given a sequence $\mathbf{S}_k \in \mathbb{R}^{T\times r}$ as the input, we have the parameters within \snet for each time step $t$ as
\begin{equation}
  \begin{split}
  \label{eq-lstm}
  \mathbf{i}_k^{(t)} &= \sigma\left(\mathbf{W}_{i}\vec{\alpha}_k^{(t)} + \mathbf{U}_{i}\mathbf{h}_k^{(t-1)} + \mathbf{b}_i\right) \\
  \mathbf{f}_k^{(t)} &= \sigma\left(\mathbf{W}_{f}\vec{\alpha}_k^{(t)} + \mathbf{U}_{f}\mathbf{h}_k^{(t-1)} + \mathbf{b}_f\right) \\
  \mathbf{o}_k^{(t)} &= \sigma\left(\mathbf{W}_{o}\vec{\alpha}_k^{(t)} + \mathbf{U}_{o}\mathbf{h}_k^{(t-1)} + \mathbf{b}_o\right) \\
  \mathbf{g}_k^{(t)} &= \tanh\left(\mathbf{W}_{c}\vec{\alpha}_k^{(t)} + \mathbf{U}_{c}\mathbf{h}_k^{(t-1)} + \mathbf{b}_c\right) \\
  \mathbf{c}_k^{(t)} &= \mathbf{f}_k^{(t)}\odot\mathbf{c}_k^{(t-1)} + \mathbf{i}_k^{(t)} \odot \mathbf{g}_k^{(t-1)} \\
  \mathbf{h}_{k}^{(t)} &= \mathbf{o}_k^{(t)} \odot \tanh\left(\mathbf{c}_k^{(t)}\right)
  \end{split}
\end{equation}
where $\sigma(z)=\frac{1}{1+e^{-z}}$ is a \texttt{sigmoid} activation function, $\odot$ is the bitwise multiplication, $\mathbf{i}_k^{(t)}$, $\mathbf{f}_k^{(t)}$ and $\mathbf{o}_k^{(t)}$ are the internal gates of the LSTM, $\mathbf{c}_k^{(t)}$ and $\mathbf{h}_k^{(t)}$ are the cell and hidden states of the LSTM. For simplicity, we denote the \snet with $d_S$ hidden units as 
\begin{equation}
\Theta_S:\mathbb{R}^{T \times r} \mapsto \mathbb{R}^{d_S}
\end{equation}
$\Theta_S$ is parameterized by bias vector set $\mathbf{b}_S=\left(\mathbf{b}_i,\mathbf{b}_f,\mathbf{b}_o,\mathbf{b}_c\right)$ and the set of weight matrices $\mathbf{W}_S = \{\mathbf{W}_{\theta_s}, \mathbf{U}_{\theta_s}\}$, where $\mathbf{W}_{\theta_s}\!=\!\left(\mathbf{W}_{i}, \mathbf{W}_{f}, \mathbf{W}_{o}, \mathbf{W}_{c}\right)\in \mathbb{R}^{4\times d_S \times r}$ and $\mathbf{U}_{\theta_s} = \left(\mathbf{U}_{i}, \mathbf{U}_{f}, \mathbf{U}_{o}, \mathbf{U}_{c}\right)\in \mathbb{R}^{4\times d_S \times d_S}$. 

\subsection{FusionNet}
\label{subsection-fusionnet}
Next, we design the \fnet to integrate \anet and \snet into one network. Here we propose three designs: (1) balanced, (2) \anetns-centric and (3) \snetns-centric. 

\begin{figure}[!ht]
    \centering
    \begin{subfigure}{0.4\linewidth}
    \centering
        \includegraphics[width=1\textwidth]{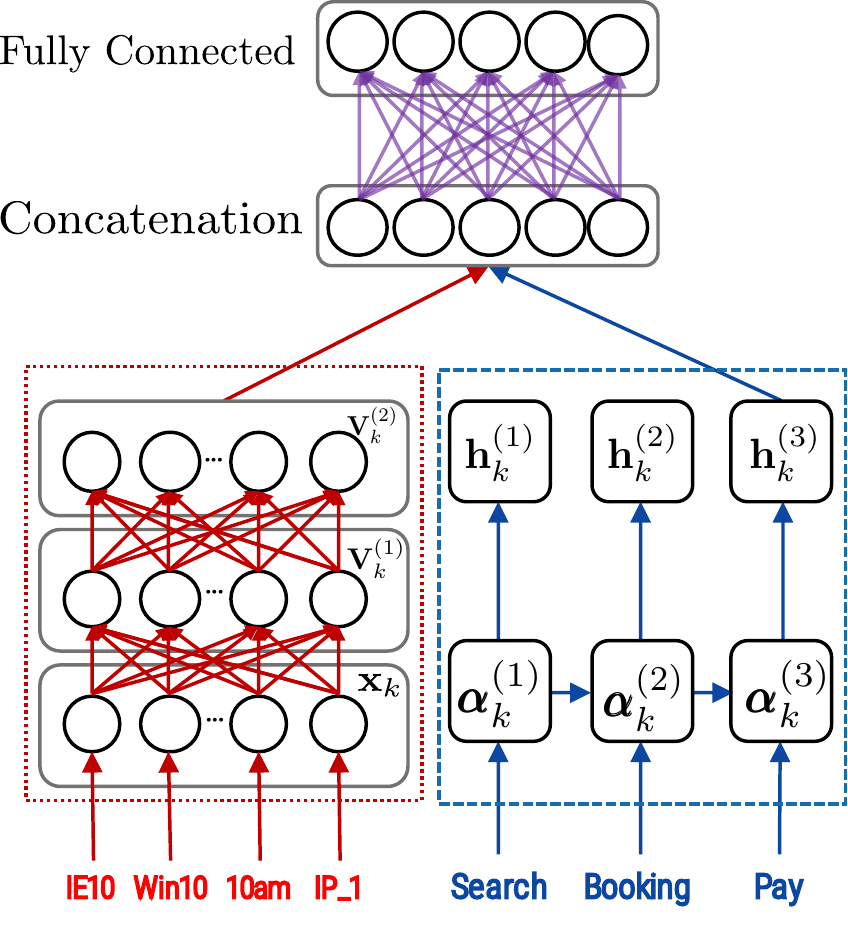}
        \captionsetup{format=centercaption}
        \caption{Balanced Design}
        \label{fig-balanced}
    \end{subfigure}
    \hspace{2mm}
    \begin{subfigure}{0.4\linewidth}
        \begin{subfigure}{1\linewidth}
          \centering
          \includegraphics[width=1\textwidth]{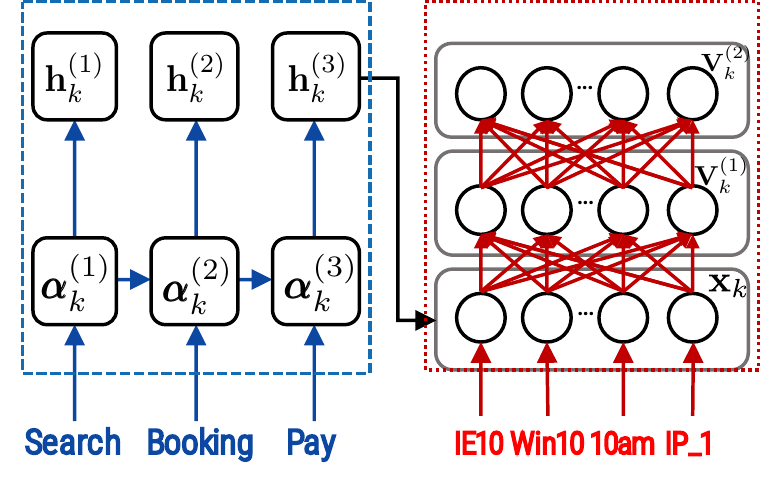}
          \captionsetup{format=centercaption}
          \caption{\anetns-centric Design}
          \label{fig-att-centric}
        \end{subfigure}
        \begin{subfigure}{1\linewidth}
          \centering
          \includegraphics[width=1\textwidth]{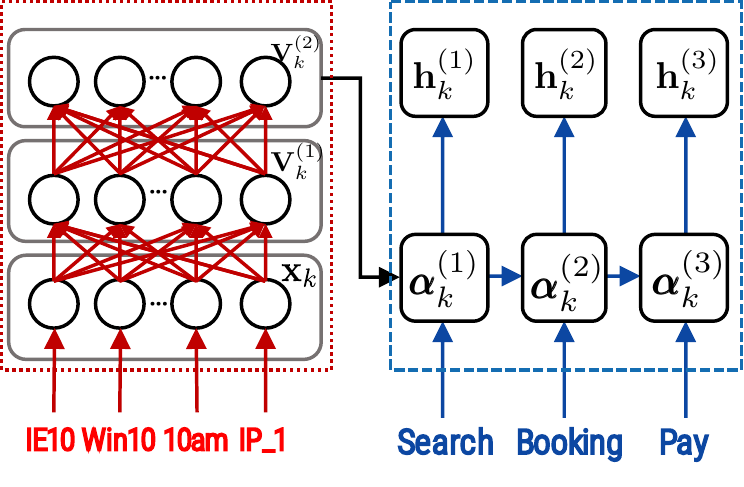}
          \captionsetup{format=centercaption}
          \caption{\snetns-centric Design} 
          \label{fig-seq-centric}
        \end{subfigure}
    \end{subfigure}
    \caption{Comparison of three different network designs. }
    \label{fig-three-designs}
\end{figure}

\noindent $\bullet$ \textbf{Balanced Design (Figure~\ref{fig-balanced}).} Both the attribute and sequence of each attributed sequence are processed simultaneously, and the results are concatenated together. The \anet and \snet are first concatenated, followed by a fully connected layer with a nonlinear function over the concatenation to capture the dependencies between attributes and sequences. The output of this fully connected layer corresponds to the output of \snet after processing the last time step in the sequence input. We denote the balanced design as 
\begin{align}
    \mathbf{y}_k &= \mathbf{V}_k^{(M)}\oplus \mathbf{h}_k^{(T_k)}\\
    \mathbf{z}_k &= \delta(\mathbf{W}_z\mathbf{y}_k + \mathbf{b}_z)
\end{align}
where $\oplus$ represents the concatenation operation, $\mathbf{W}_z\in\mathbb{R}^{d\times(d_M+d_S)}$ and $\mathbf{b}_z\in\mathbb{R}^{d}$ denote the weight matrix and bias vector in this fully connected layer, respectively. 

\noindent $\bullet$ \textbf{\anetns-centric Design (Figure~\ref{fig-att-centric}).} Here, the sequence is first transformed by sequence network, \ie, the function $\Theta_S$, and then used as an input of the attribute network, \ie, the function $\Theta_A$. Specifically, we modify Eq.~\ref{eq-att-network} to incorporate sequence representation as an input. We use the output of \snet after processing the last time step in the sequence input. The modified $\mathbf{V}_k^{(1)}$ is written as 
\begin{equation}
    \mathbf{V}_k^{(1)} = \delta(\mathbf{W}_A^{(1)}(\mathbf{x_k}\oplus\mathbf{h}_k^{(T_k)})+\mathbf{b}_A^{(1)})
\end{equation}
where the $\mathbf{W}_A^{(1)}\in \mathbb{R}^{d_1\times(u+d_S)}$ and $\mathbf{b}_A^{(1)}\in\mathbb{R}^{d_1}$. 

\noindent $\bullet$ \textbf{\snetns-centric Design (Figure~\ref{fig-seq-centric}).} The attribute vector is first transformed by $\Theta_A$ and then used as an additional input for $\Theta_S$. Specifically, we modify Eq.~\ref{eq-lstm} to integrate attribute representations at the first time step as an input. The modified $\mathbf{h}_k^{(t)}$ is 
\begin{equation}
\mathbf{h}_{k}^{(1)} = \mathbf{o}_k^{(1)} \odot \tanh(\mathbf{c}_k^{(1)}) + \mathbf{V}_k^{(M)}
\end{equation}
In order to fuse \anet and \snet using the \snetns-centric design, the dimension of $\mathbf{V}_k^{(M)}$ has to be the same as $\mathbf{o}_k^{(1)}$ and $\mathbf{c}_k^{(1)}$. That is, $d_S = d_M$.

Without loss of generality, we summarize the above three designs as
\begin{equation}
\label{eq-func-theta}
\Theta: (\mathbb{R}^{u}, \mathbb{R}^{T \times r}) \mapsto \mathbb{R}^{d}
\end{equation}

\subsection{MetricNet}
\label{subsection-metricnet}

Without loss of generality, we present the \mnet using the proposed balanced design due to space limitations. In the balanced design (as shown in Figure~\ref{fig-balanced}), the explicit form of Eq.~\ref{eq-func-theta} can be written as
\begin{equation}
\label{eq-balanced-specific}
    \Theta(J_k) = \Theta_A(\Theta_A(\mathbf{x}_k)\oplus\Theta_S(\mathbf{S}_k))
\end{equation}

Given an attributed sequence feedback instance $(p_i, p_j, \ell_{ij})$, where $p_i = (\mathbf{x}_i, \mathbf{S}_i)$ and $p_j = (\mathbf{x}_j, \mathbf{S}_j)$, $\ell_{ij} \in \{1, 0\}$. This input is transformed to $\Theta(p_i) \in \mathbb{R}^d$ and $\Theta(p_j)\in \mathbb{R}^d$ by the nonlinear transformation $\Theta$. 

The \mnet is designed using a contrastive loss function~\cite{hadsell2006dimensionality} so that attributed sequences in each similar pair in $\mathcal{S}$ have a smaller distance compared to those in $\mathcal{D}$ after learning the distance metric. The \mnet computes the Euclidean distance between each pair using the labels and back-propagates the gradients through all components in our network. The learning objective of \mnet can be written as
\begin{equation}
    \label{eq-constrastive-loss} 
L(p_i, p_j, l_{ij}) = \frac{1}{2}(1-\ell_{ij})(D_\Theta)^2 + \frac{1}{2}\ell_{ij}\{\max(0, g - D_\Theta)\}^2
\end{equation}
where $g$ is the margin parameter, meaning that the pairs with a dissimilar label ($\ell_{ij}=1$) contribute to the learning objective if and only if when the Euclidean distance between them is smaller than $g$~\cite{hadsell2006dimensionality}. 
We note that the \mnet can augment all three designs in the same way. Figure~\ref{fig-fdml} illustrates the \mnet with the proposed balanced design.

\section{Learning Feedback}
\label{section-learn-feedback}

\begin{figure}[!ht]
\centering
    \includegraphics[width=0.8\linewidth]{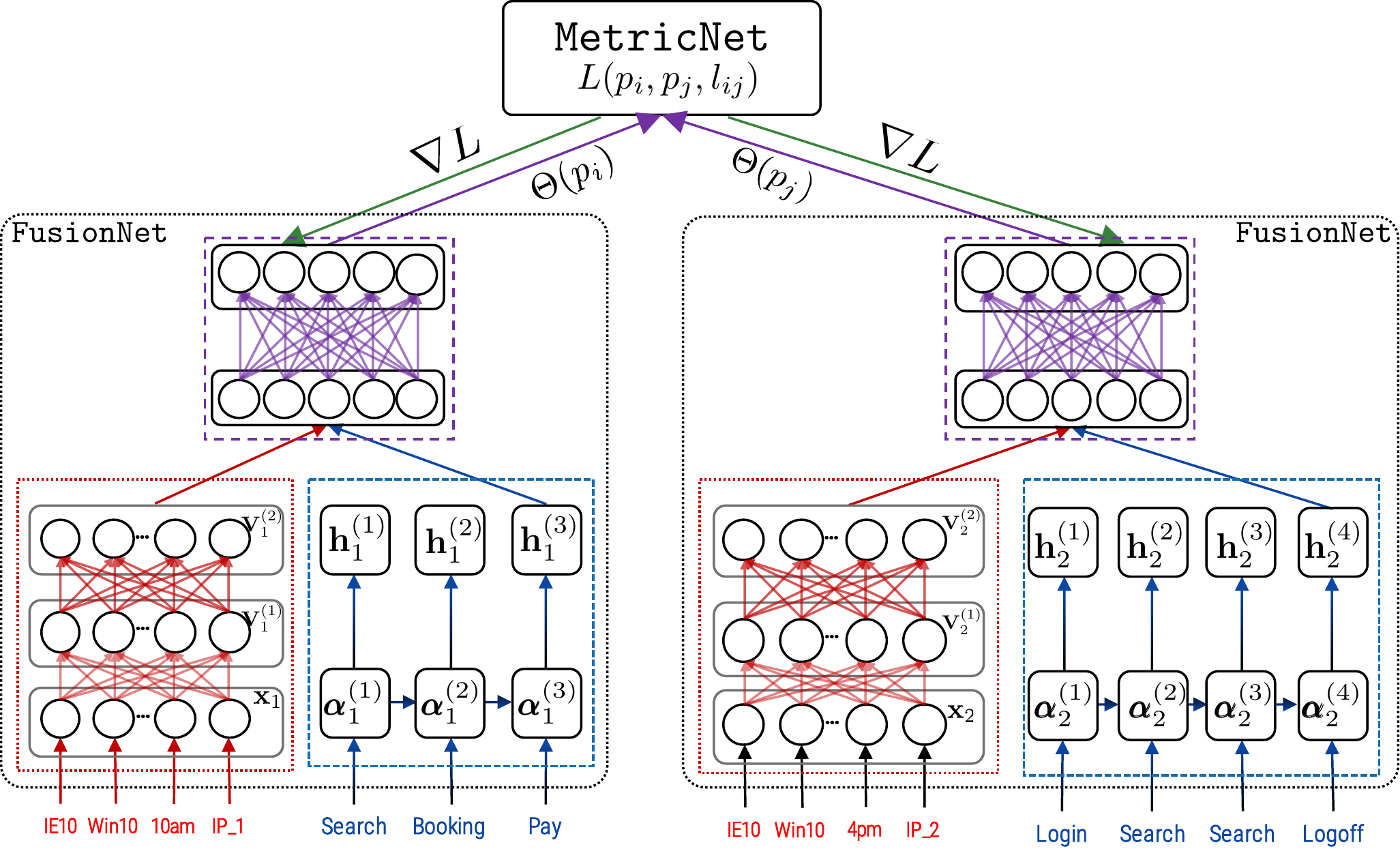}    
    \caption[MLAS Network with Balanced Design]{\mlas network with balanced design. Parameters in the two \fnet are identical. Gradient $\nabla L$ is used to update all layers.}
    \label{fig-fdml}
\end{figure}

In this section, we present the feedback learning mechanism of our \mlas network. Given two attributed sequences $p_i$ and $p_j$ as inputs, with Eq.~\ref{eq-balanced-specific}, we have 
\begin{equation}
\label{eq-nabla-l}
    \nabla L \equiv \left[\frac{\partial{L}}{\partial{\mathbf{W}_A}}, \frac{\partial{L}}{\partial{\mathbf{b}_A}}, \frac{\partial{L}}{\partial{\mathbf{W}_S}}, \frac{\partial{L}}{\partial{\mathbf{b}_S}}\right]
\end{equation}
where the explicit form can be written as
\begin{equation}
\nabla L = \frac{\partial{L}}{\partial{D_\Theta}}\frac{\partial D_\Theta}{\partial \Theta}
 \left[\frac{\partial \mathbf{V}_k^{(M)}}{\partial{\mathbf{W}_A}},\frac{\partial \mathbf{V}_k^{(M)}}{\partial{\mathbf{b}_A}}, \frac{\partial \mathbf{h}_k^{(T_k)}}{\partial{\mathbf{W}_S}}, \frac{\partial \mathbf{h}_k^{(T_k)}}{\partial{\mathbf{b}_S}}\right]
\end{equation}
where
\begin{equation}
\label{eq-l-theta}
\frac{\partial L}{\partial D_\Theta} = (1-\ell_{ij})D_{\Theta} - \ell_{ij}\max(0, g-D_{\Theta})
\end{equation}
\begin{equation}
\label{eq-d-theta}
\frac{\partial D_\Theta}{\partial \Theta} = \Big(\Theta(p_i) - \Theta(p_j)\Big) \cdot \Big(\mathds{1}-(\Theta(p_i) - \Theta(p_j))\Big)
\end{equation}
where $\mathds{1}$ is a vector filled with ones with the same shape as $\Theta(p_i)$ and $\Theta(p_j)$. 

For the $m$-th layer in \anetns, we employ the following update functions
\begin{equation}
    \begin{split}
        \frac{\partial \mathbf{V}_k^{(m)}}{\partial \mathbf{W}_A^{(m)}} &= \mathbf{V}_{k}^{(m)}\left(1-\mathbf{V}_{k}^{(m)}\right)\mathbf{V}_{k}^{(m-1)} \\
        \frac{\partial \mathbf{V}_k^{(m)}}{\partial \mathbf{b}_A^{(m)}} &= \mathbf{V}_{k}^{(m)}\left(1-\mathbf{V}_{k}^{(m)}\right) 
    \end{split}
\end{equation}

With the learning rate $\gamma$, the parameters $\mathbf{W}_A, \mathbf{W}_S, \mathbf{b}_A$ and $\mathbf{b}_S$ can be updated by the following equations until convergence: 
\vspace{-5pt}
\begin{equation}
\label{eq-update-all}
    \begin{split}
        \mathbf{W}_A &= \mathbf{W}_A - \gamma \frac{\partial L}{\partial \mathbf{W}_A} \\
        \mathbf{b}_A &= \mathbf{b}_A - \gamma \frac{\partial L}{\partial \mathbf{b}_A} \\
        \mathbf{W}_S &= \mathbf{W}_S - \gamma \frac{\partial L}{\partial \mathbf{W}_S} \\
        \mathbf{b}_S &= \mathbf{b}_A - \gamma \frac{\partial L}{\partial \mathbf{b}_S}
    \end{split}
\end{equation}

\begin{algorithm}[!ht]
    \begin{algorithmic}[1]
    \caption{\mlas Learning}
    \label{alg-fase}
    \INPUT A set of attributed sequences $\mathcal{J} = \{J_1, \cdots, J_n\}$, a set of feedback as pairwise attributed sequences $\mathcal{C} = \{(p_i, p_j, \ell_{ij})|p_i, p_j \in \mathcal{J}, \forall i, j = 1,\cdots, n, i\ne j\}$, the number of layers $M$, learning rate $\gamma$, number of iterations $\vartheta$ and convergence error $\epsilon$. 
    \OUTPUT Parameter sets $\{\mathbf{W}_A, \mathbf{b}_A, \mathbf{W}_S, \mathbf{b}_S\}$.
    \State{Initialize \mlas network $\Theta$.}
    \ForEach{$\vartheta^{\prime} = 1, \cdots, \vartheta$}
        \ForEach{$(p_i, p_j, \ell_{ij}) \in \mathcal{C}$}
            \State{//Forward propagation.}
            \State{Calculate $\Theta(p_i)$ and $\Theta(p_j)$.}
            \State{Calculate $D_\Theta$ using Eq.~\ref{eq-euclidean}.}
            \State{Calculate loss $L_{\vartheta^{\prime}}(p_i, p_j, \ell_{ij})$ according to Eq.~\ref{eq-constrastive-loss}.}
            \If{$|L_{\vartheta^{\prime}}(p_i, p_j, \ell_{ij}) - L_{\vartheta^{\prime}-1}(p_i, p_j, \ell_{ij})| < \epsilon$}
                \State{\textbf{break}}
            \EndIf
            \State{//Back-propagation.}
            \State{Calculate $\frac{\partial L}{\partial \Theta}$ according to Eq.~\ref{eq-l-theta},~\ref{eq-d-theta}.}
            \State{Calculate $\nabla L$ according to Eq.~\ref{eq-nabla-l}.}
            \State{Use Eq.~\ref{eq-update-all} to update network parameters.}
        \EndFor
    \EndFor
    \end{algorithmic}
\end{algorithm}

We summarize the algorithms for updating the \mlas network in Algorithm~\ref{alg-fase}.

\section{Experiments}
\subsection{Datasets}
We evaluate the proposed methods using four real-world datasets. Two of them are derived from application log files\footnote{Personal identity information is not collected. } at Amadeus~\cite{amadeus} (denoted as AMS-A and AMS-B). The other two datasets are derived from the Wikispeedia~\cite{west2009wikispeedia} dataset (denoted as Wiki-A and Wiki-B). 

\begin{itemize}
    \item \textbf{AMS-A}: We extracted $\sim$58k user sessions from log files of an internal application from our Amadeus. This internal application from Amadeus has motivated this research. Each record is composed of a user profile containing information ranging from system configurations to office name, and a sequence of functions invoked by click activities on the web interface. There are 288 distinct functions, 57,270 distinct user profiles in this dataset. The average length of the sequences is 18. 100 attributed sequence feedback pairs were selected by domain experts. 
    \item \textbf{AMS-B}: There are $\sim$106k user sessions derived from internal application log files with 575 distinct functions and 106,671 distinct user profile. The average length of the sequences is 22. 84 attributed sequence feedback pairs were selected by domain experts. 
    \item \textbf{Wiki-A}: This dataset is sampled from Wikispeedia dataset. Wikispeedia dataset originated from an online computation game~\cite{west2009wikispeedia}, in which each user is given two pages ({\ie}, source, and destination) from a subset of Wikipedia pages and asked to navigate from the source to the destination page. We use a subset of $\sim$2k paths from Wikispeedia with the average length of the path as 6. We also extract 11 sequence context as attributes (\eg, the category of the source page, average time spent on each page, \etc). There are 200 feedback instances selected based on the criteria of frequent subsequences and attribute value. 
    \item \textbf{Wiki-B}: This dataset is also sampled from Wikispeedia dataset. We use a subset of $\sim$1.5k paths from Wikispeedia with the average length of the path as 8. We also extract 11 sequence context (\eg, the category of the source page, average time spent on each page, \etc) as attributes. 220 feedback instances have been selected based on the criteria of frequent subsequences and attribute value. 
\end{itemize}
\begin{table}[t]
  \caption{Summary of Compared Methods}
  \label{table-compared-methods}
  \centering
  \begin{tabular}{ccp{2.25in}p{2.3cm}}
    \toprule
    Method & {Data Used} & {Short Description} & {Reference} 
    \\ \midrule
    \multirow[c]{2}{*}{\texttt{ATT}} & \multirow[c]{2}{*}{Attributes} & \multirow[c]{2}{*}{\shortstack[l]{Only \textit{attribute} feedback \\ is used in the model.}} & \multirow[c]{2}{*}{\cite{cvpr-face-verify}}
    \\ & & &
    \\ \midrule
    \multirow[c]{2}{*}{\texttt{SEQ}} & \multirow[c]{2}{*}{Sequences} & \multirow[c]{2}{*}{\shortstack[l]{Only \textit{sequence} feedback \\ is used in the model.}}& \multirow[c]{2}{*}{\cite{mueller2016siamese}}
    \\ & & &
    \\ \midrule
    \multirow[c]{3}{*}{\texttt{ASF}} & \multirow[c]{3}{*}{\shortstack[c]{Attributes \\ Sequences}} & \multirow[c]{3}{*}{\shortstack[l]{Feedback of attributes and \\sequences are used to train \\ two models \textbf{\textit{separatedly}}. }} & \multirow[c]{3}{*}{\cite{cvpr-face-verify} + \cite{mueller2016siamese}}
    \\ & & &
    \\ & & &
    \\ \midrule
    \multirow[c]{3}{*}{\mlasns\texttt{-B}} & \multirow[c]{3}{*}{\shortstack[c]{Attributes \\ Sequences \\ Dependencies}} &\multirow[c]{3}{*}{\shortstack[l]{Balanced design using attri- \\ -buted sequence feedback \\ to train one \textbf{\textit{unified}} model. }} & \multirow[c]{3}{*}{This Work} 
    \\ & & &
    \\ & & &
    \\ \midrule
    \multirow[c]{4}{*}{\mlasns\texttt{-A}} & \multirow[c]{4}{*}{\shortstack[c]{Attributes \\ Sequences \\ Dependencies}} & \multirow[c]{4}{*}{\shortstack[l]{Attribute-centric design \\ using attributed sequence \\ feedback to train one \\ \textbf{\textit{unified}} model. }} & \multirow[c]{4}{*}{This Work} 
    \\ & & &
    \\ & & &
    \\ & & &
    \\ \midrule
    \multirow[c]{4}{*}{\mlasns\texttt{-S}} & \multirow[c]{4}{*}{\shortstack[c]{Attributes \\ Sequences \\ Dependencies}} & \multirow[c]{4}{*}{\shortstack[l]{Sequence-centric design \\ using attributed sequence \\ feedback to train one \\ \textbf{\textit{unified}} model. }} & \multirow[c]{4}{*}{This Work} 
    \\ & & &
    \\ & & &
    \\ & & &    
    \\ \bottomrule
  \end{tabular}
\end{table}

\subsection{Compared Methods} 

We validate the effectiveness of our proposed \mlas solution compared with state-of-the-art baseline methods. To well understand the advancements of the proposed methods, we use baselines that are working on only attributes (denoted as \texttt{ATT}) or sequences (denoted as \texttt{SEQ}), as well as methods without exploiting the dependencies between attributes and sequences (denoted as \texttt{ASF}). We summarize the compared methods and references in Table~\ref{table-compared-methods}. 
\begin{itemize}
  \item \underline{Att}ribute-only Feedback (\texttt{ATT})~\cite{cvpr-face-verify}: Only attribute feedback is used in this model. This model first transforms fixed-size input data into feature vectors, then learns the similarities between these two inputs. 
  \item \underline{Seq}uence-only Feedback (\texttt{SEQ})~\cite{mueller2016siamese}: Only sequence feedback is used in this model. This model utilizes a long short-term memory (LSTM) to learn the similarities between two sequences.
  \item \underline{A}ttribute and \underline{S}equence \underline{F}eedback (\texttt{ASF})~\cite{cvpr-face-verify} + \cite{mueller2016siamese}: This method is a combination of the \texttt{ATT} and \texttt{SEQ} methods as above, where the two networks are trained \textit{separately} using attribute feedback and sequence feedback, respectively. The feature vectors generated by the two models are then concatenated. 
  \item Balanced Network Design with Attributed Sequence Feedback (\mlasns\texttt{-B}): The balanced design model using attributed sequence feedback to train a {\textit{unified}} model.
  \item \anetns-centric Network Design with Attributed Sequence Feedback (\mlasns\texttt{-A}): The \anetns-centric design using attributed sequence feedback to train a {\textit{unified}} model. 
  \item \snetns-centric Network Design with Attributed Sequence Feedback (\mlasns\texttt{-S}): The \snetns-centric design using attributed sequence feedback to train a {\textit{unified}} model.
\end{itemize}

\subsection{Experimental Settings} 

\subsubsection{Network Initialization and Training} Initializing the network parameters is important for models using gradient descent based approaches~\cite{erhan2009difficulty}. The weight matrices $\mathbf{W}_A$ in $\Theta_A$ and the $\mathbf{W}_{S}$ in $\Theta_S$ are initialized using the uniform distribution~\cite{glorot2010understanding}, the biases $\mathbf{b}_A$ and $\mathbf{b}_S$ are initialized with zero vector $\pmb0$ and the recurrent matrix $\mathbf{U}_{S}$ is initialized using orthogonal matrix~\cite{saxe2013exact}. We use one hidden layer ($M=1$) for \anet and \texttt{ATT} in the experiments to make the training process easier.  

After that, we pre-train each compared method. Pre-training is an important step to initialize the neural network-based models~\cite{erhan2009difficulty}. Our pre-training uses the attributed sequences as the inputs for \fnetns, and use the generated feature representations to reconstruct the attributed sequence inputs. We also pre-train the \texttt{ATT} and \texttt{SEQ} networks in a similar fashion that reconstruct the respective attributes or sequences. 
We utilize $\ell_2$-regularization and early stopping strategy to avoid overfitting. Twenty percents of feedback pairs are used in the validation set. We choose \texttt{ReLU} activation function~\cite{nair2010rectified} in our \anet to accelerate the stochastic gradient descent convergence.

\subsubsection{Performance Evaluation Setting.}
We evaluate the performance by using the feature representations generated by each method for clustering tasks. The feature representations are generated through a forward pass. 

Clustering tasks have been widely used in distance metric learning work~\cite{xing2003distance,wang2015survey}. In this set of experiments, we use HDBSCAN~\cite{campello2015hierarchical} clustering algorithm. HDBSCAN is a deterministic algorithm, which gives identical output when using the same input. We measure the normalized mutual information (NMI)~\cite{mcdaid2011normalized} score. The maximum NMI score is 1. Specifically, we conduct the below two experiments:
\begin{enumerate}
  \item The effect of feedback. We compare the performance of the clustering algorithm using the feature representations generated by \fnet before and after incorporating the feedback. 
  \item The effect of varying parameters in the clustering algorithm. After the metric learning process, we evaluate the feature representations generated by all compared methods under various parameters of the clustering algorithm. 
\end{enumerate}

\subsubsection{Parameter Study Settings.} 
We first evaluate the effect of output dimensions (\ie, the dimension of the hidden layer), which not only affect the model size but also affects the performance of subsequent mining algorithms. 

The other parameter we evaluate is the relative importance of attribute data (denoted as $\omega_A$) in the attributed sequences. The pre-training phase is essential to gradient descent-based methods~\cite{erhan2009difficulty}. The relative importance of attribute data and sequence data are represented by the weights of $\Theta_A$ and $\Theta_S$, denoted as $\omega_A$ and $\omega_S$, where $\omega_A + \omega_S = 1$. The intuition is that with one data type more important, the other one becomes relatively less important. 

\subsection{Results and Analysis}
\subsubsection{Effect of Feedback.} 
\begin{figure}[!ht]
    \centering
    \begin{subfigure}[t]{0.35\linewidth}
      \centering
      \includegraphics[page=1, width=1.05\linewidth]{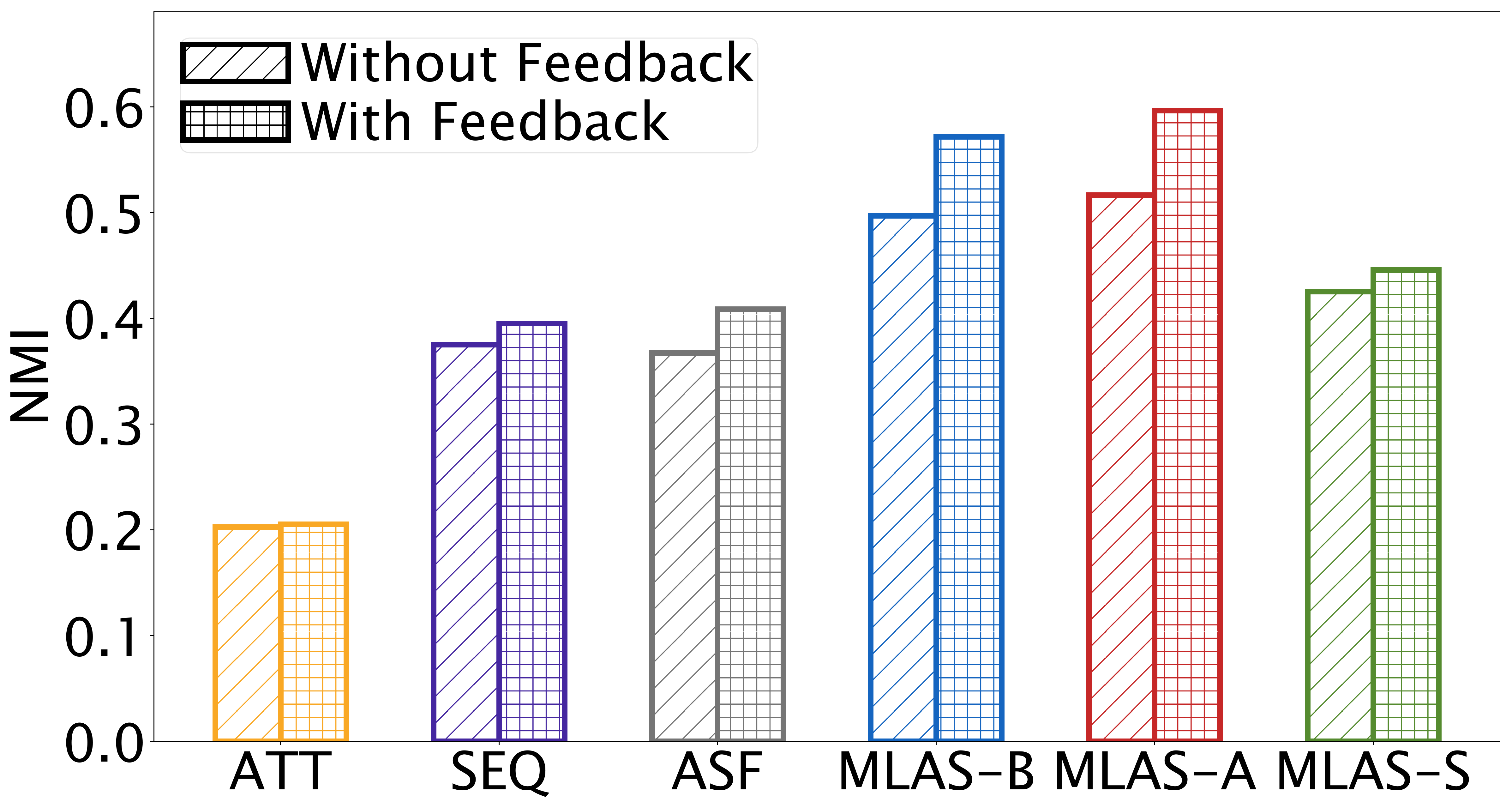}
      \label{fig-exp-fb1}      
      \caption{AMS-A Dataset}
    \end{subfigure}
    \hspace{5mm}
    \begin{subfigure}[t]{0.35\linewidth}
      \centering
      \includegraphics[page=2, width=1.05\linewidth]{figures/MLAS/exp/nmi_dim_bar}
      \label{fig-exp-fb2}
      \caption{AMS-B Dataset}
    \end{subfigure}

    \begin{subfigure}[t]{0.35\linewidth}
      \centering
      \includegraphics[page=3, width=1.05\linewidth]{figures/MLAS/exp/nmi_dim_bar}
      \label{fig-exp-fb3}
      \caption{Wiki-A Dataset}
    \end{subfigure}
    \hspace{5mm}
    \begin{subfigure}[t]{0.35\linewidth}
      \centering
      \includegraphics[page=4, width=1.05\linewidth]{figures/MLAS/exp/nmi_dim_bar}
      \label{fig-exp-fb4}
      \caption{Wiki-B Dataset}
    \end{subfigure}
  \caption[The effect of feedback in \mlas]{The effectiveness of using feedback. Using feedback could boost performance of all methods. The three methods we proposed (\mlasns\texttt{-B/A/S}) are capable of exploiting the information of attributes, sequences, and more importantly, the attribute-sequence dependencies to outperform other methods. }
  \label{fig-exp-feedback}
\end{figure}

We present the performance comparisons in clustering tasks using feature representations generated using the parameters of all methods in Figure~\ref{fig-exp-feedback}. Two sets of feature representations are generated, the first set is generated after the pre-training (denoted as \textit{without feedback}), the other set is generated after the metric learning step (denoted as \textit{with feedback}). We fix the output dimension to 10, minimum cluster size to 100 and $\omega_A=0.5$ (for \mlasns\texttt{-B/A/S}). We have observed that the feedback can boost the performance of all methods, and the three methods (\mlasns\texttt{-B/A/S}) proposed in this work are capable of outperforming other methods. Also, we also observe that the proposed three \mlas variations have better performance compared to the \texttt{ASF}, which also uses the information from attributes and sequences but \textit{without} using the attribute-sequence dependencies. 

Based on the above observations, we can conclude that the performance boost of our three architectures (\mlasns\texttt{-B/A/S}) is a result of taking advantages of attribute data, sequence data, and more importantly, the attribute-sequence dependencies. 

\subsubsection{Performance in Clustering Tasks.} 
\begin{figure}[!ht]
  \centering
  \includegraphics[width=0.8\linewidth]{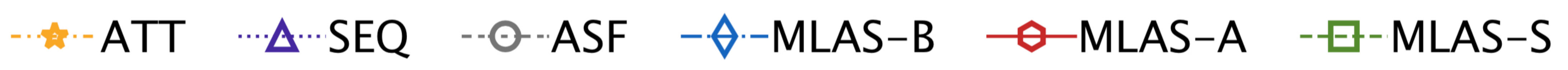}

    \begin{subfigure}[t]{0.35\linewidth}
      \centering
      \includegraphics[page=1, width=\linewidth]{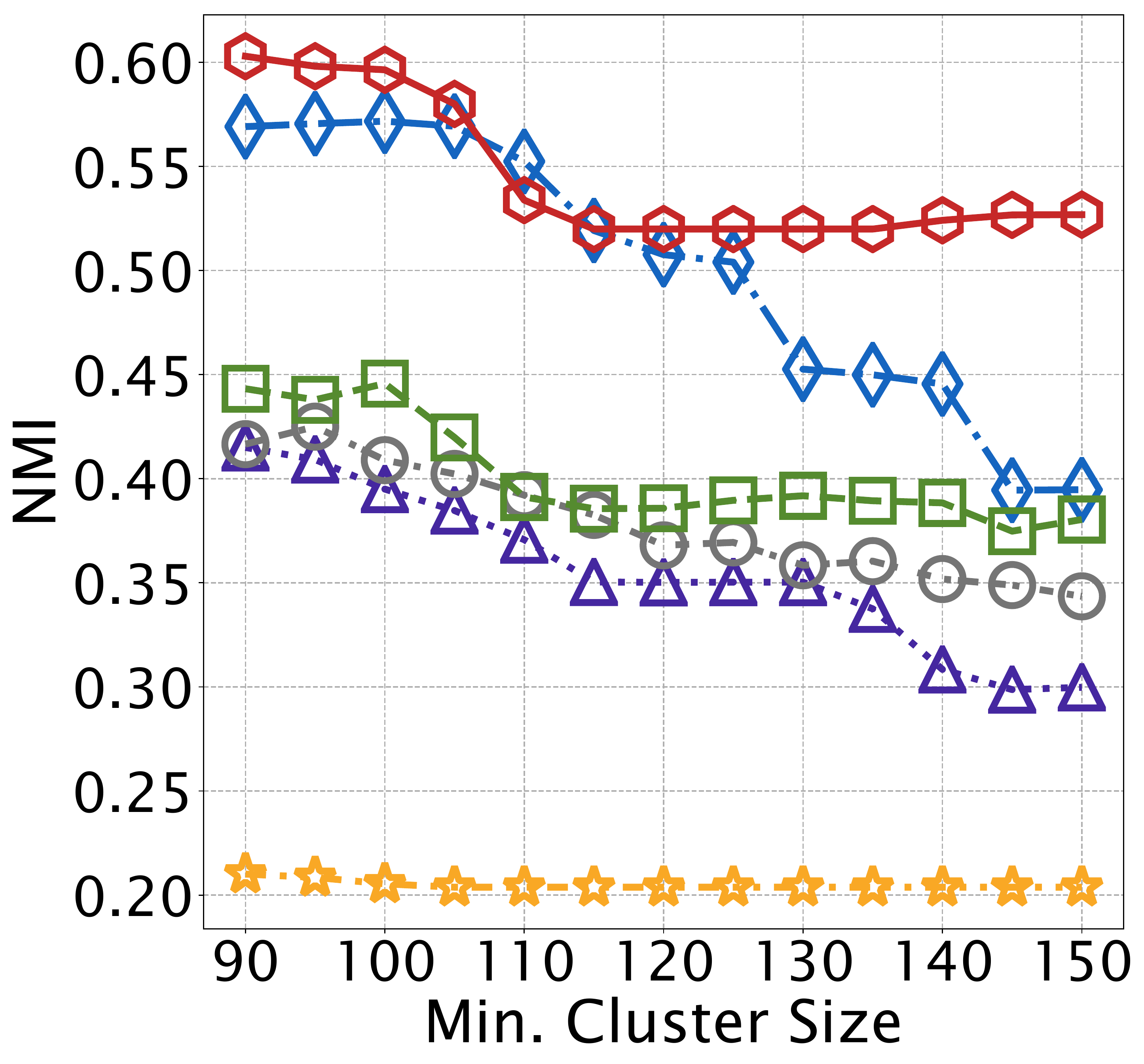}
      \label{fig-exp-cs1}
      \caption{AMS-A Dataset}
    \end{subfigure}
    \hspace{5mm}
    \begin{subfigure}[t]{0.35\linewidth}
      \centering
      \includegraphics[page=2, width=\linewidth]{figures/MLAS/exp/nmi_cs}
      \label{fig-exp-cs2}
      \caption{AMS-B Dataset}
    \end{subfigure}

    \begin{subfigure}[t]{0.35\linewidth}
      \centering
      \includegraphics[page=3, width=\linewidth]{figures/MLAS/exp/nmi_cs}
      \label{fig-exp-cs3}
      \caption{Wiki-A Dataset}
    \end{subfigure}
    \hspace{5mm}
    \begin{subfigure}[t]{0.35\linewidth}
      \centering
      \includegraphics[page=4, width=\linewidth]{figures/MLAS/exp/nmi_cs}
      \label{fig-exp-cs4}
      \caption{Wiki-B Dataset}
    \end{subfigure}
  \caption[\mlas performance with varying clustering parameters]{Performance with varying clustering parameters. Clustering results using the feature representations produced by \mlas are the best among the compared methods. }
  \label{fig-exp-clustering}
\end{figure}

\begin{figure}[!ht]
  \centering
  \includegraphics[width=0.8\linewidth]{figures/MLAS/exp/legend-6}
  
      \begin{subfigure}[t]{0.35\linewidth}
        \centering
        \includegraphics[page=1, width=\linewidth]{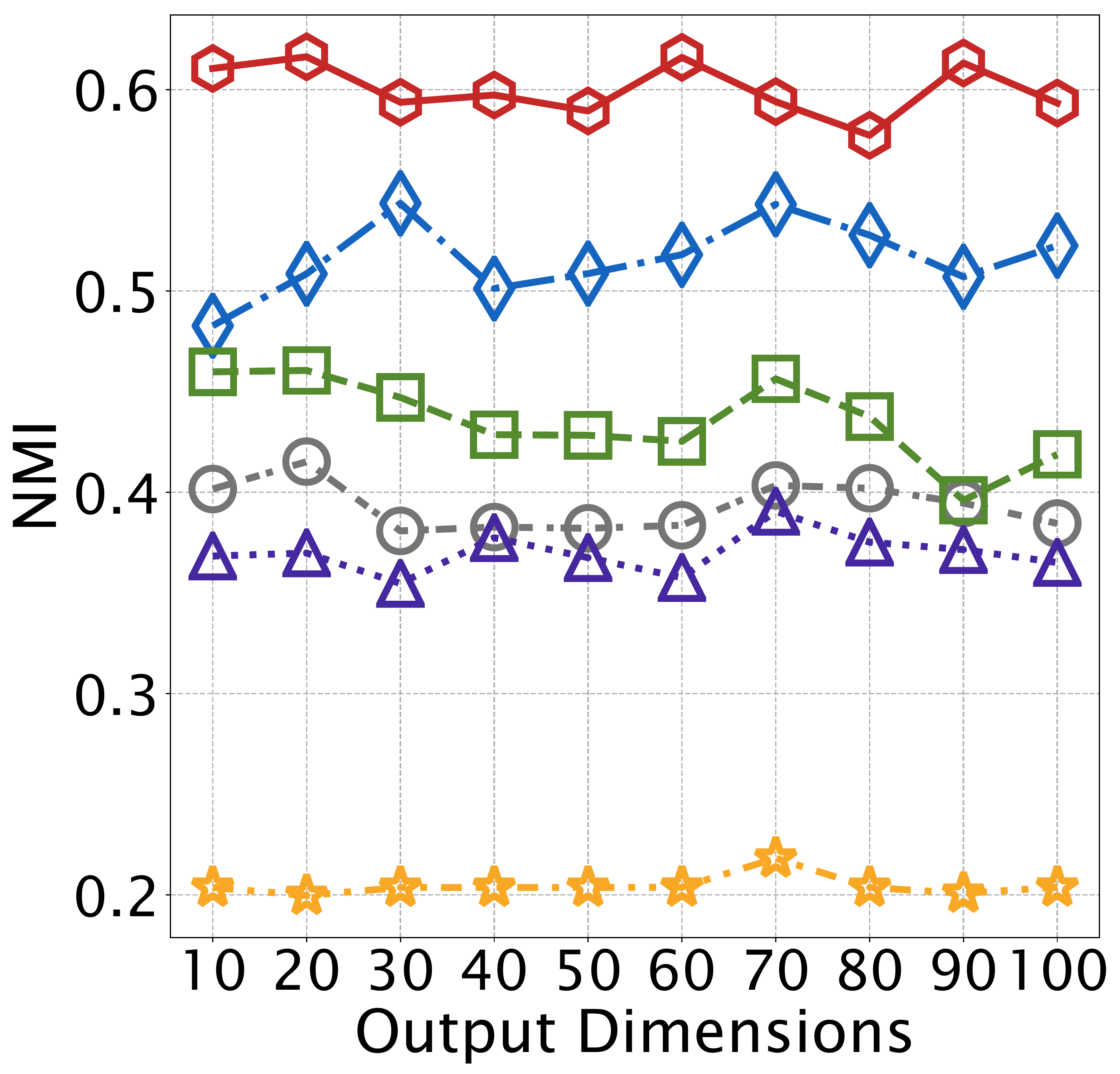}
        \label{fig-exp-dim1}
        \vspace{-5mm}
        \caption{AMS-A Dataset}
      \end{subfigure}
      \hspace{5mm}
      \begin{subfigure}[t]{0.35\linewidth}
        \centering
        \includegraphics[page=2, width=\linewidth]{figures/MLAS/exp/nmi_dim}
        \label{fig-exp-dim2}
        \vspace{-5mm}
        \caption{AMS-B Dataset}
      \end{subfigure}

      \begin{subfigure}[t]{0.35\linewidth}
        \centering
        \includegraphics[page=3, width=\linewidth]{figures/MLAS/exp/nmi_dim}
        \label{fig-exp-dim3}
        \vspace{-5mm}
        \caption{Wiki-A Dataset}
      \end{subfigure}
      \hspace{5mm}
      \begin{subfigure}[t]{0.35\linewidth}
        \centering
        \includegraphics[page=4, width=\linewidth]{figures/MLAS/exp/nmi_dim}
        \label{fig-exp-dim4}
        \vspace{-5mm}
        \caption{Wiki-B Dataset}
      \end{subfigure}
      \vspace{-3mm}
     \caption[\mlas parameter study with different network dimensions]{The effect of output dimensions (higher is better). Output dimension is an important factor for (1) the size of model; and (2) the cost of computations in downstream data mining tasks. Using the feature representations produced by \mlas can  constantly achieve the best performance among the compared methods. }
    \label{fig-exp-dimension}
    \vspace{-3mm}
  \end{figure}

\begin{figure}[!ht]
  \centering
  \includegraphics[width=0.5\linewidth]{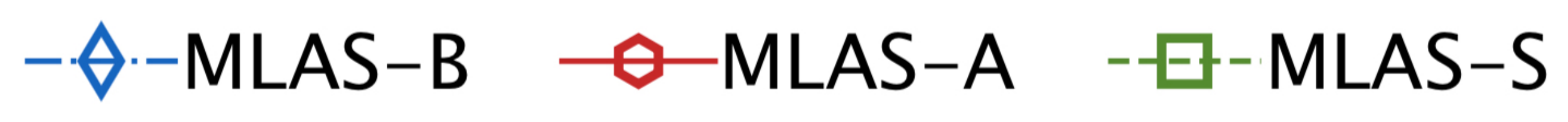}
  
      \begin{subfigure}[t]{0.35\linewidth}
        \centering
        \includegraphics[page=1, width=\linewidth]{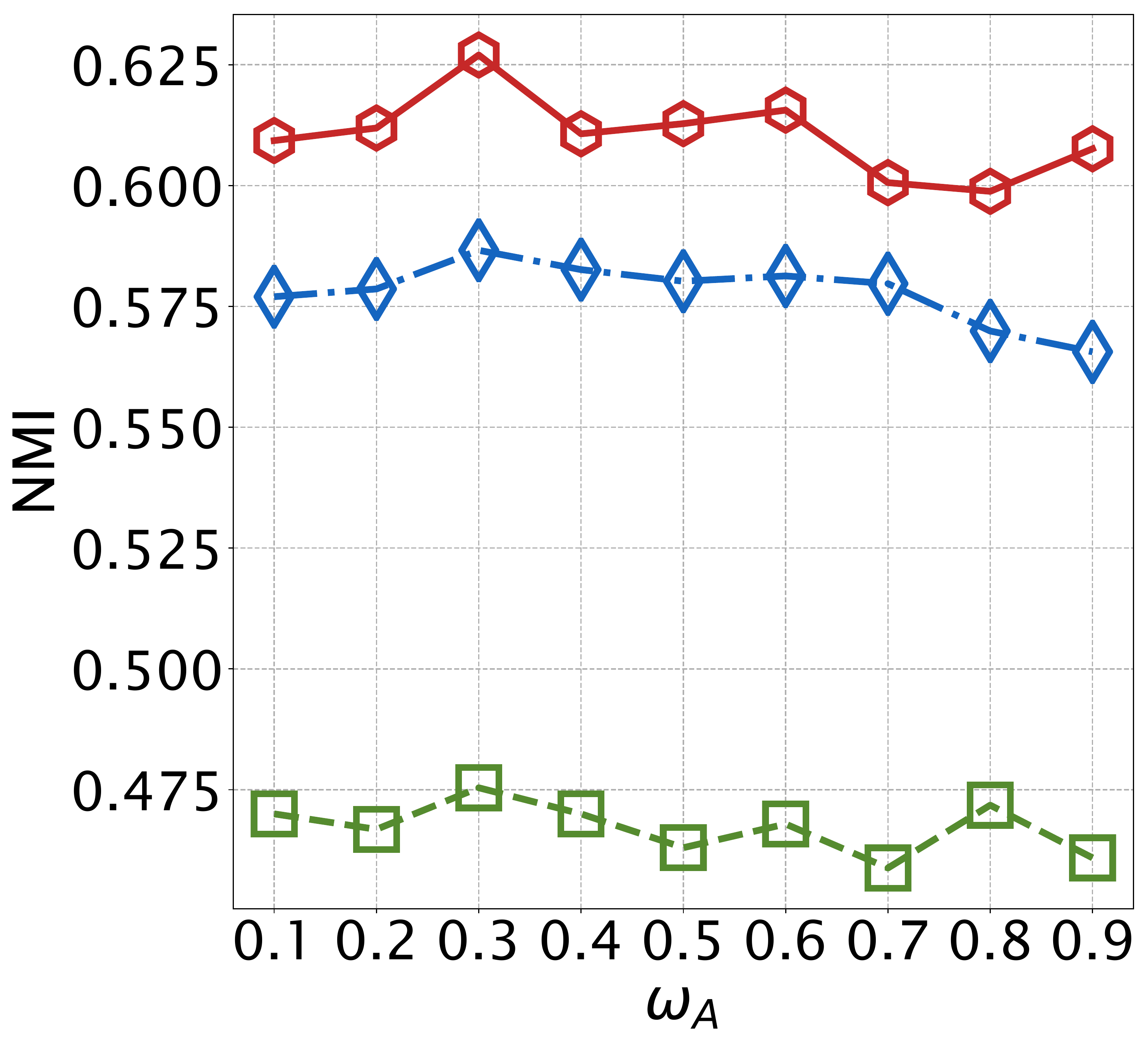}
        \label{fig-exp-w1}
        \vspace{-5mm}
        \caption{AMS-A Dataset}
      \end{subfigure}
      \hspace{5mm}
      \begin{subfigure}[t]{0.35\linewidth}
        \centering
        \includegraphics[page=2, width=\linewidth]{figures/MLAS/exp/nmi_w}
        \label{fig-exp-w2}
        \vspace{-5mm}
        \caption{AMS-B Dataset}
      \end{subfigure}

      \begin{subfigure}[t]{0.35\linewidth}
        \centering
        \includegraphics[page=3, width=\linewidth]{figures/MLAS/exp/nmi_w}
        \label{fig-exp-w3}
        \vspace{-5mm}
        \caption{Wiki-A Dataset}
      \end{subfigure}
      \hspace{5mm}
      \begin{subfigure}[t]{0.35\linewidth}
        \centering
        \includegraphics[page=4, width=\linewidth]{figures/MLAS/exp/nmi_w}
        \label{fig-exp-w4}
        \vspace{-5mm}
        \caption{Wiki-B Dataset}
      \end{subfigure}
      \vspace{-3mm}
     \caption[Effect of pre-training parameters in \mlas]{The effect of pre-training parameters in \mlasns. The pre-training parameter $\omega_A$ decides the relative importance of attributes in the model. We observe that \mlasns-\texttt{A} is capable of achieving the best performance on AMS-A and AMS-B datasets while \mlasns-\texttt{S} has the best performance on Wiki-A and Wiki-B datasets. }
    \label{fig-exp-wa}
    \vspace{-3mm}
  \end{figure}

\begin{figure}[!ht]
\centering
    \begin{subfigure}{\linewidth}
    \centering
      \includegraphics[width=\linewidth]{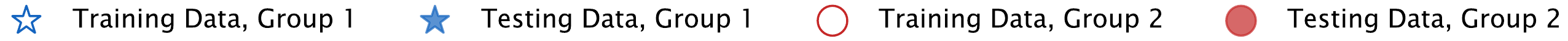}
    \end{subfigure}

    \begin{subfigure}[t]{0.16\linewidth}
      \centering
      \includegraphics[page=1, width=\linewidth]{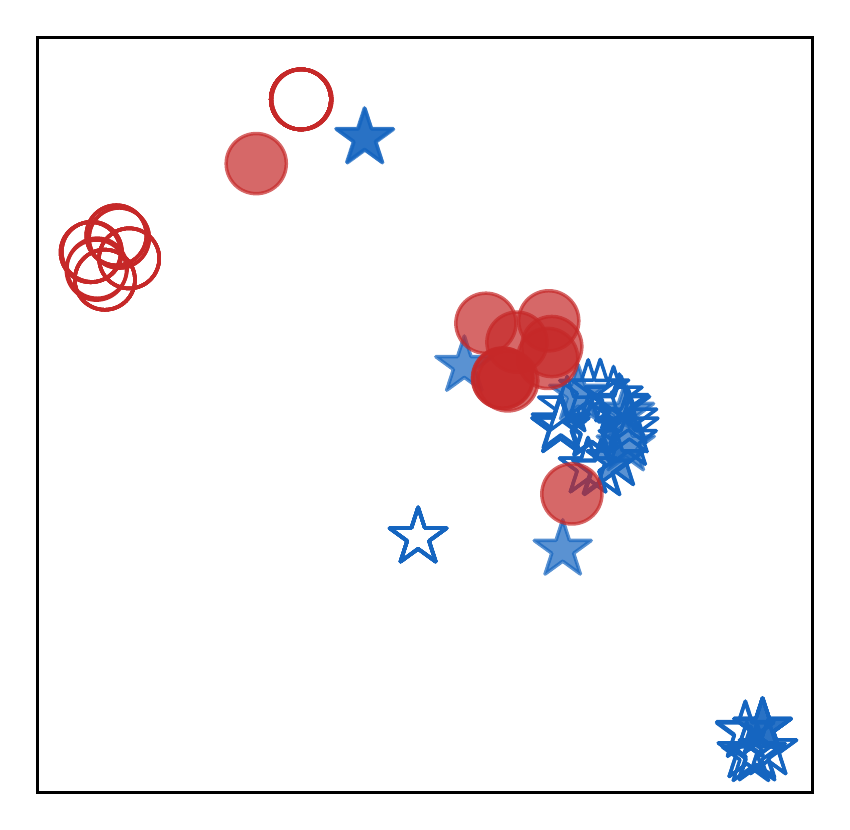}
      \captionsetup{format=centercaption}
      \caption{\texttt{ATT} \\ without Feedback}
      \label{fig-exp-tsne-attn}
      \end{subfigure}
	  \hfill
      \begin{subfigure}[t]{0.16\linewidth}
      \centering
      \includegraphics[page=3, width=\linewidth]{figures/MLAS/exp/tsne_plot}
        \captionsetup{format=centercaption}
      \caption{\texttt{SEQ} \\ without Feedback}
      \label{fig-exp-tsne-seqn}
      \end{subfigure}
    \hfill
    \begin{subfigure}[t]{0.16\linewidth}
      \centering
      \includegraphics[page=11, width=\linewidth]{figures/MLAS/exp/tsne_plot}
        \captionsetup{format=centercaption}
      \caption{\texttt{ASF} \\ without Feedback}
      \label{fig-exp-tsne-asfn}
      \end{subfigure}
      \hfill
      \begin{subfigure}[t]{0.16\linewidth}
      \centering
      \includegraphics[page=9, width=\linewidth]{figures/MLAS/exp/tsne_plot}
        \captionsetup{format=centercaption}
      \caption{\mlasns\texttt{-B} \\ without Feedback}
      \label{fig-exp-tsne-fcn}
      \end{subfigure}
    \hfill
    \begin{subfigure}[t]{0.16\linewidth}
      \centering
      \includegraphics[page=5, width=\linewidth]{figures/MLAS/exp/tsne_plot}
        \captionsetup{format=centercaption}
      \caption{\mlasns\texttt{-A} \\ without Feedback}
      \label{fig-exp-tsne-fan}
      \end{subfigure}
    \hfill
    \begin{subfigure}[t]{0.16\linewidth}
      \centering
      \includegraphics[page=7, width=\linewidth]{figures/MLAS/exp/tsne_plot}
        \captionsetup{format=centercaption}
      \caption{\mlasns\texttt{-S} \\ without Feedback}
      \label{fig-exp-tsne-fsn}
      \end{subfigure}
    \hspace{-2mm}
    \begin{subfigure}[t]{0.16\linewidth}
      \centering
      \reflectbox{\includegraphics[page=2, width=\linewidth]{figures/MLAS/exp/tsne_plot}}
      \captionsetup{format=centercaption}
      \caption{\texttt{ATT} \\ with Feedback}
      \label{fig-exp-tsne-attf}
      \end{subfigure}
    \hfill
    \begin{subfigure}[t]{0.16\linewidth}
      \centering
      \includegraphics[page=4, width=\linewidth]{figures/MLAS/exp/tsne_plot}
      \captionsetup{format=centercaption}
      \caption{\texttt{SEQ} \\ with Feedback}
      \label{fig-exp-tsne-seqf}
      \end{subfigure}
    \hfill
    \begin{subfigure}[t]{0.16\linewidth}
      \centering
      \includegraphics[page=12, width=\linewidth]{figures/MLAS/exp/tsne_plot}
      \captionsetup{format=centercaption}
      \caption{\texttt{ASF} \\ with Feedback}
      \label{fig-exp-tsne-asff}
      \end{subfigure}
    \hfill
    \begin{subfigure}[t]{0.16\linewidth}
      \centering
      \includegraphics[page=10, width=\linewidth]{figures/MLAS/exp/tsne_plot}
      \captionsetup{format=centercaption}
      \caption{\mlasns\texttt{-B} \\ with Feedback}
      \label{fig-exp-tsne-fcf}
    \end{subfigure}
    \hfill
    \begin{subfigure}[t]{0.16\linewidth}
      \centering
      \reflectbox{\includegraphics[page=6, width=\linewidth]{figures/MLAS/exp/tsne_plot}}
      \captionsetup{format=centercaption}
      \caption{\mlasns\texttt{-A} \\ with Feedback}
      \label{fig-exp-tsne-faf}
      \end{subfigure}
    \hfill
    \begin{subfigure}[t]{0.16\linewidth}
      \centering
      \includegraphics[page=8, width=\linewidth]{figures/MLAS/exp/tsne_plot}
      \captionsetup{format=centercaption}
      \caption{\mlasns\texttt{-S} \\ with Feedback}
      \label{fig-exp-tsne-fsf}
      \end{subfigure}
  \caption[Plots of \mlas feature space]{Plots of the feature representations. The \mlas is capable of exploiting the feedback and separating the instances from two different groups while keeping the instances from the same group together. }
  \label{fig-exp-tsne}
\end{figure}

The primary parameter in HDBSCAN is the minimum cluster size~\cite{campello2015hierarchical}, denoting the smallest set of instances to be considered as a group. 
Intuitively, while the minimum cluster size increases, each cluster may include instances that do not belong to it and the performance decreases. 
Figure~\ref{fig-exp-clustering} presents the results with the output dimension is 10 and $\omega_A = 0.5$. 

Compared to the best baseline method \texttt{ASF}, \mlasns-\texttt{A} achieves up to 18.3\% and 25.4\% increase of performance on AMS-A and AMS-B datasets, respectively. On Wiki-A and Wiki-B datasets, \mlasns-\texttt{S} is capable of achieving up to 26.3\% and 24.8\% performance improvement compared to \texttt{ASF}, respectively. 
We can further confirm that \mlas network is capable of exploiting the \textit{attribute-sequence dependencies} to improve the performance of the clustering algorithm with various parameter settings.

\subsubsection{Output Dimensions} 

We evaluate \mlas under a wide range of output dimension choices. The number of output dimensions relates to a variety of impacts, such as the usability of feature representations in downstream data mining tasks. In this set of experiments, we fix the minimum cluster size at 50, $\omega_A=0.5$ and vary output dimensions from 10 to 100. From Figure~\ref{fig-exp-dimension} we conclude that our proposed approaches outperform the baseline methods with various output dimensions. 

In particular, compared to the baseline method with the best performance, namely \texttt{ASF}, \mlasns-\texttt{A} achieves 20.7\% improvement on average on AMS-A dataset and 19.4\% improvement on average on the AMS-B dataset. When evaluated using Wiki-A and Wiki-B datasets, \mlasns-\texttt{S} outperforms \texttt{ASF} by 20.8\% and 10.6\% on average, respectively. 

\subsubsection{Pre-training Parameters} 
We evaluate \mlas under different pre-training parameters in this set of experiments. 
\texttt{ATT} and \texttt{SEQ} are not included in this set of experiments since they only utilize one data type. Output dimension is set to 5. Minimum cluster size is set at 50. 
Figure~\ref{fig-exp-wa} presents the results under different pre-training parameters. This confirms that our proposed \mlas method is not sensitive to different pre-training parameters. 

We notice the performance differences among the three \mlas architectures in the above experiments, where \mlasns-\texttt{A} has the best performance on AMS-A and AMS-B datasets, and \mlasns-\texttt{S} has the best performance on Wiki-A and Wiki-B datasets. We conclude this difference may relate to the datasets.

\subsection{Case Studies} In Figure~\ref{fig-exp-tsne}, we apply t-SNE~\cite{maaten2008visualizing} to the feature representations generated by all compared methods. The set of feature representations without feedback is generated after the pre-training phase and before the distance metric learning process. 

Our goal is to demonstrate the differences in the feature space of each method. 
We randomly select data points from both training and testing sets with a ground truth of two groups. We have the following findings: 
\begin{enumerate}
  \item The methods using either attribute data (\texttt{ATT}) or  sequence data (\texttt{SEQ}) \textit{only} cannot use the attributed sequence feedback. 
  \item The method using \textit{both} attributes and sequences \textit{separately} (\texttt{ASF}) is capable of better separating the two groups than the methods using single data type (\texttt{ATT} and \texttt{SEQ}). 
  \item Our methods using attributed sequence feedback as a unity to train unified models (\mlasns-\texttt{B/A/S}) are capable of separating the two groups the furthest, and thus achieve the best results. 
\end{enumerate}
These observations confirm that all three designs of \mlas can effectively learn the distance metric and result in better separation of two groups of data points.

 \newpage
\chapter{One-shot Learning on Attributed Sequences}
\label{chapter-task3}
\newpage

\section{Problem Definition} 
Inspired by the work in~\cite{bertinetto2016learning}, we formulate our problem as finding the parameters $\theta$ of a predictor $\Theta$ that minimizes the loss $\mathcal{L}_{\text{one-shot}}$. Given a training set of $g$ attributed sequences $\mathcal{G} = \{ (p_1, c_1), \cdots, (p_g, c_g) \}$, where each attributed sequence $p_i$ has a unique class label $c_i$, we formulate the objective for one-shot learning for attributed sequences as:
\begin{equation}
    \label{eq-problem-definition}
    \minimize_\theta \sum_{(p_i, c_i)\in \mathcal{G}} \mathcal{L}_{\text{one-shot}} \left( \Theta\left( p_i  ; \theta  \right), c_i \right)
\end{equation}
That is, we want to minimize the loss calculated using the label predicted using parameter $\theta$ and the true label. 
One-shot learning is known as a hard problem~\cite{koch2015siamese} mainly as a result of unavoidable overfitting caused by insufficient data. With a complex data type, such as attributed sequences, the number of parameters that need to be trained is even larger, which further complicates the problem. 

\begin{table}[t]
    \centering
    \normalsize
    \caption[\olas notations]{Important Mathematical Notations}
    \label{tab-notation}
    \begin{tabular}{cl}
        \hline
        Notation & Description \\ \hline
        $\mathbb{R}$ & The set of real numbers \\
        $r$ & The number of possible items in sequences. \\ 
        $s_i$ & A sequence of categorical items. \\
        $x_i^{(t)}$ & The $t$-th item in sequence $s_i$. \\
        $t_{\text{max}}$ & The maximum length of sequences in a dataset.\\ 
        $\mathbf{s}_i$ & A one-hot encoded sequence in the form of a matrix $\mathbf{s}_i  \in \mathbb{R}^{t_{\text{max}\times r}}$. \\
        $\textbf{x}_i^{(t)}$ & A one-hot encoded item at $t$-th time step in a sequence. \\
        $\mathbf{v}_i$ & An attribute vector. \\
        $p_i$ & An attributed sequence. \ie, $p_i = (\mathbf{v}_i, \mathbf{s}_i)$\\
        $\mathbf{p}_i$ & An $n$-dimensional feature vector of attributed sequence $p_i$. \\
        $\Omega$ & A function transforming each attributed sequence to a feature vector. \\
        $d$ & A distance function. \eg, Mahalanobis distance, Manhattan distance. \\
        $\gamma$ & An activation function within fully connected neural networks. \\ & Possible choices include \texttt{ReLU} and \texttt{tanh}. \\
        $\sigma$ & A logistic activation function within LSTM, \ie, $\sigma(z)=\frac{1}{1+e^{-z}}$ \\
        \hline
    \end{tabular}
\end{table}

\section{The \textsf{OLAS} Model}
\label{section-olas-model}
\subsection{Approach}
In this work, we adopt an approach from the distance metric learning perspective. Distance metric learning methods are well known for several important applications, such as face recognition, image classification, \etc. Distance metric learning is capable of disseminating data based on their dissimilarities using pairwise training samples. Recent work~\cite{koch2015siamese} has empirically demonstrated the effectiveness of the distance metric learning approach. 
In addition to the pairwise training samples, there are two key components in distance metric learning: a \textit{similarity} label depicting whether the training pair is similar and a distance function $d$. The similar and dissimilar pairs can be randomly generated using the class labels~\cite{koch2015siamese}. We define \textit{attributed sequence triplets} in Definition~\ref{def-triplets}. 
\begin{definition}[Attributed Sequence Triplets]
{\rm \label{def-triplets}
    An attributed sequence triplet $(p_i, p_j, \ell_{ij})$ consists of two attributed sequences $p_i, p_j$, and a \textit{similarity} label $\ell_{ij} \in \{0, 1\}$. The similarity label indicates whether $p_i$ and $p_j$ belong to the same class ($\ell_{ij} = 0$) or different classes ($\ell_{ij} = 1$). We denote $\mathcal{P} = \{(p_i, p_j, \ell_{ij}) | \ell_{ij}=0\}$ as the \textit{positive set} and  $\mathcal{N} = \{(p_i, p_j, \ell_{ij}) | \ell_{ij}=1\}$ as the \textit{negative set}.
    }
\end{definition}
However, attributed sequences are not naturally represented as feature vectors. Therefore, we define a transformation function $\Omega(p_i; \omega)$ parameterized by $\omega$ as a part of the predictor $\Theta$. $\Omega$ uses attributed sequences as the inputs and generates the corresponding feature vectors as the outputs. With two attributed sequences $p_i$ and $p_j$ as inputs, the $n$-dimensional feature vectors of the respective attributed sequences are:
\begin{equation}
\label{eq-two-attseq-vectors}
    \begin{split}        
        \mathbf{p}_i = \Omega(p_i ; \omega)\\
        \mathbf{p}_j = \Omega(p_j ; \omega)\\ 
        \mathbf{p}_i, \mathbf{p}_j \in \mathbb{R}^n
    \end{split}
\end{equation}


The other key component in distance metric learning approaches is a distance function (\eg, Mahalanobis distance ~\cite{cvpr-face-verify}, Manhattan distance~\cite{bertinetto2016learning}). A distance function is applied to the feature vectors in distance metric learning. 

Distance metric learning-based approaches often use the Mahalanobis distance~\cite{cvpr-face-verify,karpathy2015deep}, which can be equivalent to the Euclidean distance~\cite{cvpr-face-verify}. Using the two feature vectors of attributed sequences in Equation~\ref{eq-two-attseq-vectors}, the Mahalanobis distance can be written as: 
\begin{equation}
    \label{eq-mahalanobis-distance}
    d_\omega(\mathbf{p}_i, \mathbf{p}_j) = \sqrt{(\mathbf{p}_i - \mathbf{p}_j)^\top \mathbf\Lambda (\mathbf{p}_i - \mathbf{p}_j)}
\end{equation}
where $d_\omega$ is a specific form of distance function $d$ denoting the inputs (\ie, $\mathbf{p}_i, \mathbf{p}_j$) are the results of transformations using parameter $\omega$. $\mathbf\Lambda \in \mathbb{R}^{n \times n}$ is a symmetric, semi-definite, and positive matrix, and $\mathbf\Lambda$ can be decomposed as:
\begin{equation}
    \mathbf\Lambda = \mathbf\Gamma^\top\mathbf\Gamma,
\end{equation}
where $\mathbf\Gamma \in \mathbb{R}^{f \times n}, f\leq n$. By~\cite{xing2003distance}, Equation~\ref{eq-mahalanobis-distance} is equivalent to:
\begin{equation}
\label{eq-euclidean}
 \begin{split}
    d_\omega(\mathbf{p}_i, \mathbf{p}_j) & = \sqrt{(\mathbf{p}_i - \mathbf{p}_j)^\top \mathbf\Gamma^\top\mathbf\Gamma (\mathbf{p}_i - \mathbf{p}_j)} \\
        &= \| \mathbf\Gamma\mathbf{p}_i- \mathbf\Gamma\mathbf{p}_j\|_2.
 \end{split}
\end{equation}

Instead of directly minimizing the loss of the predictor function $\Theta$ predicting a label of each attributed sequence as in Equation~\ref{eq-problem-definition}, we can now achieve the same training goal by minimizing the loss of predicting whether a pair of attributed sequences belong to the same class using distance metric learning-based methods. The overall objective can be written as:
\begin{equation}
    \label{eq-problem-def}
    \minimize_{\omega} \sum_{(p_i, p_j, \ell_{ij})\in \mathcal{P}\cup\mathcal{N}} 
    \mathcal{L} \left(
    d_\omega\left(\mathbf{p}_i, \mathbf{p}_j\right), \ell_{ij}
    \right)
\end{equation}
In recent work on distance metric learning applications~\cite{bertinetto2016learning,cvpr-face-verify}, deep neural networks are serve as the nonlinear transformation function $\Omega$. Deep neural networks can effectively learn the features from input data without requiring domain-specific knowledge~\cite{koch2015siamese}, and also generalize the knowledge for future predictions and inferences. These advantages make neural networks become an ideal solution for one-shot learning. 

\subsection{\textsc{OLAS} Model Design}
We next describe the design of the two key components of the \olas model. First, we design a \cnet for the nonlinear transformation of attributed sequences. Then, a \vnet is designed to learn from the contrast of attributed sequences with different class labels. The specific parameters of the \olas used in our experiments are detailed in Section~\ref{section-olas-experiments}. 

\begin{figure}[t]
    \centering
        \includegraphics[width=0.82\linewidth]{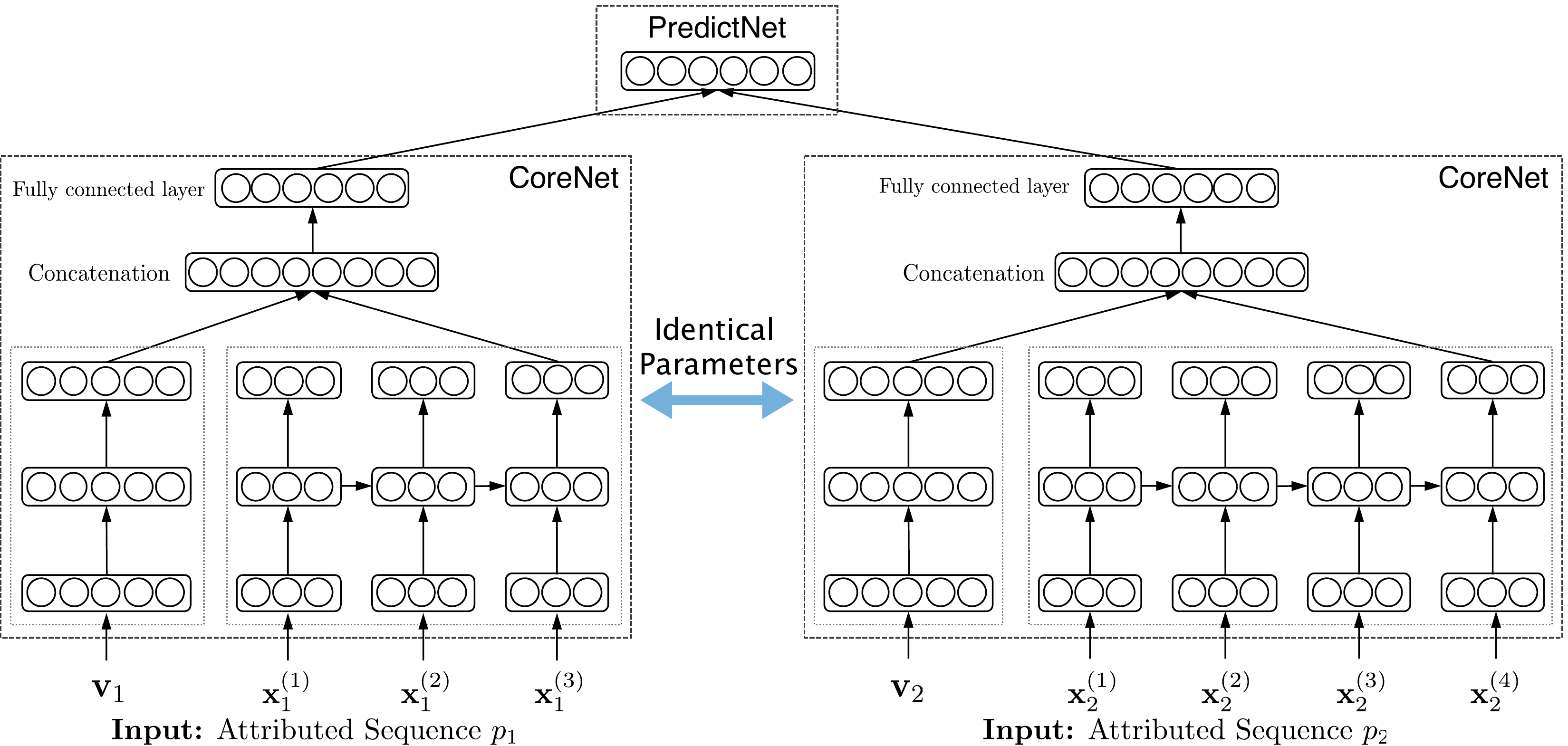}   
        \caption[\olas network structure]{The network architecture of \olasns. The concatenation only happens after the \textit{last time step} of the sequence so the information of the complete sequence is used. }
        \label{fig-fdml}
        \vspace{-5mm}
    \end{figure}
The two main networks in \cnetns, a fully connected neural network with $m$ layers and a long short-term memory (LSTM) network~\cite{hochreiter1997long}, correspond to the tasks of encoding the information from attributes and sequences in attributed sequences, respectively. By augmenting with another layer of fully connected neural network on top of the concatenation of the above networks, \cnet is also capable of learning the attribute-sequence dependencies.

Given the input of an attribute vector $\mathbf{v}_k\in\mathbb{R}^u$, we define a fully connected neural network with $m$ layers as: 
\begin{equation}
\begin{split}
\label{eq-att-network}
\pmb\upalpha_1 &= \gamma\left(\mathbf{W}_1\mathbf{v}_i + \mathbf{b}_1\right) \\
\pmb\upalpha_2 &= \gamma\left(\mathbf{W}_2\pmb\upalpha_1 + \mathbf{b}_2\right) \\
\vdots \\
\pmb\upalpha_m &= \gamma\left(\mathbf{W}_m\pmb\upalpha_{m - 1} + \mathbf{b}_m\right)
\end{split}
\end{equation}
where $\gamma$ is a nonlinear transformation function. Although we use hyperbolic tangent $\tanh$ in our model, other nonlinear functions such as rectified linear unit (ReLu)~\cite{nair2010rectified} can also be used depending on the empirical results. We denote the weights and bias parameters as: 
\begin{equation}
    \mathbf{W}_{\text{F}} = [\mathbf{W}_{1}, \cdots, \mathbf{W}_{m}]^\top, \mathbf{b}_{\text{F}} = [\mathbf{b}_{1}, \cdots, \mathbf{b}_{m}]^\top%
\end{equation}
Note that the choice of $m$ is task-specific. Although neural networks with more layers are better at learning hierarchical structure in the data, it is also observed that such networks are challenging to train due to the multiple nonlinear mappings that prevent the information and gradient passing along the computation graph~\cite{pham2017column}.

$\mathbf{W}_{\text{F}}$ and $\mathbf{b}_{\text{F}}$ are used to transform the input of each layer to a lower dimension. This transformation is imperative given the often large number of dimensions of attribute vectors in real-world applications. 
Different from attribute vectors, the categorical items in the sequences in attributed sequences obey temporal ordering. The information of sequences is not only in the item values, but more importantly, in the temporal ordering of these items. In this vein, the \cnet also utilizes an LSTM network. LSTM is capable of handling not only the ordering of items, but also the dependencies between different items in the sequences. Given a sequence $\mathbf{s}_i$ as the input, we use an LSTM ~\cite{hochreiter1997long} to process each item $\mathbf{x}_k^{(t)}$ in this sequence as: 
\begin{equation}
  \begin{split}
  \label{eq-lstm}
  \mathbf{i}^{(t)} &= \sigma\left(\mathbf{W}_{\text{i}}\mathbf{x}_k^{(t)} + \mathbf{U}_{\text{i}}\mathbf{h}^{(t-1)} + \mathbf{b}_{\text{i}}\right) \\[2pt]
  \mathbf{f}^{(t)} &= \sigma\left(\mathbf{W}_{\text{f}}\mathbf{x}_k^{(t)} + \mathbf{U}_{\text{f}}\mathbf{h}^{(t-1)} + \mathbf{b}_{\text{f}}\right) \\[2pt]
  \mathbf{o}^{(t)} &= \sigma\left(\mathbf{W}_{\text{o}}\mathbf{x}_k^{(t)} + \mathbf{U}_{\text{o}}\mathbf{h}^{(t-1)} + \mathbf{b}_{\text{o}}\right) \\[2pt]
  \mathbf{g}^{(t)} &= \tanh\left(\mathbf{W}_{\text{c}}\mathbf{x}_k^{(t)} + \mathbf{U}_{\text{c}}\mathbf{h}^{(t-1)} + \mathbf{b}_{\text{c}}\right) \\[2pt]
  \mathbf{c}^{(t)} &= \mathbf{f}^{(t)}\odot\mathbf{c}^{(t-1)} + \mathbf{i}^{(t)} \odot \mathbf{g}^{(t)} \\[2pt]
  \mathbf{h}^{(t)} &= \mathbf{o}^{(t)} \odot \tanh\left(\mathbf{c}^{(t)}\right)
  \end{split}
\end{equation}
where $\sigma$ is a sigmoid activation function, $\odot$ denotes the bitwise multiplication, $\mathbf{i}^{(t)}$, $\mathbf{f}^{(t)}$ and $\mathbf{o}^{(t)}$ are the internal gates of the LSTM, $\mathbf{c}^{(t)}$ and $\mathbf{h}^{(t)}$ are the cell and hidden states of the LSTM. Without loss of generality, we denote LSTM kernel parameters $\mathbf{W}_{\text{L}}$, recurrent parameters $\mathbf{U}_{\text{L}}$ and bias parameters $\mathbf{b}_{\text{L}}$ as:
\begin{equation}
    \begin{split}
        \mathbf{W}_{\text{L}} &= [\mathbf{W}_{\text{i}}, \mathbf{W}_{\text{f}}, \mathbf{W}_{\text{o}}, \mathbf{W}_{\text{c}}]^\top \\[-1pt]
        \mathbf{U}_{\text{L}} &= [\mathbf{U}_{\text{i}}, \mathbf{U}_{\text{f}}, \mathbf{U}_{\text{o}}, \mathbf{U}_{\text{c}}]^\top \\[-1pt]
        \mathbf{b}_{\text{b}} &= [\mathbf{b}_{\text{i}}, \mathbf{b}_{\text{f}}, \mathbf{b}_{\text{o}}, \mathbf{b}_{\text{c}}]^\top
    \end{split}
\end{equation}
The attribute vectors and sequences are processed simultaneously and the outputs of both networks are concatenated together. Instead of using the outputs of the LSTM at every time step, we only concatenate the last output from the LSTM to the output of the fully connected neural network so that the complete sequence information is used. 
After that, another layer of the fully connected neural network is used to capture the dependencies between attributes and sequences. 
Given the output dimensions of $\pmb\upalpha_m$ and $\mathbf{h}^{(t)}$ as $n_m$ and $n_l$, respectively, the concatenation and the last fully connected layer of \cnet can be written as: 
\begin{equation}
    \normalsize
    \mathbf{p}_i = \gamma\left(\mathbf{W}_{\text{p}}\left( \pmb\upalpha_m \oplus \mathbf{h}^{(t_i)} \right) + \mathbf{b}_{\text{p}}\right)
\end{equation}
where $\oplus$ represents the concatenation of two vectors, $\mathbf{W}_{\text{p}}\in\mathbb{R}^{n\times(n_m+n_{{l}})}$ and $\mathbf{b}_{\text{p}}\in\mathbb{R}^{n}$ denote the weight matrix and bias vector in this fully connected layer for an $n$-dimensional output. In summary, the \cnet can be written as: 
\begin{equation}
\label{eq-func-omega}
\Omega: \left(\mathbb{R}^{u}, \mathbb{R}^{t_{\text{max}} \times r}\right) \mapsto \mathbb{R}^{n}
\end{equation}

\begin{algorithm}[t]
    \normalsize
    \begin{algorithmic}[1]
    \caption{Training using attributed sequence triplets}
    \label{alg-fase}
    \INPUT A positive set $\mathcal{P}$ and a negative set $\mathcal{N}$ of attributed sequence triplets, the number of layers in fully connected neural networks $m$, learning rate $\lambda$, number of iterations $\phi$ and convergence error $\epsilon$. 
    \OUTPUT Parameters of \olas ($\{\mathbf{W}_{\text{F}}, \mathbf{b}_{\text{F}}, \mathbf{W}_{\text{L}}, \mathbf{U}_{\text{L}}, \mathbf{b}_{\text{L}}\}$).
    \State{Initialize \olas network.}
    \ForEach{$\phi^{\prime} = 1, \cdots, \phi$} {\Comment{\small $\phi$ is the maximum number of training epochs.}}
        \ForEach{$(p_i, p_j, \ell_{ij}) \in \mathcal{P}\cup\mathcal{N}$}
            \State{$\mathbf{p}_i \leftarrow \Omega(p_i; \omega)$.}
            \State{$\mathbf{p}_j \leftarrow \Omega(p_j; \omega)$.}
            \State{Compute $d_\omega$.} \Comment{Equation~\ref{eq-euclidean}.}
            \State{Compute the loss $\mathcal{L}_{\phi^{\prime}}(\mathbf{p}_i, \mathbf{p}_j, \ell_{ij})$.} \Comment{Equation~\ref{eq-contrastive-loss}.}
            \If{$|\mathcal{L}_{\phi^{\prime}}(\mathbf{p}_i, \mathbf{p}_j, \ell_{ij}) - \mathcal{L}_{\phi^{\prime}-1}(\mathbf{p}_i, \mathbf{p}_j, \ell_{ij})| < \epsilon$} 
                \State{\textbf{break}} \Comment{Early stopping to avoid overfitting. }
            \Else
                \State{Compute $\frac{\partial \mathcal{L}}{\partial d_\omega}, \frac{\partial d_\omega}{\partial \Omega}$. } \Comment{Equation~\ref{eq-l-omega},~\ref{eq-d-omega}.}
                \State{Compute $\nabla \mathcal{L}$.} \Comment{Equation~\ref{eq-nabla-l}.}
                \State{Update network parameters.} \Comment{Equation~\ref{eq-update-all}.}
            \EndIf
        \EndFor
    \EndFor
    \end{algorithmic}
\end{algorithm}
The two outputs of \cnet ($\mathbf{p}_i$ and $\mathbf{p}_j$) are first generated. Then, $\mathbf{p}_i$, $\mathbf{p}_j$ and the similarity label $\ell_{ij}$, are used by the \vnet to learn the similarities and differences between them. 
The \vnet is designed to utilize a contrastive loss function~\cite{hadsell2006dimensionality} so that attributed sequences in different categories are disseminated. 
The contrastive loss function is composed of two parts: a partial loss for the dissimilar pairs and a partial loss for similar pairs. 
The specific form of contrastive loss of \vnet can be written as:
\begin{equation}
\label{eq-contrastive-loss} 
\begin{split}
    \mathcal{L}(\mathbf{p}_i, \mathbf{p}_j, \ell_{ij}) = \underbrace{\frac{1}{2} \ell_{ij} \Big[\max \big( 0, \upxi - d_\omega(\mathbf{p}_i, \mathbf{p}_j)\big)\Big]^2}_{\text{\small Partial loss for \textit{dissimilar} pairs. }} \\[6pt]  +\underbrace{\frac{1}{2}(1-\ell_{ij})d_\omega^2(\mathbf{p}_i, \mathbf{p}_j)}_{\text{\small Partial loss for \textit{similar} pairs}}
\end{split}
\end{equation}
where $\upxi$ is a margin parameter used to prevent the dataset being reduced to a single point~\cite{xing2003distance}. That is, the attributed sequences with $\ell_{ij} = 1$ are only used to adjust the parameters in the transformation function $\Omega$ if the distance between them is larger than $\upxi$. The architecture of \olas is illustrated in Figure~\ref{fig-fdml}.

\subsection{OLAS Model Training}
With the contrastive loss $\mathcal{L}$ computed using Equation~\ref{eq-contrastive-loss}, we can now calculate the gradient $\nabla \mathcal{L}$, which is used to adjust parameters in the network as:
\begin{equation}
\label{eq-nabla-l}
    \nabla \mathcal{L} \equiv 
    \left[
        \frac{\partial{\mathcal{L}}}{\partial{\mathbf{W}_{\text{F}}}}, 
        \frac{\partial{\mathcal{L}}}{\partial{\mathbf{b}_{\text{F}}}}, 
        \frac{\partial{\mathcal{L}}}{\partial{\mathbf{W}_{\text{L}}}}, 
        \frac{\partial{\mathcal{L}}}{\partial{\mathbf{U}_{\text{L}}}}, 
        \frac{\partial{\mathcal{L}}}{\partial{\mathbf{b}_{\text{L}}}}
    \right]
\end{equation}
With the transformation function $\Omega$ and distance function $d$, the explicit form of $\nabla \mathcal{L}$ can be written as:
\begin{equation}
\nabla \mathcal{L} = 
\frac{\partial{\mathcal{L}}}{\partial{d_\omega}}\frac{\partial d_\omega}{\partial \Omega}
 \left[
     \frac{\partial\pmb{\upalpha}_m}{\partial{\mathbf{W}_{\text{F}}}}, 
     \frac{\partial \pmb{\upalpha}_m}{\partial{\mathbf{b}_{\text{F}}}}, 
     \frac{\partial \mathbf{h}^{(t_i)}}{\partial{\mathbf{W}_{\text{L}}}}, 
     \frac{\partial \mathbf{h}^{(t_i)}}{\partial{\mathbf{U}_{\text{L}}}}, 
     \frac{\partial \mathbf{h}^{(t_i)}}{\partial{\mathbf{b}_{\text{L}}}}
\right]
\end{equation}
where
\begin{equation}
\label{eq-l-omega}
\begin{split}
    \frac{\partial \mathcal{L}}{\partial d_\omega} = - \ell_{ij}\max(0, \upxi-&d_\omega(\mathbf{p}_i, \mathbf{p}_j))\\ &+ (1-\ell_{ij})d_\omega(\mathbf{p}_i, \mathbf{p}_j) 
\end{split}
\end{equation}
\begin{equation}
\label{eq-d-omega}
\frac{\partial d_\omega}{\partial \Omega} = \left(\mathbf{p}_i - \mathbf{p}_j\right) \cdot \left(\mathds{1}-(\mathbf{p}_i - \mathbf{p}_j)\right)
\end{equation}
where $\mathds{1}$ is a vector filled with ones. 

For the $m$-th layer in a fully connected neural network, we employ the following update functions:
\begin{equation}
    \begin{split}
        \frac{\partial \pmb{\upalpha}_m}{\partial \mathbf{W}_{m}} &= \pmb{\upalpha}_m\left(\mathds{1}-\pmb{\upalpha}_m\right)\pmb{\upalpha}_{m - 1} \\
        \frac{\partial \pmb{\upalpha}_m}{\partial \mathbf{b}_{m}} &= \pmb{\upalpha}_m\left(\mathds{1}-\pmb{\upalpha}_{m - 1}\right) 
    \end{split}
\end{equation}

Here we use three steps to explain how \olas back-propagates the gradients. We use a $\delta_{\mu, \nu}$ function to simplify the equations with $\mu = \{\text{i, f, o}\}$ and $\nu = \{\text{i, f, o, c}\}$:
\begin{equation}
  \delta_{\mu, \nu}=\left\{
  \begin{array}{@{}ll@{}}
    1, & \text{if}\ \mu = \nu \\
    0, & \text{otherwise}
  \end{array}\right.
\end{equation} 
First, we have the following equations for $\mathbf{h}^{(t)}$ and $\mathbf{c}^{(t)}$: 
\begin{equation}
 \begin{split}
\frac{\partial   \mathbf{h}^{(t)}}{\partial \mathbf{W}_{\nu}} =   \frac{\partial \mathbf{o}^{(t)}}{\partial \mathbf{W}_{\nu}}\odot\tanh\big(&\mathbf{c}^{(t)}\big) +   \mathbf{o}^{(t)} \\ &\odot(1 - \tanh^2(\mathbf{c}^{(t)}))   \frac{\partial \mathbf{c}^{(t)}}{\partial \mathbf{W}_{\nu}}\\
\frac{\partial   \mathbf{h}^{(t)}}{\partial \mathbf{U}_{\nu}} =   \frac{\partial \mathbf{o}^{(t)}}{\partial \mathbf{U}_{\nu}}\odot\tanh(&\mathbf{c}^{(t)}) +   \mathbf{o}^{(t)} \\ &\odot(1 - \tanh^2(\mathbf{c}^{(t)}))   \frac{\partial \mathbf{c}^{(t)}}{\partial \mathbf{U}_{\nu}}\\
\frac{\partial   \mathbf{h}^{(t)}}{\partial \mathbf{b}_{\nu}} =   \frac{\partial \mathbf{o}^{(t)}}{\partial \mathbf{b}_{\nu}}\odot\tanh(&\mathbf{c}^{(t)}) +   \mathbf{o}^{(t)}\\ &\odot(1 - \tanh^2(\mathbf{c}^{(t)}))   \frac{\partial \mathbf{c}^{(t)}}{\partial \mathbf{b}_{\nu}}
\end{split}
\end{equation}
\vspace{-3mm}
\begin{equation}
  \begin{split}
\frac{\partial    \mathbf{c}^{(t)}}{\partial \mathbf{W}_{\nu}} =   \frac{\partial \mathbf{f}^{(t)}}{\partial \mathbf{W}_{\nu}}\odot\mathbf{c}^{(t-1)} &+   \mathbf{f}^{(t)}\odot  \frac{\partial \mathbf{c}^{(t-1)}}{\partial \mathbf{W}_{\nu}}   + \\  &\frac{\partial \mathbf{i}^{(t)}}{\partial \mathbf{W}_{\nu}}\odot\mathbf{g}^{(t)} +   \mathbf{i}^{(t)}\odot  \frac{\partial \mathbf{g}^{(t)}}{\partial \mathbf{W}_{\nu}} \\
\frac{\partial    \mathbf{c}^{(t)}}{\partial \mathbf{U}_{\nu}} =   \frac{\partial \mathbf{f}^{(t)}}{\partial \mathbf{U}_{\nu}}\odot\mathbf{c}^{(t-1)} &+   \mathbf{f}^{(t)}\odot  \frac{\partial \mathbf{c}^{(t-1)}}{\partial \mathbf{U}_{\nu}}   +   \\ &\frac{\partial \mathbf{i}^{(t)}}{\partial \mathbf{U}_{\nu}}\odot\mathbf{g}^{(t)} +   \mathbf{i}^{(t)}\odot  \frac{\partial \mathbf{g}^{(t)}}{\partial \mathbf{U}_{\nu}} \\
\frac{\partial    \mathbf{c}^{(t)}}{\partial \mathbf{b}_{\nu}} =   \frac{\partial \mathbf{f}^{(t)}}{\partial \mathbf{b}_{\nu}}\odot\mathbf{c}^{(t-1)} &+   \mathbf{f}^{(t)}\odot  \frac{\partial \mathbf{c}^{(t-1)}}{\partial \mathbf{b}_{\nu}}   +  \\  &\frac{\partial \mathbf{i}^{(t)}}{\partial \mathbf{b}_{\nu}}\odot\mathbf{g}^{(t)} +   \mathbf{i}^{(t)}\odot  \frac{\partial \mathbf{g}^{(t)}}{\partial \mathbf{b}_{\nu}}
\end{split}
\end{equation}

Then, we have the following equations for $\mathbf{i}^{(t)}, \mathbf{f}^{(t)}$ and $\mathbf{o}^{(t)}$:
\begin{equation}
\begin{split}
\frac{\partial  \Delta_\mu}{\partial \mathbf{W}_{\nu}} =  \Delta_\mu  (1 - \Delta_\mu) \vec{\alpha}^{(t)} \delta_{\mu, \nu} \\
\frac{\partial  \Delta_\mu}{\partial \mathbf{U}_{\nu}} =  \Delta_\mu  (1 -  \Delta_\mu) \mathbf{h}^{(t-1)} \delta_{\mu, \nu} \\
\frac{\partial  \Delta_\mu}{\partial \mathbf{b}_{\nu}} =  \Delta_\mu  (1 - \Delta_\mu) \delta_{\mu, \nu}
\end{split}
\end{equation}
where $ \Delta_{\text{i}} = \mathbf{i}^{(t)}$, $ \Delta_{\text{f}} = \mathbf{f}^{(t)}$ and $ \Delta_{\text{o}} = \mathbf{o}^{(t)}$. 

Finally, we have the gradients for $\mathbf{g}^{(t)}$ as:
\begin{equation}
\begin{split}
\frac{\partial \mathbf{g}^{(t)}}{\partial \mathbf{W}_{\nu}} =   (1 -  (\mathbf{g}^{(t)}) ^2)\vec{\alpha}^{(t)} \delta_{c, \nu} \\[1pt]
\frac{\partial \mathbf{g}^{(t)}}{\partial \mathbf{U}_{\nu}} = (1 - (\mathbf{g}^{(t)}) ^2)\mathbf{h}^{(t-1)} \delta_{c, \nu} \\[1pt]
\frac{\partial \mathbf{g}^{(t)}}{\partial \mathbf{b}_{\nu}} =  (1 -  (\mathbf{g}^{(t)}) ^2) \delta_{c, \nu}
\end{split}
\end{equation}

With the learning rate $\lambda$, the parameters $\mathbf{W}_{\text{F}}, \mathbf{W}_{\text{L}}, \mathbf{U}_{\text{L}}, \mathbf{b}_{\text{F}}$ and $\mathbf{b}_{\text{L}}$ can be updated by the following equation until convergence is achieved: 
\begin{equation}
\label{eq-update-all}
    \begin{split}
        \mathbf{W}_{\text{F}} &= \mathbf{W}_{\text{F}} - \lambda  \frac{\partial \mathcal{L}}{\partial \mathbf{W}_{\text{F}}} \\[3pt]
        \mathbf{b}_{\text{F}} &= \mathbf{b}_{\text{F}} - \lambda  \frac{\partial \mathcal{L}}{\partial \mathbf{b}_{\text{F}}} \\[3pt]
        \mathbf{W}_{\text{L}} &= \mathbf{W}_{\text{L}} - \lambda  \frac{\partial \mathcal{L}}{\partial \mathbf{W}_{\text{L}}} \\[3pt]
        \mathbf{U}_{\text{L}} &= \mathbf{U}_{\text{L}} - \lambda  \frac{\partial \mathcal{L}}{\partial \mathbf{U}_{\text{L}}} \\[3pt]
        \mathbf{b}_{\text{L}} &= \mathbf{b}_{\text{L}} - \lambda  \frac{\partial \mathcal{L}}{\partial \mathbf{b}_{\text{L}}} \\[3pt]
    \end{split}
\end{equation}
We summarize the algorithms for updating the \olas network in Algorithm~\ref{alg-fase}. 

\subsection{Labeling Attributed Sequences}
Once we have trained the \olas network to recognize the similarities and dissimilarities between exemplars of attributed sequence pairs. The \olas is then ready to be used to assign labels to unlabeled attributed sequences in one-shot learning. 
Given a test attributed sequence $p_k$ from a set $\mathcal{K}$ of unlabeled instances, a set $\mathcal{G}=\{p_g\}_{g=1}^{G}$ of attributed sequences with $G$ categories, in which there is only one instance per category, and the goal is to classify $p_k$ into one of $G$ categories. 
We can now use the \olas network with only one forward pass to calculate the distance between $p_k$ with each of the $G$ attributed sequences and the label of the instance that is closest to $p_k$ is then assigned as the label of $p_k$. This process can be defined using maximum similarity as:
\begin{equation}
    \widehat{c_k} = \argmin_g d_\omega(\mathbf{p}_k, \mathbf{p}_g)
\end{equation}
where $\widehat{c_k}$ is the predicted label of $\mathbf{p}_k$. 

\section{Experiments}
\label{section-olas-experiments}
\subsection{Datasets}
Our solution has been motivated in part by use case scenarios observed at  Amadeus related to attributed sequences. For this reason, we now work with the log files of an Amadeus~\cite{amadeus} internal application. 
Also, we apply our methodology to real-world, public available Wikispeedia data~\cite{west2009wikispeedia}. We summarize the data descriptions as follows:

\begin{itemize}
    \item \textbf{Amadeus data (AMS1$\sim$AMS6)}. We sampled six datasets from the log files of an internal application at Amadeus IT Group. Each attributed sequence is composed of a user profile containing information (\eg, system configuration, office name) and a sequence of function names invoked by web click activities (\eg, login, search) ordered by time. 
    \item \textbf{Wikispeedia data (WS1$\sim$WS6)}. Wikispeedia is an online game requiring participants to click through from a given start page to an end page using fewest clicks~\cite{west2009wikispeedia}. We select the \textit{finished} path and extract several properties of each path (\eg, the category of the start path, time spent per click). We also sample six datasets from Wikispeedia. The Wikispeedia data is available through the Stanford Network Analysis Project\footnote{https://snap.stanford.edu/data/wikispeedia.html}~\cite{SNAPwiki}.  
\end{itemize}

Following the protocols in recent work~\cite{koch2015siamese}, we utilize the attributed sequences associated with 60\% of categories to generate attributed sequence triplets and use them in training.  

The class labels used in training and one-shot learning are disjoint sets. Similar to the strategy in~\cite{lake2013one}, where the authors designed a 20-way classification task that attempts to match an alphabet with one of the twenty possible classes, we randomly select one instance in the one-shot learning set and attempt to give it a correct label. We selected 2000 instances for each set used in one-shot learning and compute the accuracy. 
We summarize the number of classes in Table~\ref{tab-classes}.
\begin{table}[t]
    \centering
    \normalsize
    \caption{Number of Classes in Datasets}
    \label{tab-classes}
    \begin{tabular}{c|c|c}
    \hline
    Dataset & Training & One-shot Learning \\ 
    \hline
    AMS1, WS1 & 6 & 4 \\
    AMS2, WS2 & 12 & 8 \\
    AMS3, WS3 & 18 & 12 \\
    AMS4, WS4 & 24 & 16 \\ 
    AMS5, WS5 & 30 & 20 \\
    AMS6, WS6 & 36 & 24 \\
    \hline 
    \end{tabular}
\end{table}


\subsection{Compared Methods} We focus on one-shot learning methods on different data types. We summarize the compared methods in Table~\ref{tab-methods}. 
\begin{table}[t]
    \centering
    \normalsize
    \caption{Compared Methods}
    \label{tab-methods}
    \begin{tabular}{c|c|c}
        \hline
        Name & Data Used & Note \\ \hline
        \textsf{OLAS} & Attributed Sequences & This Work\\ 
        \textsf{OLASEmb} & {\scriptsize Attributed Sequence Embeddings} & This work + \cite{zhuang2018nas}\\
        \textsf{ATT} & Attributes Only & \cite{koch2015siamese}\\
        \textsf{SEQ} & Sequence Only & \cite{sutskever2014sequence} + \cite{koch2015siamese}\\
        \hline
    \end{tabular}
    \vspace{-7mm}
\end{table}
Specifically, we compare the performance of the following one-shot learning methods: 
\begin{itemize}
    \item \textsf{OLAS}: We first evaluate our proposed method using attributed sequences data. 
    \item \textsf{OLASEmb}: Instead of using the attributed sequence instances as input, we use the embeddings of attributed sequences as the input. 
    We want to find out whether a simpler heuristic combination of state-of-the-art would achieve better performance.  
    \item \textsf{ATT}: This is the state-of-the-art method~\cite{koch2015siamese} using only attributes of the data. 
    \item \textsf{SEQ}: We combine the state-of-the-art in one-shot learning~\cite{koch2015siamese} with sequence-to-sequence learning~\cite{sutskever2014sequence} to be able to utilize sequences in one-shot learning.  
\end{itemize}

\begin{algorithm}[t]
    \normalsize
    \begin{algorithmic}[1]
    \caption{One-shot learning for attributed sequences. }
    \label{alg-olas}
    \INPUT Trained networks \cnet $\Omega$ and \vnetns, a set of unlabeled attributed sequences $\mathcal{K}$, a set of labeled attributed sequence with one example per class $\mathcal{G}$ and a distance function $d$.
    \OUTPUT A set of labeleled attributed sequences $\mathcal{K}^{\prime}$.
    \State{$\mathcal{K}^{\prime} \leftarrow \emptyset$}
    \ForEach{$p_k \in \mathcal{K}$}
    \State{$\varepsilon \leftarrow +\infty$ \Comment{Set initial minimum distance to $+\infty$}}
        \State{$\mathbf{p}_k \leftarrow \Omega(p_k; \omega)$}
        \ForEach{$(p_g, c_g) \in \mathcal{G}$}
            \State{$\mathbf{p}_g \leftarrow \Omega(p_g; \omega)$}
            \If{$d{(\mathbf{p}_k, \mathbf{p}_g)} \le \varepsilon$ }
                \State{$\varepsilon \leftarrow d_\omega{(\mathbf{p}_k, \mathbf{p}_g)}$\Comment{Using \vnetns.}}
                \State{$\widehat{c_k} \leftarrow c_g$} \Comment{Assign the same label of $p_g$ to $p_k$.}
            \EndIf
        \State{$\mathcal{K}^{\prime} \leftarrow (p_k, \widehat{c_k})$} 
        \EndFor
    \EndFor
    \State{\Return $\mathcal{K}$}
    \end{algorithmic}
\end{algorithm}
\begin{figure*}[t]
    \centering
    \includegraphics[width=0.55\textwidth]{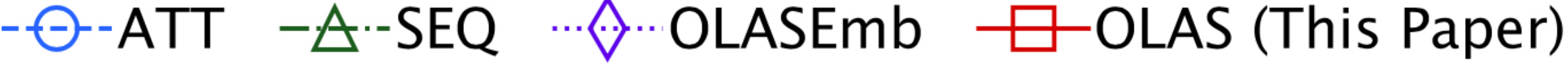}
    
    \begin{subfigure}[t]{0.32\linewidth}
        \centering
        \includegraphics[width=\textwidth, page=1]{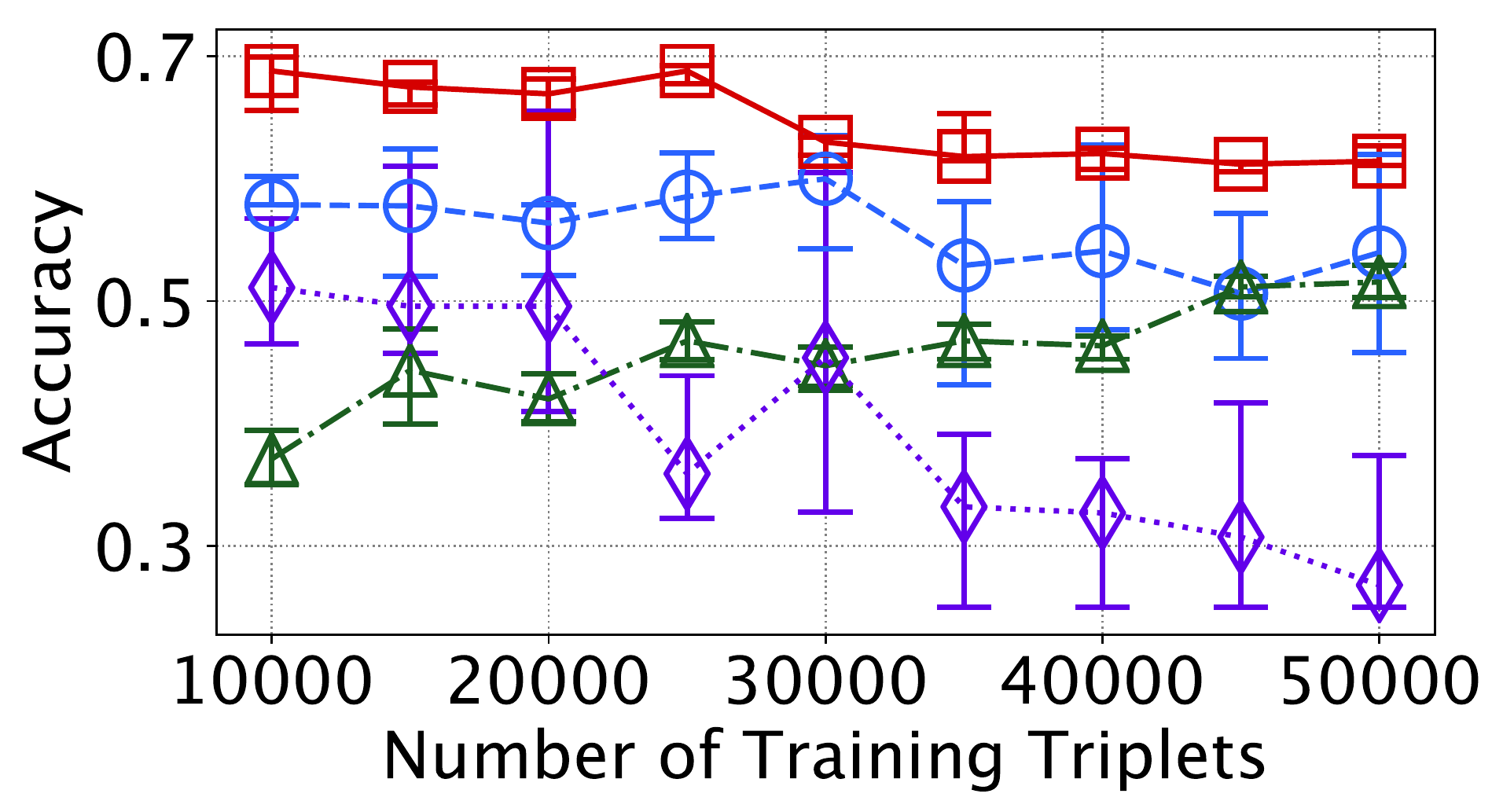}
        \vspace{-6mm}
        \caption{Dataset AMS1}
        \label{fig-exp-euc-ams-1}
    \end{subfigure}
    \begin{subfigure}[t]{0.32\linewidth}
        \centering
        \includegraphics[width=\textwidth, page=2]{figures/OLAS/experiments/0816/euclidean_deviation}
        \vspace{-6mm}
        \caption{Dataset AMS2}
        \label{fig-exp-euc-ams-2}
    \end{subfigure}
    \begin{subfigure}[t]{0.32\linewidth}
        \centering
        \includegraphics[width=\textwidth, page=3]{figures/OLAS/experiments/0816/euclidean_deviation}
        \vspace{-6mm}
        \caption{Dataset AMS3}
        \label{fig-exp-euc-ams-3}
    \end{subfigure}
    
    \begin{subfigure}[t]{0.32\linewidth}
        \centering
        \includegraphics[width=\textwidth, page=4]{figures/OLAS/experiments/0816/euclidean_deviation}
        \vspace{-6mm}
        \caption{Dataset AMS4}
        \label{fig-exp-euc-ams-4}
    \end{subfigure}
    \begin{subfigure}[t]{0.32\linewidth}
        \centering
        \includegraphics[width=\textwidth, page=5]{figures/OLAS/experiments/0816/euclidean_deviation}
        \vspace{-6mm}
        \caption{Dataset AMS5}
        \label{fig-exp-euc-ams-5}
    \end{subfigure}
    \begin{subfigure}[t]{0.32\linewidth}
        \centering
        \includegraphics[width=\textwidth, page=6]{figures/OLAS/experiments/0816/euclidean_deviation}
        \vspace{-6mm}
        \caption{Dataset AMS6}
        \label{fig-exp-euc-ams-6}
    \end{subfigure}
    \vspace{-2mm}
    \caption[\olas accuracy on AMS data with the Euclidean distance]{Accuracy of the label prediction on AMS datasets using \textbf{Euclidean} distance function. }
    \vspace{-2mm}
    \label{fig-exp-euc-ams}
\end{figure*}
\begin{figure*}[t]
    \centering
    \begin{subfigure}[t]{0.32\linewidth}
        \centering
        \includegraphics[width=\textwidth, page=1]{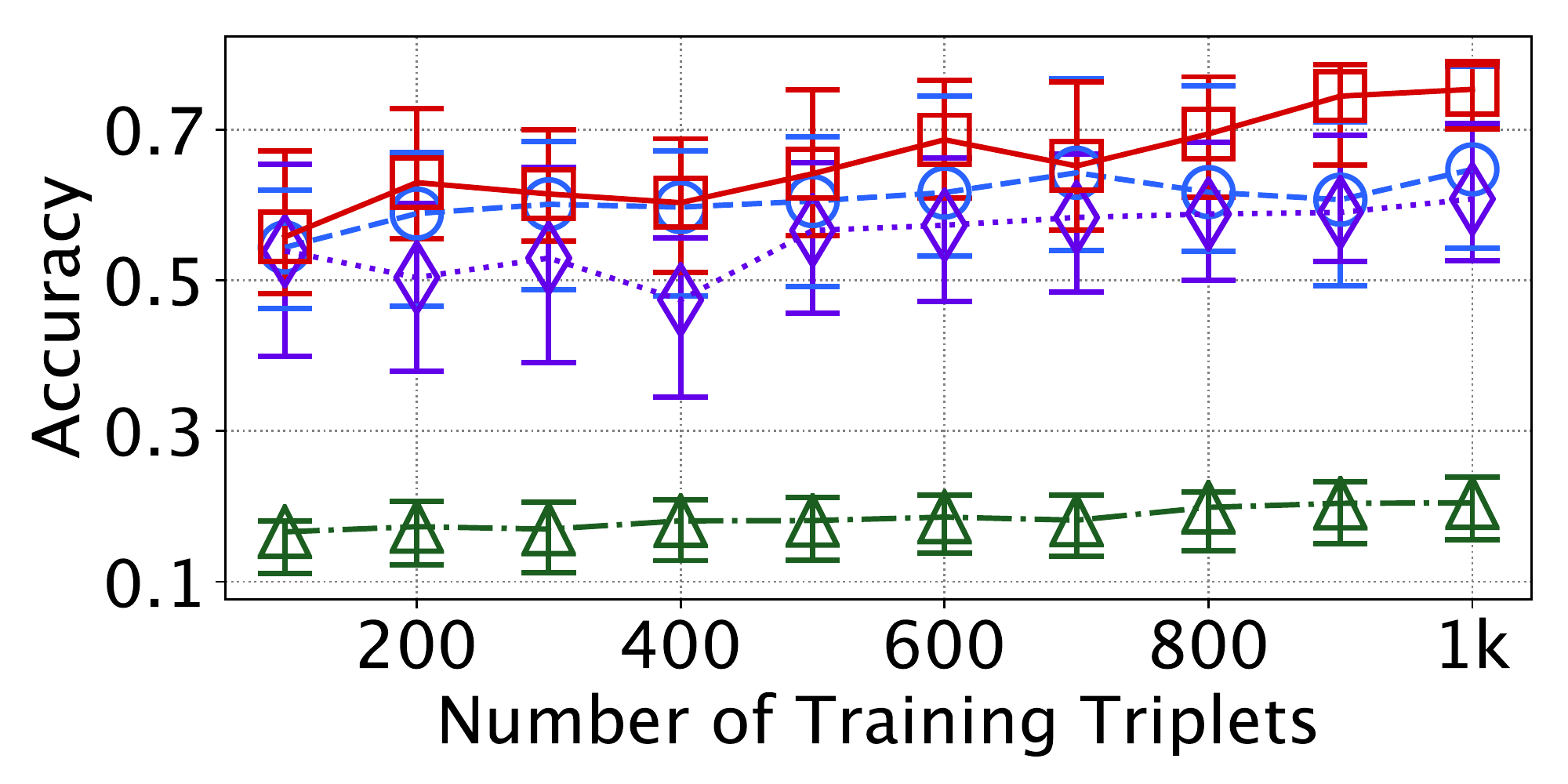}
        \vspace{-6mm}
        \caption{Dataset WS1}
        \label{fig-exp-euc-ws-1}
    \end{subfigure}
    \begin{subfigure}[t]{0.32\linewidth}
        \centering
        \includegraphics[width=\textwidth, page=2]{figures/OLAS/experiments/0816/ws_euclidean_deviation}
        \vspace{-6mm}
        \caption{Dataset WS2}
        \label{fig-exp-euc-ws-2}
    \end{subfigure}
    \begin{subfigure}[t]{0.32\linewidth}
        \centering
        \includegraphics[width=\textwidth, page=3]{figures/OLAS/experiments/0816/ws_euclidean_deviation}
        \vspace{-6mm}
        \caption{Dataset WS3}
        \label{fig-exp-euc-ws-3}
    \end{subfigure}
    
    \begin{subfigure}[t]{0.32\linewidth}
        \centering
        \includegraphics[width=\textwidth, page=4]{figures/OLAS/experiments/0816/ws_euclidean_deviation}
        \vspace{-6mm}
        \caption{Dataset WS4}
        \label{fig-exp-euc-ws-4}
    \end{subfigure}
    \begin{subfigure}[t]{0.32\linewidth}
        \centering
        \includegraphics[width=\textwidth, page=5]{figures/OLAS/experiments/0816/ws_euclidean_deviation}
        \vspace{-6mm}
        \caption{Dataset WS5}
        \label{fig-exp-euc-ws-5}
    \end{subfigure}
    \begin{subfigure}[t]{0.32\linewidth}
        \centering
        \includegraphics[width=\textwidth, page=6]{figures/OLAS/experiments/0816/ws_euclidean_deviation}
        \vspace{-6mm}
        \caption{Dataset WS6}
        \label{fig-exp-euc-ws-6}
    \end{subfigure}
    \vspace{-2mm}
    \caption[\olas accuracy on Wikispeedia data with the Euclidean distance]{Accuracy of the label prediction on Wikispeedia datasets using \textbf{Euclidean} distance function. }
    \vspace{-5mm}
    \label{fig-exp-euc-ws}
\end{figure*}
\subsection{Experiment Settings} 
\subsubsection{Protocols} The goal of one-shot learning is to correctly assign class labels to each instance. In order to compare with state-of-the-art work~\cite{koch2015siamese,bertinetto2016learning}, we also use accuracy to evaluate the performance. 
A higher accuracy score means a method could make more correct class label predictions. For each experiment setting, we repeat ten times and report the median, 25 percentile and 75 percentile of the results using error bars. For each training process using attributed sequence triplets, we hold out 20\% of the training data as the validation set. The holdout portion is not limited to the instances with certain labels, but instead, they are randomly chosen from all possible classes. 
\subsubsection{Network Initialization and Settings} 
Gradient-based methods often require a careful initialization of the neural networks. In our experiments, we use normalized random distribution~\cite{glorot2010understanding} to initialize weight matrices $\mathbf{W}_{\text{F}}$ and $\mathbf{W}_{\text{L}}$, orthogonal matrix is used to initialize recurrent matrices $\mathbf{U}_{\text{L}}$ and biases are initialized to zero vector $\pmb 0$. 
 Specifically, the $m$-th layer of the fully connected neural network is initialized as:
 \begin{equation*}
     \mathbf{W}_m \sim \text{Uniform}\left[-\frac{\sqrt{6}}{\sqrt{{n}_{m-1} + {n}_{m+1}}}, \frac{\sqrt{6}}{\sqrt{{n}_{m-1} + {n}_{m+1}}}\right]
 \end{equation*}
 where $n_{m}$ is the output dimension of the $m$-th layer. There are three layers used in our experiments. Meanwhile, the weight matrices $\mathbf{W}_{\text{i}},\mathbf{W}_{\text{f}},\mathbf{W}_{\text{o}},\mathbf{W}_{\text{c}}$ are initialized as:  
 \begin{equation*}
     \mathbf{W}_{\text{i}},\mathbf{W}_{\text{f}},\mathbf{W}_{\text{o}},\mathbf{W}_{\text{c}} \sim \text{Uniform}\left[-\sqrt{\frac{6}{n_l}}, \sqrt{\frac{6}{n_l}}\right]
 \end{equation*}
 where $n_l$ is the dimension of the output. In our experiments, we use 50 dimensions for both $n_l$ and $n_m$.
We utilize $\ell_2$-regularization with early stopping to avoid overfitting. The validation set is composed of 20\% of the total amount of attributed sequence triplets in the training set.  
\subsubsection{Performance Studies}
In this section, we present the performance studies of the proposed \olas network and compare it with techniques in the state-of-the-art. 

\begin{figure*}[t]
    \centering
    \includegraphics[width=0.55\textwidth]{figures/OLAS/experiments/legend}
    
    \begin{subfigure}[t]{0.32\linewidth}
        \centering
        \includegraphics[width=\textwidth, page=1]{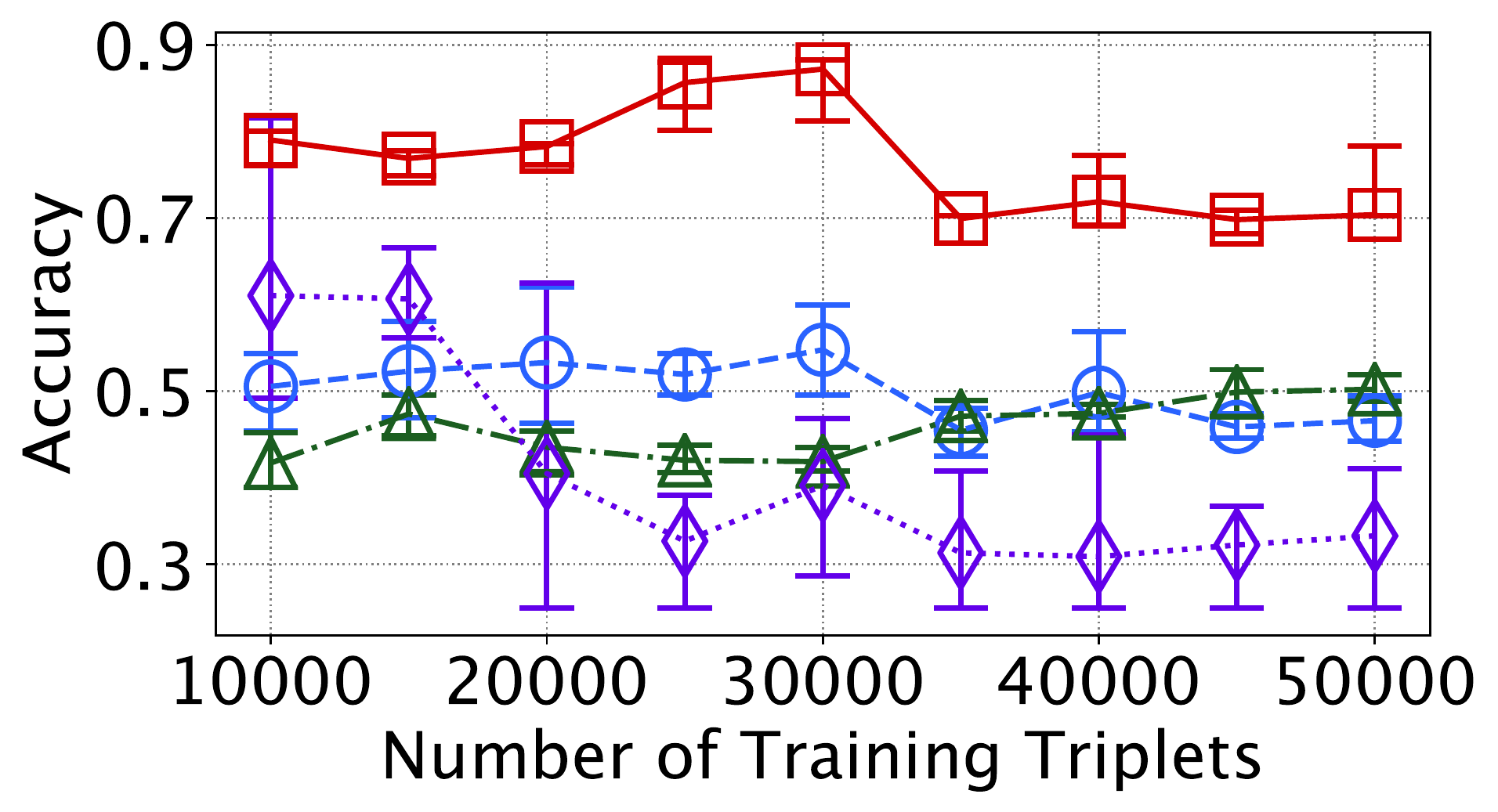}
        \vspace{-6mm}
        \caption{Dataset AMS1}
        \label{fig-exp-mah-ams-1}
    \end{subfigure}
    \begin{subfigure}[t]{0.32\linewidth}
        \centering
        \includegraphics[width=\textwidth, page=2]{figures/OLAS/experiments/0816/manhattan_deviation}
        \vspace{-6mm}
        \caption{Dataset AMS2}
        \label{fig-exp-mah-ams-2}
    \end{subfigure}
    \begin{subfigure}[t]{0.32\linewidth}
        \centering
        \includegraphics[width=\textwidth, page=3]{figures/OLAS/experiments/0816/manhattan_deviation}
        \vspace{-6mm}
        \caption{Dataset AMS3}
        \label{fig-exp-mah-ams-3}
    \end{subfigure}
    
    \begin{subfigure}[t]{0.32\linewidth}
        \centering
        \includegraphics[width=\textwidth, page=4]{figures/OLAS/experiments/0816/manhattan_deviation}
        \vspace{-6mm}
        \caption{Dataset AMS4}
        \label{fig-exp-mah-ams-4}
    \end{subfigure}
    \begin{subfigure}[t]{0.32\linewidth}
        \centering
        \includegraphics[width=\textwidth, page=5]{figures/OLAS/experiments/0816/manhattan_deviation}
        \vspace{-6mm}
        \caption{Dataset AMS5}
        \label{fig-exp-mah-ams-5}
    \end{subfigure}
    \begin{subfigure}[t]{0.32\linewidth}
        \centering
        \includegraphics[width=\textwidth, page=6]{figures/OLAS/experiments/0816/manhattan_deviation}
        \vspace{-6mm}
        \caption{Dataset AMS6}
        \label{fig-exp-mah-ams-6}
    \end{subfigure}
    \vspace{-2mm}
    \caption[\olas accuracy on AMS data with the Manhattan distance]{Accuracy of the label prediction on AMS datasets using \textbf{Manhattan} distance function. }
    \vspace{-3mm}
    \label{fig-exp-mah-ams}
\end{figure*}
\begin{figure*}[t]
    \centering
    \begin{subfigure}[t]{0.32\linewidth}
        \centering
        \includegraphics[width=\textwidth, page=1]{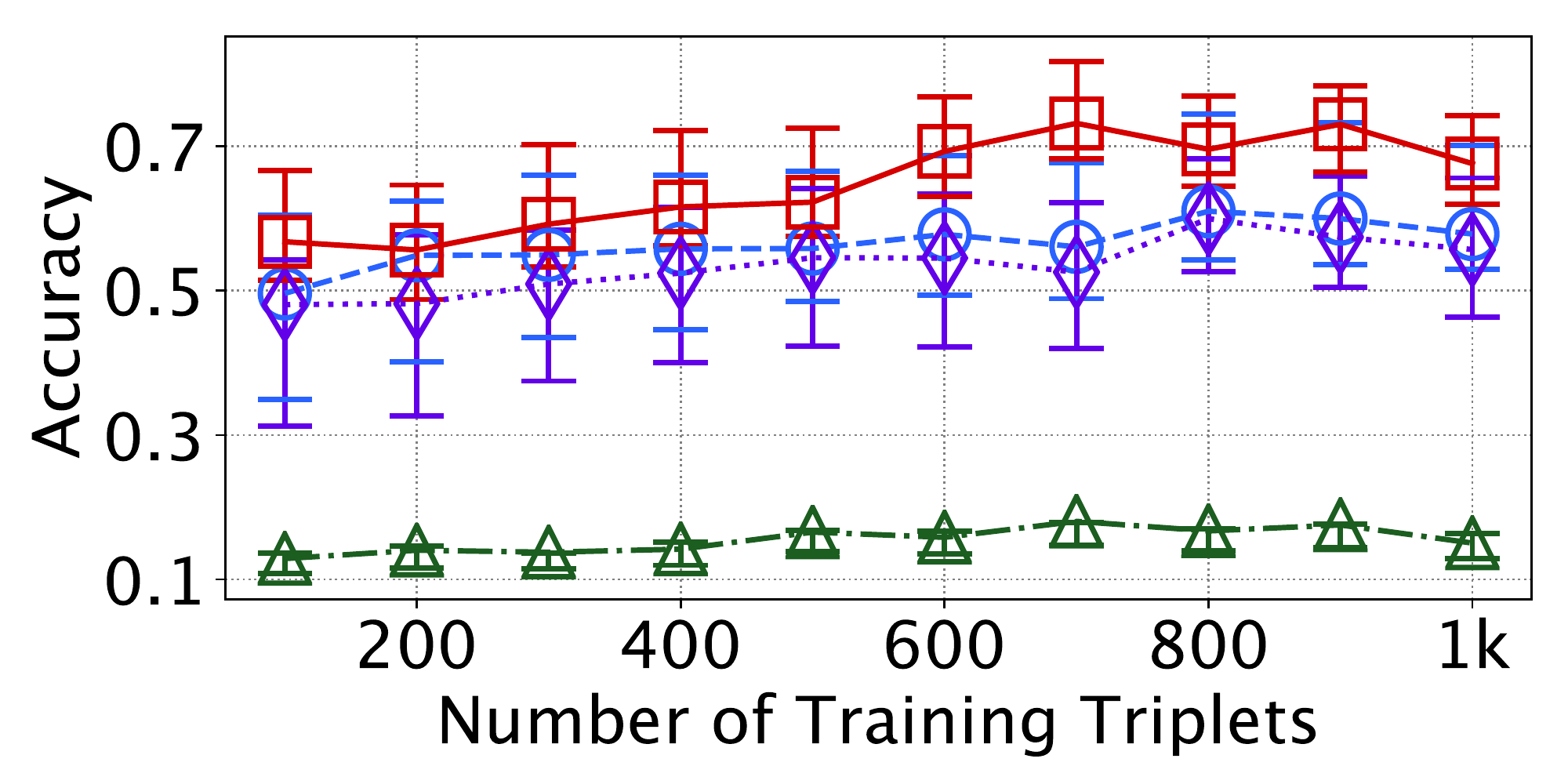}
        \vspace{-6mm}
        \caption{Dataset WS1}
        \label{fig-exp-mah-ws-1}
    \end{subfigure}
    \begin{subfigure}[t]{0.32\linewidth}
        \centering
        \includegraphics[width=\textwidth, page=2]{figures/OLAS/experiments/0816/ws_manhattan_deviation}
        \vspace{-6mm}
        \caption{Dataset WS2}
        \label{fig-exp-mah-ws-2}
    \end{subfigure}
    \begin{subfigure}[t]{0.32\linewidth}
        \centering
        \includegraphics[width=\textwidth, page=3]{figures/OLAS/experiments/0816/ws_manhattan_deviation}
        \vspace{-6mm}
        \caption{Dataset WS3}
        \label{fig-exp-mah-ws-3}
    \end{subfigure}
    
    \begin{subfigure}[t]{0.32\linewidth}
        \centering
        \includegraphics[width=\textwidth, page=4]{figures/OLAS/experiments/0816/ws_manhattan_deviation}
        \vspace{-6mm}
        \caption{Dataset WS4}
        \label{fig-exp-mah-ws-4}
    \end{subfigure}
    \begin{subfigure}[t]{0.32\linewidth}
        \centering
        \includegraphics[width=\textwidth, page=5]{figures/OLAS/experiments/0816/ws_manhattan_deviation}
        \vspace{-6mm}
        \caption{Dataset WS5}
        \label{fig-exp-mah-ws-5}
    \end{subfigure}
    \begin{subfigure}[t]{0.32\linewidth}
        \centering
        \includegraphics[width=\textwidth, page=6]{figures/OLAS/experiments/0816/ws_manhattan_deviation}
        \vspace{-6mm}
        \caption{Dataset WS6}
        \label{fig-exp-mah-ws-6}
    \end{subfigure}
    \vspace{-2mm}
    \caption[\olas accuracy on Wikispeedia data with the Manhattan distance]{Accuracy of the label prediction on Wikispeedia datasets using \textbf{Manhattan} distance function. }
    \vspace{-5mm}
    \label{fig-exp-mah-ws}
\end{figure*}

\subsubsection{Varying number of training triplets} 
Figure~\ref{fig-exp-euc-ams} and~\ref{fig-exp-euc-ws} present the results where each setting has a fixed number of labels while the number of training triplets increases. Based on the experiment result figures, we have the following observations: 
\begin{itemize}
    \item As more triplets being used in the training process, the accuracy of one-shot learning keeps increasing with the trained \olas network. Intuitively, with more examples demonstrated to the \olasns, it could better gain a better capability of generalization, even though the data instances used in one-shot learning are previously unseen. 
    \item Overfitting challenges the performance of all one-shot learning approaches. Although we use early stopping and $\ell_2$-regularization in all experiments, overfitting can still be challenging due to there is only one example per class in one-shot learning. 
    \item \olas can achieve better performance than other baseline methods when there are more possible classes. While \olas maintains a stable performance outperforming state-of-the-art under various parameter settings, \olas can achieve a better performance when the classification task become \textit{harder} with more possible class labels.  
\end{itemize}

\subsubsection{Observations using different datasets} Different from the synthetic datasets, the real-world applications often consist of diverse and noisy data instances. It is also interesting to examine the results using different real-world datasets. We find that the performance of \olas remains superior when we use the twelve datasets sampled from two real-world applications.  

\subsubsection{Advantage of the \textit{end-to-end} model} Although it is possible to use attributed sequence embeddings~\cite{zhuang2018nas} with one-shot learning, the experiment results have proven that the performance of the end-to-end solution in this work is far superior to and more stable than \textsf{OLASEmb}. Specifically, the performance of our closest baseline method \textsf{OLASEmb} has varied more compared to all other methods.  Building an end-to-end model allows the back-propagation of gradient throughout all layers in the \olas model. On the other hand, the two gradients in \textsf{OLASEmb}, \ie, the gradient in the model for generating attributed sequence embedding and the gradient in one-shot learning model, are independent and thus the parameters within this method cannot be better adjusted than our solution \olasns.

\subsubsection{Effect of different distance functions. } Recent work~\cite{bertinetto2016learning} has observed significant differences in performance when using different distance functions. Here, we substitute the Euclidean distance function with the Manhattan distance function to see the performance of all compared methods. We observe that the proposed \olas model is capable of achieving the best results despite which one of the two distance functions are used.

 \chapter{Attention Model for Attributed Sequence Classification}
\label{chapter-task4}
\newpage 

\section{Problem Definition.} 
We formulate the attributed sequence classification problem as the problem of finding the parameters $\theta$ of a predictor $\Theta$ that minimizes the prediction error of class labels. 
Intuitively, we want to maximize the possibility of correctly predicting labels when given a training set $\mathbb{P} = \{ p_1, \cdots, p_k \}$ of $k$ attributed sequences. Thus, we formulate the training process as an optimization process: 
\begin{equation}
    \label{eq-training-goal}  
        \argmin_{\theta} - \sum_{i} \Pr(l_i) \log \Pr\left(\Theta\left(p_i\right)\right)
\end{equation} 
Our goal is to find the parameters that minimize the categorical cross-entropy loss between the predicted labels using parameters in function $\Theta$ and the true labels for all attributed sequences in the dataset. 

\begin{table}[!ht]
    \centering
    \caption{Important Mathematical Notations}
    \vspace{-10pt}
    \label{tab-notation}
    \begin{tabular}{cp{12cm}}
        \hline
        \textbf{Notation} & \textbf{Description} \\ \hline
        $\mathbb{R}$ & The set of real numbers. \\
        $\mathbb{P}$ & A set of attributed sequences. \\
        $r$ & The number of all possible items in sequences. \\ 
        $s_i$ & A sequence of categorical items. \\
        $x_i^{(t)}$ & The $t$-th item in sequence $s_i$. \\
        $t_{\text{max}}$ & The maximum length of sequences.\\ 
        $\mathbf{s}_i$ & A one-hot encoded sequence in the form of a matrix $\mathbf{s}_i  \in \mathbb{R}^{t_{\text{max}\times r}}$. \\
        $\textbf{x}_i^{(t)}$ & A one-hot encoded item at $t$-th time step. \\
        $\mathbf{v}_i$ & An attribute vector. \\
        $p_i$ & An attributed sequence. \ie, $p_i = (\mathbf{v}_i, \mathbf{s}_i)$\\
        $\pmb{p}_i$ & A feature vector of attributed sequence $p_i$. \\
        $\pmb\mu_i$ & Attention weights.   \\
        $\pmb\upalpha_i$ & Attention vector. \\
        \hline
    \end{tabular}
    \vspace{-10pt}
\end{table}

\section{Attributed Sequence Attention Mechanism}
The proposed \amas model has three components, one \anet for learning the attribute information, one \snet to learn the sequential information, and one \attn to learn the attention from both attributes and sequences. 

\subsection{Network Components.}
\label{sec-anet-snet}
\subsubsection{\anetns}
We build \anet using fully connected neural network denoted as:
\begin{equation}
    \label{eq-att-func}
    f\left(\mathbf{A}; \mathbf{W_r}, \mathbf{b_r}\right) = \tanh (\mathbf{W}_{\text{r}} \mathbf{A} + \mathbf{b}_{\text{r}})
\end{equation}
where $\mathbf{W}_a$ and $\mathbf{b}_a$ are two trainable parameters in the \anetns, denoting the weight matrix and bias vector, respectively. We use the activation function $\tanh$ here in our \anet based on our empirical studies. Other choices, such as \texttt{ReLu} or \texttt{sigmoid}, may work equally well in other real-world scenarios. 

When given an attributed sequence $p_i = (\mathbf{v}_i, \mathbf{s}_i)$, \anet takes the attributes as input and generates an attribute vector $\pmb{r}_i = f(\mathbf{v}_i; \mathbf{W_r}, \mathbf{b_r})$. 
Different from previous work in~\cite{chen2017outlier, wang2014generalized} using stacked fully connected neural network as autoencoder, where the training goal is to minimize the reconstruction error, our goal of \anet is to work together with other network components to maximize the possibility of predicting the correct labels. 
\begin{figure}[!ht]
    \captionsetup{justification=centering}
    \begin{subfigure}[t]{0.45\linewidth}
        \centering
        \includegraphics[width=0.98\linewidth]{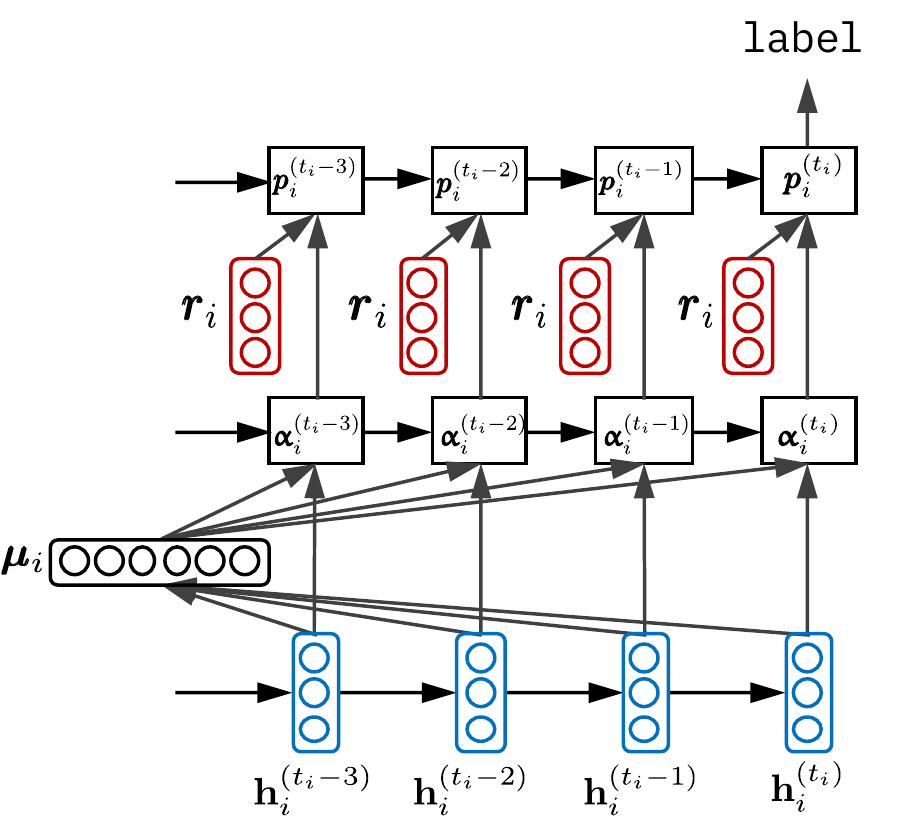}
        \caption{Attributed Sequence Attention (ASA).  }
        \vspace{-3mm}
        \label{fig-asa}
    \end{subfigure}
    \hspace{1cm}
    \begin{subfigure}[t]{0.5\linewidth}
        \centering
        \includegraphics[width=0.98\linewidth]{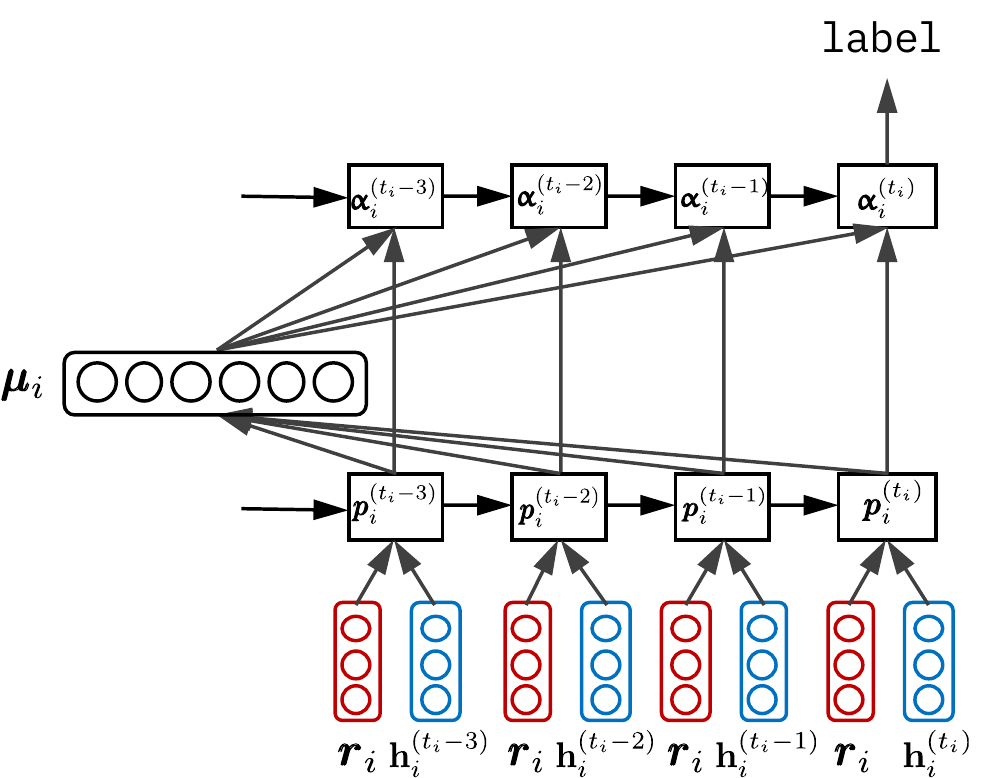}
        \caption{Attributed Sequence Hybrid Attention (ASHA). }
        \vspace{-3mm}
        \label{fig-asha}
    \end{subfigure}
    \caption[Two types of attention for attributed sequence classification]{Two types of attention for attributed sequence classification in this task, where $\pmb\mu_i = \left[\pmb\mu_i^{(1)}, \cdots, \pmb\mu_i^{(t_i)}\right]$ is the attention weights. }
    \vspace{-5mm}
\end{figure}
\subsubsection{\snetns}
Different from the attributes being unordered, items in our sequences have a temporal ordering. The information about the sequences is in both the item values and the ordering of items. The temporal orderings require a model that is capable of handling the dependencies between different items. There have been extensive studies on using recurrent neural networks (RNN) to handle temporal dependencies. However, RNN suffers from the problem of exploding and vanishing gradient during the training, where the gradient value becomes too large or too small and thus the network becomes untrainable. Long Short-Term Memory (LSTM)~\cite{hochreiter1997long} is designed as one variation and expansion of the RNN to handle such issues. LSTM is capable of ``remembering'' values over long time intervals by introducing additional internal variables (\ie various ``gates'' and ``cell states''). We use an LSTM to handle the dependencies. 
With a variable $\mathbf{X}^{(t)}$ at time $t$, the \snet can be expressed as:
\begin{equation}
    \begin{split}
    \label{eq-lstm}
    \mathbf{i}^{(t)} &= \sigma\left(\mathbf{W}_{\text{i}}\mathbf{X}^{(t)} + \mathbf{U}_{\text{i}}\mathbf{h}^{(t-1)} + \mathbf{b}_{\text{i}}\right) \\
    \mathbf{f}^{(t)} &= \sigma\left(\mathbf{W}_{\text{f}}\mathbf{X}^{(t)} + \mathbf{U}_{\text{f}}\mathbf{h}^{(t-1)} + \mathbf{b}_{\text{f}}\right) \\
    \mathbf{o}^{(t)} &= \sigma\left(\mathbf{W}_{\text{o}}\mathbf{X}^{(t)} + \mathbf{U}_{\text{o}}\mathbf{h}^{(t-1)} + \mathbf{b}_{\text{o}}\right) \\
    \mathbf{g}^{(t)} &= \tanh\left(\mathbf{W}_{\text{c}}\mathbf{X}^{(t)} + \mathbf{U}_{\text{c}}\mathbf{h}^{(t-1)} + \mathbf{b}_{\text{c}}\right) \\
    \mathbf{c}^{(t)} &= \mathbf{f}^{(t)}\odot\mathbf{c}^{(t-1)} + \mathbf{i}^{(t)} \odot \mathbf{g}^{(t)} \\[2pt]
    \mathbf{h}^{(t)} &= \mathbf{o}^{(t)} \odot \tanh\left(\mathbf{c}^{(t)}\right)
    \end{split}
  \end{equation}
  where $\odot$ denotes the bitwise multiplication, $\sigma$ is a \texttt{sigmoid} activation function, $\mathbf{i}^{(t)}$, $\mathbf{f}^{(t)}$ and $\mathbf{o}^{(t)}$ are the internal gates of the LSTM, and $\mathbf{c}^{(t)}$ and $\mathbf{h}^{(t)}$ are the cell and hidden states of the LSTM, respectively. We denote the \snet as:
  \begin{equation}
    \label{eq-seq-func}
    g(\mathbf{X}^{(t)}; \mathbf{W}_{\text{s}}, \mathbf{U}_{\text{s}}, \mathbf{b}_{\text{s}}) = \mathbf{h}^{(t)}
  \end{equation}
  where $\mathbf{W}_{\text{s}} = [\mathbf{W}_{\text{i}}, \mathbf{W}_{\text{f}}, \mathbf{W}_{\text{o}}, \mathbf{W}_{\text{c}}]$, $\mathbf{U}_{\text{s}} = [\mathbf{U}_{\text{i}}, \mathbf{U}_{\text{f}}, \mathbf{U}_{\text{o}}, \mathbf{U}_{\text{c}}]$ and $\mathbf{b}_{\text{s}} = [\mathbf{b}_{\text{i}}, \mathbf{b}_{\text{f}}, \mathbf{b}_{\text{o}}, \mathbf{b}_{\text{c}}]$. 
  
  With the sequence $\mathbf{s}_i = \left[\mathbf{x}_i^{(1)}, \cdots, \mathbf{x}_i^{(t_i)} \right]$ as part of an attributed sequence $p_i$, the hidden states for input $\mathbf{x}_i^{(t)}$ are $g\left(\mathbf{x}_i^{(t)}; \mathbf{W}_{\text{s}}, \mathbf{U}_{\text{s}}, \mathbf{b}_{\text{s}}\right) = \mathbf{h}_i^{(t)}$.

  \subsubsection{\attns} 
  Recent work~\cite{mnih2014recurrent, chorowski2015attention} has identified that even LSTM-based solutions cannot fully handle the sequence learning on long sequences that the information over a long time may be lost. One popular solution to this problem is to incorporate the attention mechanism into the model. The attention mechanism effectively summarizes the data with the aim to leverage the importance of each item in the sequential input. 
  
  \paragraph{Attributed Sequence Attention (ASA).} Different from the common sequence attention models, we now need to incorporate the attribute information into the learning process. Here, we design the \attn as follows: First, we need to compute the attention weight $\pmb\mu_i^{(t)}$ at $t$-th time as: 
  \vspace{-3mm}
  \begin{gather*}
        g\left(\mathbf{x}_i^{(t)}\right) = \mathbf{h}_i^{(t)} \\
        \pmb\mu_i^{(t)} = \frac{
          \exp\left(
              g\left(\mathbf{x}_i^{(t)}\right)
              \right)
      }{
          \sum_{j=1}^{t_i}g\left(\mathbf{x}_i^{(j)}\right)
          }
          \vspace{-10mm}
  \end{gather*}
  Then, the attention weight is multiplied with the hidden state at each time step: 
  \begin{equation}
      \pmb\upalpha_i^{(t)} = \pmb\mu_i^{(t)} \odot g\left(\mathbf{x}_i^{(t)}\right), t=1,2,\cdots,t_i
  \end{equation}
  The attention weight $\mu_i^{(t)}$ at $t$ time is randomly initialized and incrementally adjusted during the training process. The output at each time step is then augmented with the outputs from \anet as:
  \begin{equation*}
      \pmb p_i^{(t)} = f(\mathbf{v}_i) \oplus \pmb\upalpha_i^{(t)}
  \end{equation*}
  At the last time step $t_i$, we denote $\pmb p_i = \pmb p_i^{(t_i)}$ to simply the notation. 
  
  \paragraph{Attributed Sequence Hybrid Attention (ASHA).} Different from the previous ASA approach, the outputs of \anet and \snet are augmented with the \attns. The attention weight is written as: 
  \begin{gather*}
    \vspace{-10mm}
    d\left(\mathbf{v}_i, \mathbf{x}_i^{(t)}\right) = f(\mathbf{v}_i) \oplus g\left(\mathbf{x}_i^{(t)}\right) \\
      \pmb\mu_i^{(t)} = \frac{
          \exp \left( d\left(\mathbf{v}_i, \mathbf{x}_i^{(t)}\right) \right)
          }{
              \sum_{j=1}^{t_i}d\left(\mathbf{v}_i, \mathbf{x}_i^{(j)}\right)
              }
              \vspace{-5mm}
  \end{gather*}
  Then, the vectors used for classification is: 
  \begin{equation}
      \pmb\upalpha_i^{(t)} = \pmb\mu_i^{(t)} \odot d\left(\mathbf{v}_i, \mathbf{x}_i^{(t)}\right), t=1,2,\cdots,t_i
  \end{equation}

\subsection{Attributed Sequence Classification}
In the solution of attributed sequence classification \textit{without} attention, the \anet and the \snet are first concatenated as:
\begin{equation}
    \label{eq-concat}
    \pmb{p}_i = d\left(\mathbf{v}_i, \mathbf{x}_i^{(t_i)}\right) = \pmb{r}_i \oplus \mathbf{h}_i^{(t_i)}
\end{equation}
Here, $\oplus$ denotes the concatenation and $t_i$ denotes the last item in $\mathbf{s}_i$. Although all attributed sequences in the dataset are zero-padded to the maximum length $t_{\text{max}}$, the padded zero values are masked and not used in the computation. We model the process of predicting the label for each attributed sequence as:
\begin{equation*}
  \Theta(p_i) = \left\{
  \begin{array}{@{}ll@{}}
    \sigma(\mathbf{W}_{\text{p}} \pmb{p}_i + \mathbf{b}_{\text{p}}), & \text{ASA or No Attention} \vspace{3mm} 
    \\
    \sigma\left(\mathbf{W}_{\text{p}} \pmb{\upalpha}_i^{(t_i)} + \mathbf{b}_{\text{p}}\right), & \text{ASHA}
  \end{array}\right.
\end{equation*} 
where $\sigma$ is a \texttt{sigmoid} activation function and $\hat{l}_i = \Theta(p_i)$ is the predicted label. The $\mathbf{W}_{\text{p}}$ and $\mathbf{b}_{\text{p}}$ are both trainable in our model. 

\subsection{Training}
\label{sec-training}

\subsubsection{Regularization}
We adopt multiple strategies for different components in our \amas network. We empirically select the following regularization strategies in our model based on: (1). For \snetns, we apply $\ell_2$-regularization to the recurrent unit. 
    (2). Dropout with a rate of 0.5 is used to regularize the fully connected layer in \anetns. 
    (3). Lastly, we use Dropout with a rate of 0.2 in other fully connected layers in the model. 
    Based on our observations, using regularization on \attn has no significant impact on the performance of \amasns. 

\subsubsection{Optimizer}
We use an optimizer that computes the adaptive learning rates for every parameters, referred to as \underline{Ada}ptive \underline{M}oment Estimation (\texttt{Adam})~\cite{kingmaadam}. The core idea is to keep (1). an exponentially decaying average of gradients in the past and (2). a squared past gradient. \texttt{Adam} counteracts the biases as:
\begin{gather*}
    \widehat{\omega^{(t)}} = \frac{
        \beta_1 \omega^{(t-1)} + (1 - \beta_1) m^{(t)}
    }{
        1 - \beta_1^t
    } \\
    \widehat{\nu^{(t)}} = \frac{
        \beta_2 \nu^{(t-1)} + (1 - \beta_2) \left(m^{(t)}\right)^2
    }{
        1 - \beta_2^t
    }
\end{gather*}
where $\beta_1$ and $\beta_2$ are the decay rates, and $m^{(t)}$ is the gradient. We adopt $\beta_1=0.9$ and $\beta_2=0.999$ as in~\cite{kingmaadam}. Finally, the \texttt{Adam} updates the parameters as:
\begin{equation*}
    \gamma^{(t+1)} = \gamma^{(t)} - \frac{\rho}{
        \sqrt{\widehat{\nu^{(t)}} + \epsilon
        }
        }\widehat{\omega^{(t)}}
\end{equation*}
where $\rho$ is a static learning rate and $\epsilon$ is a constant with a small value to avoid division errors, such as division by zero. We empirically select $\rho=0.01$.

\section{Experiments}
\begin{table}[!ht]
    \centering
    \caption{Compared Methods}
    \begin{tabular}{c|c|c|c}
        \hline
        \textbf{Name} & \textbf{Data Used} & \textbf{Attention} & \textbf{Note} \\ \hline
        \textsf{BLA} & Attributes & No & \cite{hinton2006reducing} \\ \hline
        \textsf{BLS} & Sequences & No & \cite{wang2016recurrent} \\\hline
        \multirow{2}{*}{\textsf{BLAS}} & Attributes & \multirow{2}{*}{No} & \multirow{2}{*}{\cite{zhuang2018nas}}\\
        & Sequences & & \\\hline
        \textsf{SOA} & Sequences & Yes & \cite{yang2016hierarchical} \\\hline
        \multirow{2}{*}{\textsf{ASA}} & Attributes & \multirow{2}{*}{Yes} &\multirow{2}{*}{This paper} \\
        & Sequences & & \\\hline
        \multirow{2}{*}{\textsf{ASHA}} & Attributes & \multirow{2}{*}{Yes} &\multirow{2}{*}{This paper} \\
        & Sequences & & \\
        \hline
    \end{tabular}
    \vspace{-5mm}
    \label{tab-methods}
\end{table}
\begin{figure}[!ht]
    \centering
    \begin{subfigure}[t]{.45\linewidth}
        \centering
        \includegraphics[page=5, width=\textwidth]{figures/AMAS/exp/amas_accuracy.pdf}
        \vspace{-7mm}
        \caption{ASM-1, 83 classes}
        \label{fig-perf-A1}
    \end{subfigure}
    \begin{subfigure}[t]{.45\linewidth}
        \centering
        \includegraphics[page=6, width=\textwidth]{figures/AMAS/exp/amas_accuracy.pdf}
        \vspace{-7mm}
        \caption{ASM-2, 83 classes}
        \label{fig-perf-A2}
    \end{subfigure}
    \begin{subfigure}[t]{.45\linewidth}
        \centering
        \includegraphics[page=3, width=\textwidth]{figures/AMAS/exp/amas_accuracy.pdf}
        \vspace{-7mm}
        \caption{Wiki-1, 64 classes}
        \label{fig-perf-W1}
    \end{subfigure}
    \begin{subfigure}[t]{.45\linewidth}
        \centering
        \includegraphics[page=4, width=\textwidth]{figures/AMAS/exp/amas_accuracy.pdf}  
        \vspace{-7mm}
        \caption{Wiki-2, 64 classes}
        \label{fig-perf-W2}
    \end{subfigure}
    \begin{subfigure}[t]{.45\linewidth}
        \centering
        \includegraphics[page=1, width=\textwidth]{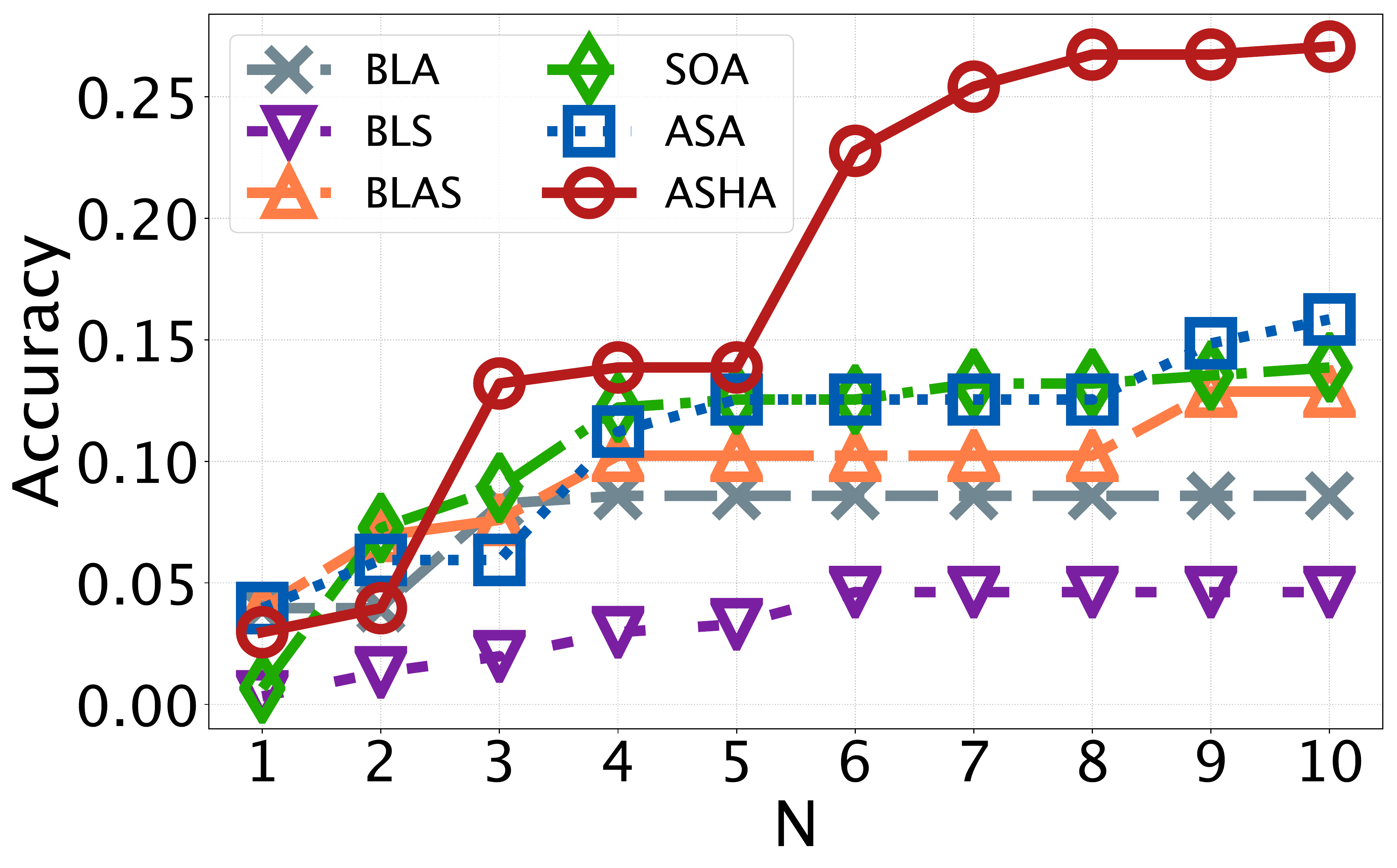}
        \vspace{-7mm}
        \caption{Reddit-1, 140 classes}
        \label{fig-perf-R1}
    \end{subfigure}
    \begin{subfigure}[t]{.45\linewidth}
        \centering
        \includegraphics[page=2, width=\textwidth]{figures/AMAS/exp/amas_accuracy.pdf}
        \vspace{-7mm}
        \caption{Reddit-2, 140 classes}
        \label{fig-perf-R2}
    \end{subfigure}
    \vspace{-3mm}
    \caption[\amas performance evaluation]{Performance comparison on all six datasets. }
    \vspace{-5mm}
    \label{fig-exp-perf}
\end{figure}
\subsection{Datasets}
\begin{figure}[!ht]
    \centering
    \begin{subfigure}[t]{.45\linewidth}
        \centering
        \includegraphics[page=5, width=\textwidth]{figures/AMAS/exp/amas_adaptive_accuracy.pdf}
        \vspace{-7mm}
        \caption{\textsf{ASA} model on AMS-1 dataset. }
        \label{fig-asa-A1}
    \end{subfigure}
    \begin{subfigure}[t]{.45\linewidth}
        \centering
        \includegraphics[page=6, width=\textwidth]{figures/AMAS/exp/amas_adaptive_accuracy.pdf}
        \vspace{-7mm}
        \caption{\textsf{ASHA} model on AMS-1 dataset. }
        \label{fig-asha-A1}
    \end{subfigure}
    \begin{subfigure}[t]{.45\linewidth}
        \centering
        \includegraphics[page=3, width=\textwidth]{figures/AMAS/exp/amas_adaptive_accuracy.pdf}
        \vspace{-7mm}
        \caption{\textsf{ASA} model on Wiki-1 dataset. }
        \label{fig-asa-W1}
    \end{subfigure}
    \begin{subfigure}[t]{.45\linewidth}
        \centering
        \includegraphics[page=4, width=\textwidth]{figures/AMAS/exp/amas_adaptive_accuracy.pdf}
        \vspace{-7mm}
        \caption{\textsf{ASHA} model on Wiki-1 dataset. }
        \label{fig-asha-W1}
    \end{subfigure}
    \begin{subfigure}[t]{.45\linewidth}
        \centering
        \includegraphics[page=1, width=\textwidth]{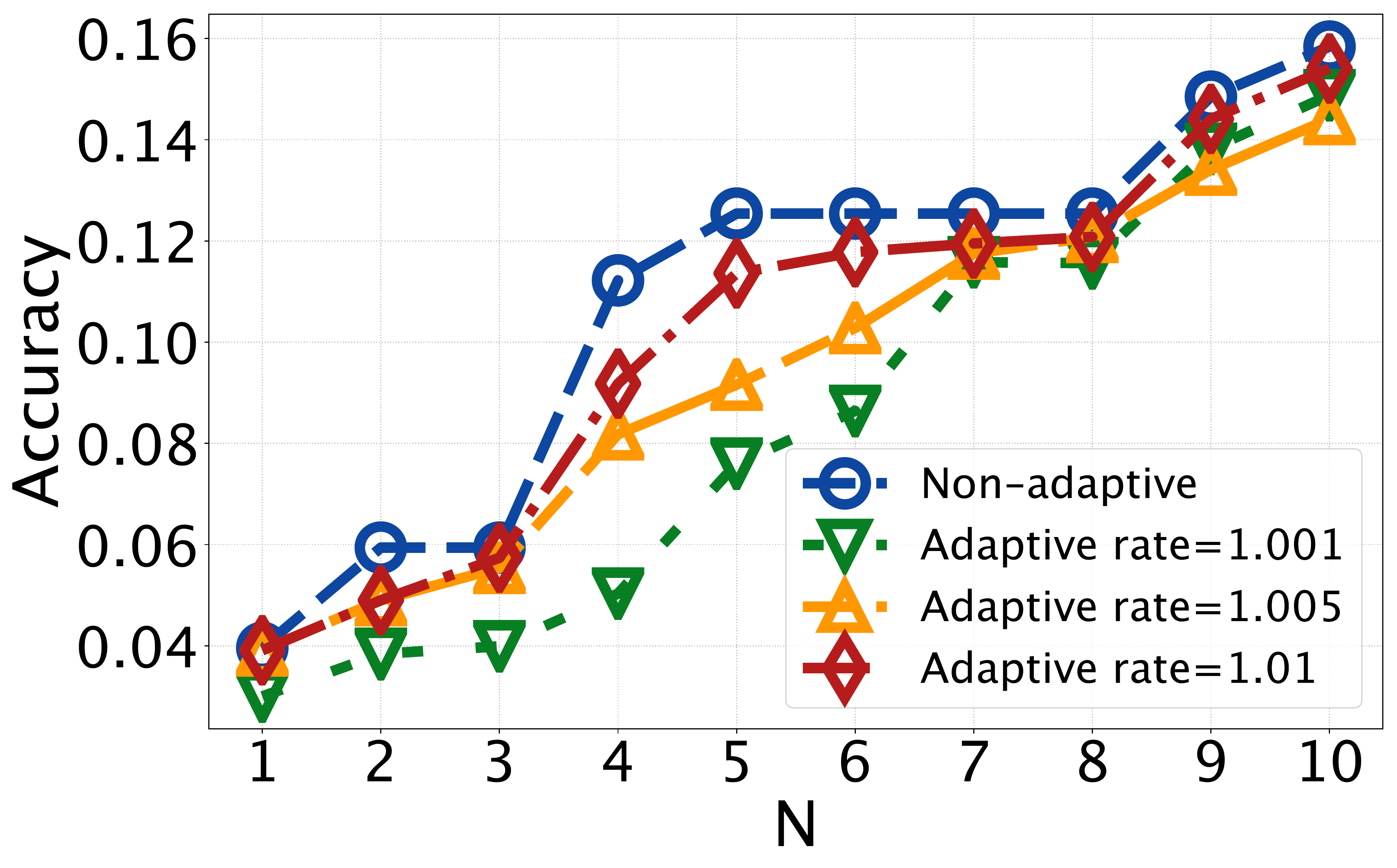}
        \vspace{-7mm}
        \caption{\textsf{ASA} model on Reddit-1 dataset. }
        \label{fig-asa-R1}
    \end{subfigure}
    \begin{subfigure}[t]{.45\linewidth}
        \centering
        \includegraphics[page=2, width=\textwidth]{figures/AMAS/exp/amas_adaptive_accuracy.pdf}
        \vspace{-7mm}
        \caption{\textsf{ASHA} on Reddit-1 dataset. }
        \label{fig-asha-R1}
    \end{subfigure}
    \vspace{-3mm}
    \caption[\amas performance with adaptive training]{The performance comparison between non-adaptive and adaptive sampling.}
    \label{fig-exp-adaptive-perf}
\end{figure}
Our solution has been motivated by use case scenarios observed at Amadeus corporation. For this reason, we work with the log files of an Amadeus~\cite{amadeus} internal application. The log files contain user sessions in the form of attributed sequences. 
Also, we apply our methodology to real-world, publicly available Wikispeedia data~\cite{west2009wikispeedia} and Reddit data~\cite{lakkaraju2013s}. 
For each type of data, we sample two subsets and conduct experiments independently. We summarize the data descriptions as follows:
\begin{itemize}
    \item \textbf{Amadeus data (AMS-1, AMS-2)}\footnote{Personal information is not collected. }. We sampled six datasets from the log files of an internal application at Amadeus IT Group. Each attributed sequence is composed of a user profile containing information (\eg system configuration, office name) and a sequence of function names invoked by web click activities (\eg login, search) ordered by time. 
    \item \textbf{Wikispeedia data (Wiki-1, Wiki-2)}. Wikispeedia is an online game requiring participants to click through from a given start page to an end page using fewest clicks~\cite{west2009wikispeedia}. We select \textit{finished} paths and extract several properties of each path (\eg, the category of the start path, time spent per click). We also sample six datasets from Wikispeedia. The Wikispeedia data is available through the Stanford Network Analysis Project\footnote{https://snap.stanford.edu/data/wikispeedia.html}~\cite{SNAPwiki}.  
    \item \textbf{Reddit data (Reddit-1, Reddit-2)}. Reddit is an online forum. Two datasets that contain the content of reddit submissions are used. The Reddit data is available through the Stanford Network Analysis Project\footnote{https://snap.stanford.edu/data/web-Reddit.html}. 
\end{itemize}
We use 60\% of the instances in each dataset for the training and the rest 40\% for testing. In the training, we holdout 20\% of the training instances for validation. 
\subsection{Compared Methods} We evaluate our two approaches, namely \textsf{ASA} and \textsf{ASHA} and compare them with the following baseline methods. We summarize all compared methods used in this research in Table~\ref{tab-methods}. 

\begin{itemize}
    \setlength{\itemsep}{1pt}
    \item \textsf{BLA} is built using a fully connected neural network to reduce the dimensionality of the input data, and then classify each instance. 
    \item \textsf{BLS} classifies sequences only data using an LSTM. 
    \item \textsf{BLAS} utilizes the information from both attributes and sequences. The resulting embeddings generated by \textsf{BLAS} are then used for classification. 
    \item \textsf{SOA} builds attention on the sequence data for classification, while the attribute data is not used. 
\end{itemize}

\subsection{Experimental Setting}
Our paper focuses on the multi-class classification problem. We thus use accuracy as the metric to evaluate the performance. 
A higher accuracy score depicts more correct predictions of class labels. For each method, we holdout 20\% as the validation dataset randomly selected from the training dataset. For each experimental setting, we report the top-1 $\sim$ top-10 accuracy for each method. 

We initialize our network using the following strategies: orthogonal matrices are used to initialize the recurrent weights, normalized random distribution~\cite{glorot2010understanding} is used to initialize weight matrices in \anetns, and bias vectors are initialized as zero vector $\pmb 0$. 

\subsection{Accuracy Results}
In Figure~\ref{fig-exp-perf}, we compare the performance of our \textsf{ASA} and \textsf{ASHA} solutions with the other state-of-the-art methods in Table~\ref{tab-methods}. \textsf{ASHA} achieves the best performance of top-1 accuracy on most datasets. In most cases, \textsf{ASHA} outperforms other solutions significantly. We also observe a significant performance improvement by \textsf{ASA} compared to other methods. That is, although the top-1 accuracy performance of \textsf{ASA} is beneath that of \textsf{ASHA}, it still outperforms \textsf{SOA} with sequence-only attention and all other methods without attention. The two closest competitors, the \textsf{SOA} utilizing the attention mechanism and classifying each instance based on only the sequential data, and \textsf{BLAS} using information from both attributes and sequences, but without the help from an attention mechanism, are outperformed by our proposed models. 

\subsection{Parameter Sensitivity Analysis}
\begin{figure}[!ht]
    \centering
    \includegraphics[width=\linewidth]{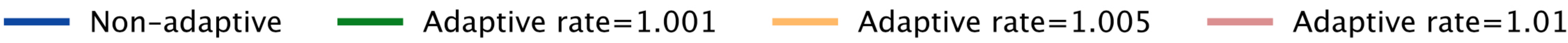}
    \begin{subfigure}[t]{.4\linewidth}
        \centering
        \includegraphics[page=3, width=\textwidth]{figures/AMAS/exp/amas_losses.pdf}
        \vspace{-7mm}
        \caption{\texttt{train_loss} of \textsf{ASHA}. }
        \label{fig-train-loss-asha}
    \end{subfigure}
    \hfill
    \begin{subfigure}[t]{.4\linewidth}
        \centering
        \includegraphics[page=4, width=\textwidth]{figures/AMAS/exp/amas_losses.pdf}
        \vspace{-7mm}
        \caption{\texttt{val_loss} of \textsf{ASHA}. }
        \label{fig-val-loss-asha}
    \end{subfigure}
    \hfill
    \begin{subfigure}[t]{.4\linewidth}
        \centering
        \includegraphics[page=1, width=\textwidth]{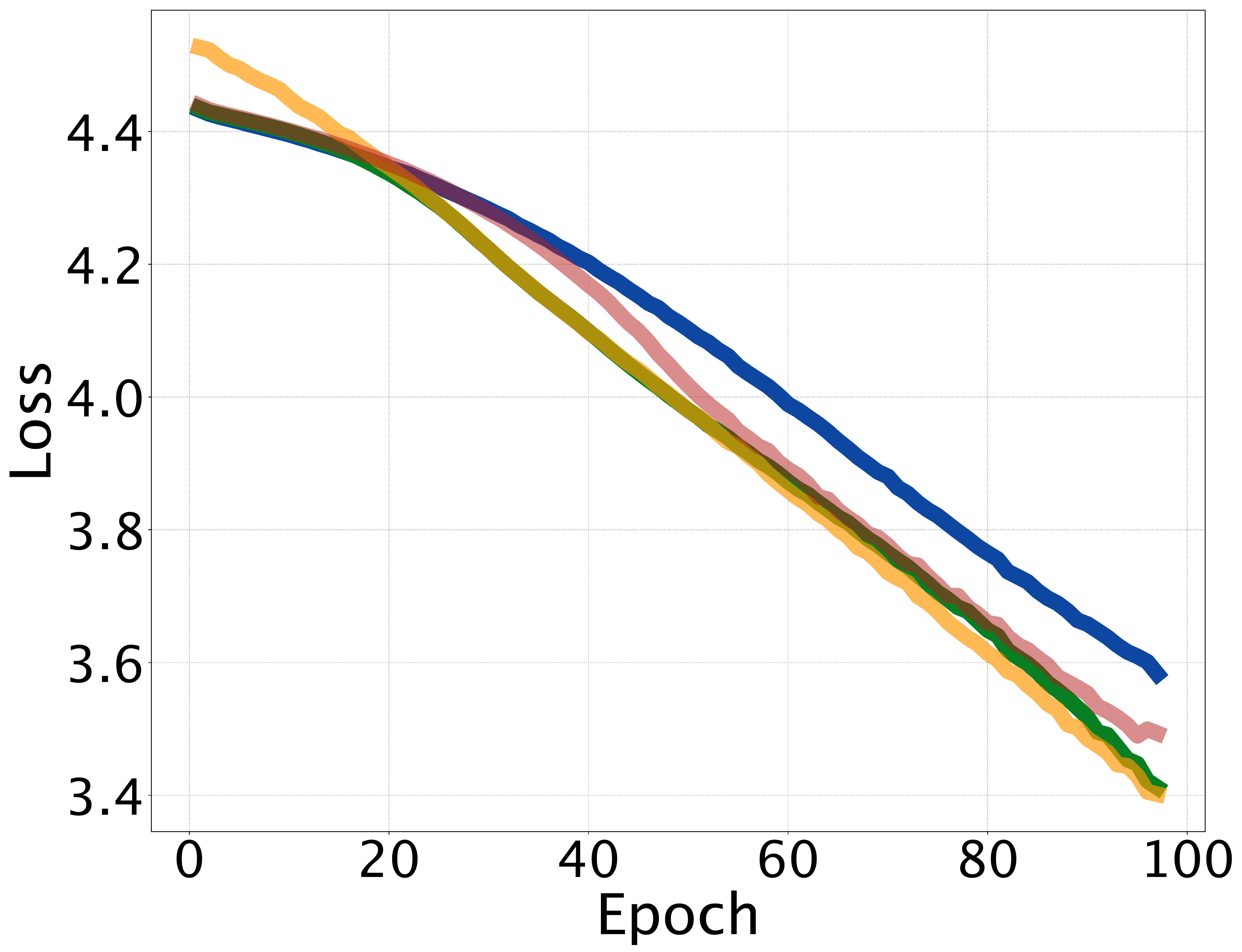}
        \vspace{-7mm}
        \caption[History of training and validation losses]{\texttt{train_loss} of \textsf{ASA}. }
        \label{fig-train-loss-asa}
    \end{subfigure}
    \hfill
    \begin{subfigure}[t]{.4\linewidth}
        \centering
        \includegraphics[page=2, width=\textwidth]{figures/AMAS/exp/amas_losses.pdf}
        \vspace{-7mm}
        \caption{\texttt{val_loss} of \textsf{ASA}. }
        \label{fig-val-loss-asa}
    \end{subfigure}
    \vspace{-3mm}
    \caption{Comparison of the history of training and validation losses.}
    \label{fig-exp-loss}
    \vspace{-5mm}
\end{figure}
\begin{figure}[!ht]
    \centering
    \begin{subfigure}[t]{\linewidth}
        \centering
        \includegraphics[page=1, width=\textwidth]{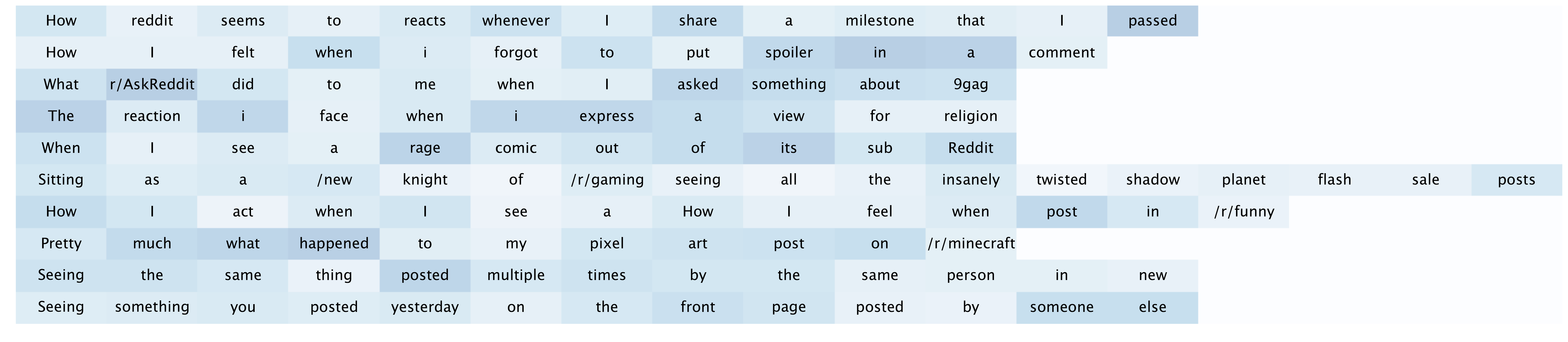}
        \vspace{-7mm}
        \caption{\textsf{ASHA}}
    \end{subfigure}
    \begin{subfigure}[t]{\linewidth}
        \centering
        \includegraphics[page=2, width=\textwidth]{figures/AMAS/exp/amas_cases}
        \vspace{-7mm}
        \caption{\textsf{ASA}}
    \end{subfigure}
    \begin{subfigure}[t]{\linewidth}
        \centering
        \includegraphics[page=3, width=\textwidth]{figures/AMAS/exp/amas_cases}
        \vspace{-7mm}
        \caption{\textsf{SOA}}
    \end{subfigure}
    \vspace{-5mm}
    \caption[Plot of weights in \amas models]{Weights of words from 10 instances in Reddit-2. Higher weights are darker.}
    \vspace{-5mm}
    \label{fig-exp-cases}
\end{figure}
\subsubsection{Adaptive Sampling Accuracy}
As pointed out in recent work~\cite{chen2017outlier}, adaptive sampling is capable of improving the efficiency of the optimization processes by adapting the training sample size in each iteration (\ie epoch). In this set of experiments, we evaluate the two models with varying adaptive sampling rates. We use the adaptive sampling function as: 
\begin{equation*}
    N_\tau = N_{1} \lambda^{(\tau-1)}
\end{equation*}
Here, $\tau$ denotes the epoch number, $N_\tau$ denotes the number of instances used in the $\tau$-th epoch and $\lambda$ the rate of adaptive sampling. We choose $\lambda=1, 1.001, 1.005$, and $1.01$ in our experiments, where $\lambda=1$ means no adaptive sampling. The results presented in Figure~\ref{fig-exp-adaptive-perf} shows that the adaptive sampling with the above sampling rates can achieve similar performance as the non-adaptive approach yet now with much less training data. 

\subsubsection{Training with Adaptive Sampling} With the continuously increasing amount of training instances, we expect the history of training loss to be ``jittery'' when a model encounters previously unseen new instances. 
Different from previous experiments, where we use \texttt{Early Stopping} strategy to avoid overfitting, we now set a fixed number of 144 epochs for the \textsf{ASHA} model and 97 epochs for \textsf{ASA} model and collect the history of training and validation to study the adaptive training strategy. In Figure~\ref{fig-train-loss-asha}, we observe the training with adaptive sampling is more aggressive compared to the non-adaptive approach. 
From Figure~\ref{fig-val-loss-asha} we conclude that with a higher adaptive rate, the model more easily becomes overfitted. Similar conclusion can also be made from Figure~\ref{fig-val-loss-asa}. Selecting a higher adaptive sampling rate can shorten the training time but risking a higher chance of overfitting. 

\subsubsection{Case Studies}
Figure~\ref{fig-exp-cases} demonstrates the weights of each word often instances from the Reddit-2 dataset. Higher weights are represented with a darker color, while lower weights are represented with a lighter color. 
Comparing the three cases, we find that the \textsf{SOA} has the most polarized weights among the three cases. This may be caused by the fact that the attention produced by \textsf{SOA} is solely based on the sequences, while \textsf{ASHA} and \textsf{ASA} have been influenced by attribute data.

 \chapter{Related Work}
\label{chapter-related}
\newpage 

\section{Deep Learning} Deep learning has received significant interests in various research areas in recent years. Deep learning models are capable of feature learning in varying granularities with hierarchical structures. Various deep learning models and optimization techniques have been proposed in a wide range of applications such as image recognition~\cite{karpathy2015deep, xu2015show} and sequence learning~\cite{cho2014learning, sutskever2014sequence, xu2017decoupling, neculoiu2016learning}. Many of these applications involve the learning of single data type~\cite{cho2014learning, sutskever2014sequence, xu2017decoupling, neculoiu2016learning}, as other applications involve more than one data type~\cite{karpathy2015deep, xu2015show}. Several work~\cite{cvpr-face-verify, mueller2016siamese, neculoiu2016learning} focuses on deep metric learning using deep learning techniques. 
The application of deep learning in sequence learning area has numerous work, one of the most popular work, sequence-to-sequence \cite{sutskever2014sequence}, uses long short-term memory model in machine translation.  
The hidden representations of sentences in the source language are transferred to a decoder to reconstruct in the target language. The idea is that the hidden representation can be used as a compact representation to transfer sequence similarities between two sequences. Multi-task learning \cite{luong2015multi} examines three multi-task learning settings for sequence-to-sequence models that aim at sharing either an encoder or decoder in an encoder-decoder model setting. Although the above work is capable of learning the dependencies within a sequence, none of them focuses on learning the dependencies between attributes and sequences. Multimodal deep neural networks~\cite{karpathy2015deep, ngiam2011multimodal, xu2015show} is designed for information sharing across multiple neural networks, but none of these work focuses on our attributed sequence embedding problem.
\section{One-shot Learning} 
One-shot learning has been known to be useful in various applications with very few training data~\cite{bertinetto2016learning, vinyals2016matching, koch2015siamese}. The common problem setting of these tasks is to use one or few training examples to train a model that is capable of generalizing from the training examples and being used to predict the classes of previously unseen test data. The one-shot learning work dates back to work in~\cite{fei2003bayesian} and~\cite{fei2006one}, where the authors use a variational Bayesian framework for one-shot learning to categorize images by using few training examples. Recent work~\cite{bertinetto2016learning, koch2015siamese} use discriminative network structure to train a learner for image classification tasks. Specifically, instead of using the siamese network for verification tasks, \cite{koch2015siamese} uses the siamese network for classification tasks. Concerning the overfitting issue, \cite{bertinetto2016learning} proposes an asymmetric variation of the siamese network, named \textit{learnet}, to further address the issues of using a few training examples. Instead of using the siamese network,~\cite{vinyals2016matching} focuses on using popular memory network and image attention to perform image classification tasks.

\section{Attention Network} 
Attention network~\cite{mnih2014recurrent} has gained a lot research interest recently, the attention network has been applied in various tasks, including image captioning~\cite{xu2015show, nam2016dual}, image generation~\cite{gregor2015draw}, speech recognition~\cite{chorowski2015attention} and document classification~\cite{yang2016hierarchical}. The goal of using attention network in these tasks is to make the neural network focus on the ``interesting'' parts of each input, such as, a small region of an image, or words that are helpful to classifying documents. There are different variations of attention network, including \textit{hierarchical attention}~\cite{yang2016hierarchical} and \textit{dual attentions}~\cite{nam2016dual,zhuang2019amas}.

\section{Sequence Mining} 
Recent work in sequence mining area aims at finding the most frequent subsequence pattern~\cite{miliaraki2013mind, fowkes2016subsequence}. 
Several recent work~\cite{bechet2015sequence, miliaraki2013mind} focus on finding the most frequent subsequence that meets certain constraints. That is, find the set of sequential patterns satisfying various linguistic constraints (e.g., syntactic, symbolic). 
Many sequence mining work focuses on frequent sequence pattern mining. 
Recent work in~\cite{miliaraki2013mind} targets finding subsequences of possible non-consecutive actions constrained by a gap within sequences. \cite{egho2015parameter} aims at solving pattern-based sequence classification problems using a parameter-free algorithm from the model space. It defines rule pattern models and a prior distribution on the model space. 
\cite{fowkes2016subsequence} builds a subsequence interleaving model for mining the most relevant sequential patterns. However, none of them learns sequence embeddings, nor do they support attribute data.  

\section{Clickstream Analysis} 
Recent studies \cite{bernhard2016clickstream, lakshminarayan2016modeling, brinton2016mining} have studied clickstream analysis in various applications. 
Recent work~\cite{bernhard2016clickstream} uses stream mining algorithms to identify frequent sequential patterns and then use the found patterns to build statistical models as Markov chains and transition matrices to capture frequent sequential patterns. 
In~\cite{brinton2016mining}, the authors propose two frameworks using sequences to represent behaviors of students and then uses Markov chains to predict quiz performance. 
Targeting massive open online courses, \cite{lakshminarayan2016modeling} and \cite{brinton2016mining} focus on learning online student behaviors and performances based on clickstreams. 
~\cite{lakshminarayan2016modeling} aims at finding the differences between the designed learning paths and the learning paths of students to improve the engagement and retention of students in online courses. 
~\cite{lakshminarayan2016modeling} handles missing transitions between clicks and predict whether a sequence of clicks will result in an online purchasing. 
However, these work only focus on the clickstreams without the attribute information, nor do they generate embeddings. 

\section{Metric Learning}
Distance metric learning, where the goal is to learn a distance metric from pairs of similar and dissimilar examples, has been studied in various work~\cite{xing2003distance, yeung2007kernel, davis2007information, wang2011integrating, mignon2012pcca, koestinger2012large, cvpr-face-verify, mueller2016siamese, neculoiu2016learning, zhuang2018mlas, zhuang2018olas}. The common objective of these tasks is to learn a distance metric that the distance between similar pairs is reduced while the distance between dissimilar pairs is enlarged as much as possible. Distance metric learning has been used in various tasks to improve the performance of mining tasks, such as clustering~\cite{xing2003distance, yeung2007kernel, davis2007information}. Many applications in various domains also involve distance metric learning, including patient similarity in health informatics~\cite{wang2011integrating}, face verification in computer vision~\cite{mignon2012pcca, koestinger2012large, cvpr-face-verify} and sentence semantic similarity analysis~\cite{mueller2016siamese, neculoiu2016learning}. 
 However, these works only focus on the problem of metric learning on a single data type by using either attribute data or sequential data. Most of them focuses on linear transformation~\cite{wang2011integrating, xing2003distance, yeung2007kernel, davis2007information}. 

 \chapter{Conclusion and Future Work}
\label{chapter-conclusion-future}
\section{Conclusion}
The goal of this dissertation is to study the deep learning applications on attributed sequences. This dissertation studies four problems on the newly proposed attributed sequence data model, including attributed sequence embedding without labels, distance metric learning on attributed sequences, one-shot learning, and using attention model to filter useful information in the attributed sequence data. Here summarize the highlights of this dissertation:

First, we study the problem of \textit{unsupervised attributed sequences embedding}. This work presents the design of the attributed sequence data model and the design of the \nas framework. \nas is a novel deep learning-based framework that generates attributed sequence embeddings in an unsupervised setting. Different from conventional feature learning approaches, which work on either sequences or attributes without considering the \textit{attribute-sequence dependencies}, we identify the three types of dependencies in attributed sequences. Our experiments on real-world tasks demonstrate that the proposed \nas effectively boosts the performance of outlier detection and clustering tasks compared to baseline methods. 

Second, we target at incorporating the pairwise feedback into the embedding. In this work, we focus on the novel problem of distance metric learning on attributed sequences. We propose one \mlas with three solution variations to this problem using neural network models. The proposed \mlas network effectively learns the nonlinear distance metric from both attribute and sequence data, as well as the attribute-sequence dependencies. In our experiments on real-world datasets, we demonstrate the effectiveness of our \mlas network over other state-of-the-art methods in both performance evaluations and case studies. 

Third, we study this new problem of one-shot learning for attributed sequences. We present the \olas network design to tackle the challenges of utilizing this new data type in one-shot learning. \olas incorporates two sub-networks, \cnet, and \vnetns, that integrated into one structure together effectively learn the patterns hidden in this data type using only one example per class. \olas uses this trained knowledge to generate labels for incoming unlabeled instances. Our experiments on real-world datasets demonstrate that \olas on attributed sequences outperforms state-of-the-art one-shot learning methods.

Lastly, we propose a \amas framework with two models for classifying attributed sequences. Our \textsf{ASHA} and \textsf{ASA} models progressively integrate the information from both attributes and sequences while weighing each item in the sequence to improve the classification accuracy. Experimental results demonstrate that our models significantly outperform state-of-the-art methods. 

\section{Future Work}
The prevalence of attributed sequence data and the broad spectrum of real-world applications using attributed sequences motivate us to keep exploring this new direction of research. Based on the work in this dissertation, there are several interesting research directions for future work:

First, in the attributed sequence embedding tasks, we currently only focus on using the information from attributes at the first time step. However, the information from attributes could be lost when encountered long sequences. One straightforward solution is to use attribute embedding to condition the sequence network at each step. Another interesting problem is to answer the question \textit{``how will the embeddings be updated?'' } That is, when the embedding model is updated, the embeddings before the update may be invalid. Other interesting directions include investigating how to building and use the index of a large number of attributed sequence embeddings. 

Second, we now assume the feedback from domain experts is \textit{triplets}, with two attributed sequences and one similarity label. However, feedback can be more complicated than pairwise attributed sequences. That is, domain experts may specify if a group of attributed sequences are similar or dissimilar. Another interesting direction would be using non-binary similarity label, where the similarity label now depicting the degrees of similarity of attributed sequences. There are also different approaches to build models for distance metric learning, such as matching network, worth investigating. 

Third, there is a strong assumption in the one-shot learning task. That is, the unlabeled attributed sequence \textit{must} belong to one of the known categories. However, this assumption may not always stand in real-world applications. For example, people will make fraud attempt that is novel to the model, in which case, the unlabeled attributed sequence may not have a match category. One-shot learning also assumes there is only one sample per category, and this problem can easily be extended to with \textit{no} samples per category in real-world applications. For example, domain experts may know certain types of fraudulent transactions, but without usable cases. It would be interesting to investigate this direction as well. 

Lastly, the goal of using the attention model in attributed sequence classification is to identify and give higher weights to the \textit{relatively} useful items in the sequence. In addition to the two models proposed in this dissertation, one can augment the attribute network and sequence network with other layers to build attention model in various ways. It would be interesting to see the differences among different ways of building attention models.

\clearpage
\singlespacing
\renewcommand{\baselinestretch}{1.5}
 \bibliographystyle{template/siamplain}
 \bibliography{chapters/final}
\end{document}